\documentclass[letterpaper, journal]{IEEEtran}

\IEEEoverridecommandlockouts
\usepackage{algorithm,algorithmic,multirow,hhline}

\newtheorem{theorem}{Theorem}

\newtheorem{lemma}{Lemma}

\usepackage{tabularx}
\usepackage{array}

\usepackage{amsmath}
\allowdisplaybreaks

\usepackage{url}
\usepackage[noadjust]{cite}

\def\ie{\emph{i.e.,\  }}
\def\eg{\emph{e.g.,\  }}

\providecommand{\norm}[1]{\left\|#1\right\|}
\providecommand{\ip}[1]{\boldsymbol{\langle}#1\boldsymbol{\rangle}}

\def\nm{\normalsize}
\usepackage{bm}

\usepackage{amsmath,graphicx}
\usepackage{placeins}
\usepackage{amsmath,epsfig}
\usepackage{epstopdf}
\usepackage{amssymb}

\usepackage{amssymb}
\usepackage{caption}
\usepackage{comment}
\usepackage{subfigure}
\usepackage{pstricks}
\usepackage{subfloat}
\usepackage{setspace}
\DeclareMathOperator{\E}{\mathbb{E}}
\usepackage[left=0.58in, right=0.58in, top=0.71in]{geometry}

\usepackage{xcolor}

\usepackage{dblfloatfix}

	\title{ \fontsize{22}{24}\selectfont Q-GADMM: Quantized Group ADMM for \\Communication Efficient Decentralized Machine Learning}
\author{Anis Elgabli, $^\dagger$Jihong Park, $^\ddagger$Amrit S. Bedi,\\ Chaouki Ben Issaid, Mehdi~Bennis, and $^*$Vaneet~Aggarwal 
\thanks{This work was supported  in part by the INFOTECH Project NOOR, in part by the EU-CHISTERA projects LeadingEdge and CONNECT, and in part by the Academy of Finland through the MISSION and SMARTER projects.} 
\thanks{A. Elgabli, C. Ben Issaid, and M. Bennis are with the Center of Wireless Communication, University of Oulu, Finland (email: \{anis.elgabli, chaouki.benissaid, mehdi.bennis\}@oulu.fi).}
\thanks{$^\dagger$J. Park is with the School of Information Technology, Deakin University, Geelong, VIC 3220, Australia (email: jihong.park@deakin.edu.au).}
\thanks{$^\ddagger$A. Bedi is with the Department of Electrical Engineering, IIT Kanpur (email: amritbd@iitk.ac.in).}
\thanks{$^*$V. Aggarwal is with the School of Industrial Engineering and the School of Electrical and Computer Engineering, Purdue University, USA (email: vaneet@purdue.edu).
}}

\begin{document}
\maketitle

\begin{abstract}
    In this article, we propose a communication-efficient decentralized machine learning (ML) algorithm, coined \emph{quantized group ADMM (Q-GADMM)}. To reduce the number of communication links, every worker in Q-GADMM communicates only with two neighbors, while updating its model via the group alternating direction method of multipliers (GADMM). Moreover, each worker transmits the quantized difference between its current model and its previously quantized model, thereby decreasing the communication payload size. However, due to the lack of centralized entity in decentralized ML, the spatial sparsity and payload compression may incur error propagation, hindering model training convergence. To overcome this, we develop a novel stochastic quantization method to adaptively adjust model quantization levels and their probabilities, while proving the convergence of Q-GADMM for convex objective functions. Furthermore, to demonstrate the feasibility of Q-GADMM for non-convex and stochastic problems, we propose quantized stochastic GADMM (Q-SGADMM) that incorporates deep neural network architectures and stochastic sampling. Simulation results corroborate that Q-GADMM significantly outperforms GADMM in terms of communication efficiency while achieving the same accuracy and convergence speed for a linear regression task. Similarly, for an image classification task using DNN, Q-SGADMM achieves significantly less total communication cost with identical accuracy and convergence speed compared to its counterpart without quantization, i.e., stochastic GADMM (SGADMM).
     
    \end{abstract}
    
    \section{Introduction}
    \label{sec:intro}

    Spurred by a plethora of Internet of things (IoT) and smart devices, the network edge has become a major source of data generation. Exploiting the sheer amount of these user-generated private data is instrumental in training high-accuracy machine learning (ML) models in various domains, ranging from medical diagnosis and disaster/epidemic forecast \cite{Google:FL19} to ultra-reliable and low latency communication (URLLC) and control systems~\cite{park2020extreme,Girgis20:SPAWC,Shiri:CL20}. However, local data is often privacy sensitive (e.g., health records, location history), which prohibits the exchange of raw data samples. 
    
    In view of this, \emph{distributed ML} has recently attracted significant attention \cite{park2018wireless,Park:2019:FLlet,park2020communicationefficient,hosseinalipour2020multi}. A notable example is federated learning (FL)~\cite{mcmahan2017federated,Google:FL19,Smith:FLSurvey}, in which edge devices, i.e., workers, locally train their own ML models that are periodically aggregated and averaged at a parameter server, without exchanging raw data samples. In contrast to classical cloud-based ML, distributed ML hinges on wireless communication and network dynamics, whereby communication may hinder its performance. To obviate this limitation, the communication cost of distributed ML can be decreased by reducing the number of communication \emph{rounds} until convergence, communication \emph{links} per round, and/or the \emph{payload size} per link.
    
    Concretely, to reduce the communication payload size, arithmetic precision of the exchanged parameters can be decreased using 1-bit gradient quantization \cite{Bernstein:2018aa}, multi-bit gradient quantization~\cite{sun2019communication}, or weight quantization with random rotation \cite{pap:jakub16}. Additionally, for large models such as large deep neural networks (DNNs) consisting of millions of model parameters, model outputs can be exchanged via knowledge distillation~\cite{Jeong18,Ahn:2019aa}. To reduce communication links, model updates can be sparsified by collecting model updates until a time deadline~\cite{Wang:2019aa}, based upon significant recent model changes~\cite{chen2018lag,sun2019communication}, or based on channel conditions~\cite{FL_Nishio,YangQuekPoor:2019aa,Chen:20019aa}. Lastly, to reduce the number of communication rounds, convergence speed can be accelerated via collaboratively adjusting the training momentum~\cite{Liu:2019aa,Yu:2019aa}. However, these methods commonly rely on a parameter server that collects all model/gradient updates, which is not scalable in large-scale networks, calling for communication-efficient \emph{decentralized ML} without any central entity.

\begin{figure}[t]
        \centering
        \subfigure[Operational structure of Q-GADMM]{\includegraphics[width=.48\textwidth]{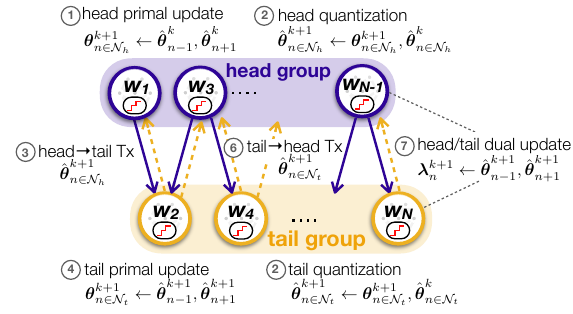}}
        \subfigure[Stochastic quantization in Q-GADMM]{\includegraphics[width=.48\textwidth]{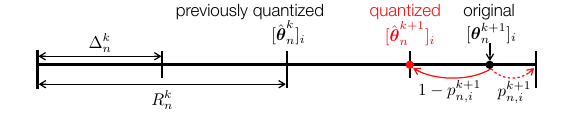}}
    \caption{\small An illustration of (a) quantized GADMM (Q-GADMM) operations, in which at every iteration $k$ every worker communicates with two neighbors, (b) quantizing the difference between the current model and the previously quantized model ($\boldsymbol{\theta}_n^k$ and $\hat{\boldsymbol{\theta}}_n^{k-1}$) with radius $R_n^k$(the infinity norm of the model difference).} \vskip -10pt
    \label{Fig:overview}
    \end{figure}

    A communication-efficient decentralized ML framework, called group ADMM (GADMM), was proposed in our prior work~\cite{elgabli2020gadmm}. Even though GADMM achieves a fast convergence rate and introduces link sparsification by allowing each worker to communicate with only two neighboring nodes, it still suffers from the communication bottleneck between neighboring nodes. We are solving this issue by proposing a novel algorithm, named \emph{quantized GADMM (Q-GADMM)}. Fig.~\ref{Fig:overview}(a) shows the architecture of Q-GADMM which follows the same architecture of GADMM. \ie the workers are divided into head and tail groups in which the workers in the same group update their models in parallel, whereas the workers in different groups update their models in an alternating way via the alternating direction method of multipliers (ADMM)~\cite{boyd2011distributed}. Every alternation entails a single communication round, in which each worker communicates only with two neighboring workers in the opposite group, significantly reducing the communication links. Then, by further integrating model quantization into GADMM, workers in Q-GADMM quantize their model differences (the difference between the current model and the previously quantized model) before every transmission, thereby decreasing the payload size. Hence, Q-GADMM reduces the number of communication rounds, links, and payload size per iteration.

    Quantization errors may lead to high unintended variance in model parameter updates. Its neighbor-based communication aggravates this problem further since the errors may easily propagate across iterations. In the view of this, we provide a stochastic quantization scheme that ensures zero mean and un-biased error at every iteration, and we theoretically analyze the convergence guarantees of Q-GADMM corroborated by numerical evaluation yielding the following contributions.
\begin{itemize}
    \item We propose Q-GADMM, in which a stochastic quantization and a mechanism to redefine the quantization range and the rounding probability to different quantization levels at every iteration are utilized on top of the original GADMM algorithm to ensure zero-mean unbiased error and convergence of Q-GADMM (see Fig.~\ref{Fig:overview}(b) and Algorithm~\ref{alhead}). .
    
    
    \item With exchanging of the quantization range, the quantizer resolution for each worker (heterogeneous quantization), and the quantized difference between the current model and the previously quantized model, we prove that Q-GADMM achieves the optimal solution for convex functions (see Theorem~\ref{theorem}).
    
    \item We additionally propose a stochastic variant of Q-GADMM, coined \emph{quantized stochastic GADMM (Q-SGADMM)}, incorporating deep NN architectures, quantization, and stochastic GADMM (see Sec.~\ref{Q-SGADMM}) to solve modern deep learning problems (stochastic and non-convex problems).
   
    \item For a linear regression task, we numerically validate that Q-GADMM converges as fast as GADMM with significantly less communication cost. Compared to parameter server based baselines using distributed gradient descent (GD), quantized GD (QGD), and Accelerated DIANA~\cite{li2020acceleration}, Q-GADMM achieves faster convergence and significantly less communication cost.

    \item For an image classification task using DNN, we numerically show that Q-SGADMM converges as fast as its variation without quantization, with significantly less communication cost. Compared to parameter server based benchmarks using SGD, and QSGD, Q-SGADMM achieves faster convergence with significantly less communication costs, respectively.
    
\end{itemize}

Note that a preliminary version of this work appeared in \cite{elgabli2019qgadmm}. Compared to this, the current version includes results of more extensive simulation experiments considering more metrics such as the communication energy and more benchmark schemes such as accelerated DIANA. It also includes sensitivity analysis to the hyper-parameters. Finally, it includes detailed convergence analysis with all proofs of Lemma~\ref{lemma1}, Theorem~\ref{theorem0}, and Theorem~\ref{theorem}, while validating the effectiveness of the proposed framework under deep learning architectures, by proposing Q-SGADMM and compare it to SGS and Q-SGD.



\section{Related Works}
\label{related}
Fundamentals of decentralized ML are rooted in decentralized optimization, i.e., distributed optimization without the aid of any central entity. To understand the gist of Q-GADMM compared to prior algorithms, communication-efficient distributed optimization and decentralized optimization frameworks are briefly reviewed as follows.

In the distributed optimization literature, there are a variety of algorithms that are broadly categorized into primal methods \cite{jakovetic2014fast,nedic2014distributed,nedic2009distributed,shi2015proximal} and primal-dual methods \cite{chang2014multi,koppel2017proximity,8624463}. Consensus optimization underlies most of the primal methods, while dual decomposition and ADMM are the most popular among the primal-dual algorithms \cite{glowinski1975approximation,gabay1975dual,boyd2011distributed,jaggi2014communication,ma2017distributed,deng2017parallel}. The performance of distributed optimization algorithms is often characterized by their computation time and communication cost. The computation time is determined by the per-iteration complexity of the algorithm. The communication cost is determined by: (i) the number of \emph{communication rounds} until convergence, (ii) the number of \emph{channel uses} per communication round, and (iii) the \emph{bandwidth/power} usage per channel use. For a large scale network, the communication cost often becomes dominant compared to the computation time, calling for communication efficient distributed optimization~\cite{zhang2012communication,park2018wireless,liu2019communication,nandan2019}.

To reduce the bandwidth/power usage per channel use, decreasing communication payload sizes is one popular solution, which is enabled by gradient quantization \cite{suresh2017distributed}, model parameter quantization \cite{Zhu:2016aa,nandan2019}, and model output exchange for large-sized models via knowledge distillation~\cite{Jeong18}. To reduce the number of channel uses per communication round, exchanging model updates can be restricted only to the workers whose computation delays are less than a target threshold~\cite{Wang:2018aa}, or to the workers whose updates are sufficiently changed from the preceding updates, with respect to gradients~\cite{chen2018lag}, or model parameters~\cite{liu2019communication}. Albeit the improvement in communication efficiency for every iteration $k$, most of the algorithms in this literature are based on distributed gradient descent whose convergence rate is~{$\mathcal{O}(1/k)$} for differentiable and smooth objective functions and can be as low as~ {$\mathcal{O}(1/\sqrt{k})$} (\eg when the objective function is non-differentiable everywhere~\cite{boyd2011distributed}), incurring a large number of communication rounds.

On the other hand, primal-dual decomposition methods are shown to be effective in enabling distributed optimization~\cite{jaggi2014communication,boyd2011distributed,ma2017distributed,glowinski1975approximation,gabay1975dual,deng2017parallel}, among which ADMM is a compelling solution that often provides fast convergence rate with low complexity~\cite{glowinski1975approximation,gabay1975dual,deng2017parallel}. It was shown in~\cite{chen2016direct} that Gauss-Seidel ADMM \cite{glowinski1975approximation} achieves the convergence rate  {$o(1/k)$}. However, all aforementioned distributed algorithms require  a parameter server being connected to every worker, which induces a costly communication link or it may be infeasible particularly for workers located beyond the server's coverage. Moreover, in parameter server based topology, all workers compete over the limited communication resources when transmitting their model updates to the parameter server at every iteration $k$. In sharp contrast, our focus is to develop a decentralized optimization framework ensuring fast convergence without any central entity.

Under decentralized architectures, the convergence of gradient descent and stochastic gradient descent algorithms has been investigated in \cite{Nedic:09,nedic2018network} and \cite{Lian:17}, respectively. To improve the communication efficiency, the communication payload size is reduced by quantizing model updates in \cite{koloskova2019decentralized,gao2020adaptive}, and by dropping some entries of the model updates (e.g., via top-k or random-k sparsification) in \cite{Stich:18, chen2020communicationefficient}. The number of communication links can be reduced by censoring the links of the workers with small model updates \cite{singh2019sparqsgd}. Furthermore, analog transmissions are applied for allowing each worker to utilize the entire bandwidth while transmitting the model updates using analog signals that are superpositioned over-the-air channels for constructing a globally averaged model update \cite{xing2020decentralized}. Nonetheless, in contrast to GADMM and Q-GADMM that allows each worker to communicate with only two neighbors, these methods pre-assume sufficient network connectivity characterized by the symmetric and doubly stochastic connectivity matrix \cite{koloskova2019decentralized,gao2020adaptive, Stich:18}, column stochastic connectivity matrix \cite{chen2020communicationefficient}, and an extended star network topology \cite{xing2020decentralized}. Moreover, the aforementioned decentralized methods fundamentally rely on decentralized gradient descent whose slow convergence rate.

Compared to the aforementioned prior works, Q-GADMM is a quantized version of GADMM from which it inherits the following key features. (i) The alternating update between the head and tail group in GADMM enables the parallel implementation per group and helps to achieve consensus. Under this alternation and connected chain (line), we were able to show that GADMM achieves linear convergence speed. (ii) GADMM sparsifies the communication links for each worker, so each worker needs to hear from only two neighboring nodes at any given iteration, thereby significantly reducing the communication cost. Note also that, as described in \cite{elgabli2020gadmm}, GADMM works under a time-varying topology in which the two neighbors of each worker may change over time, yet the algorithm can still converge to the optimal solution for convex functions. (iii) By alternating between the two groups (heads and tails), at most only half of the workers are competing the limited bandwidth resources compared to all workers in the parameter-server based topology. Hence, more bandwidth is available to each worker when utilizing GADMM. Therefore, by performing quantization on top of GADMM, we are benefitting from the key features of GADMM while further reducing the communication payload per channel use.

Finally, it is worth mentioning that for quantization based schemes, it is effective to quantize the difference between successive gradients rather than each raw gradient, as the differences are commonly smaller than individual gradients, enabling to allocate less bits to represent the desired information. Stochastic rounding is a well-known technique achieving unbiased quantization, but existing works \cite{wen:2017_6749,NIPS2017_Alistarh,Reisizadeh:2019aa,Horvath:19,Stich:18} focus on quantizing individual gradients, not the gradient differences. The quantization method in \cite{sun2019communication} is based on the gradient difference, yet it is biased. To ensure convergence of Q-GADMM, we propose to use unbiased stochastic quantization method for compressing the model differences.
\linespread{1.33}
\section{Problem Formulation and Proposed Algorithm}
\label{probForm}
We consider a set of $N$ workers storing their local batch of input samples. The $n$-th worker has its model vector {$\boldsymbol{\theta}_n \in \mathbb{R}^d$\nm}, and aims to solve the following decentralized learning problem:
\begin{align}
    \underset{\{\boldsymbol{\theta}_1,\boldsymbol{\theta}_2,\cdots,\boldsymbol{\theta}_N\}}{\text{Minimize }} &\sum_{n=1}^N f_n(\boldsymbol{\theta}_n)\nonumber
\\
    \text{subject to  } &\boldsymbol{\theta}_{n} = \boldsymbol{\theta}_{n+1}, \forall n=1,\cdots, N-1.
    \label{com_admm_c1}
\end{align}
In \cite{elgabli2020gadmm}, GADMM is proposed to solve \eqref{com_admm_c1} where the workers are divided into two groups, {\it heads} and {\it tails}, such that each worker in the head (or tail) group is communicating with two tail (or head) workers. In order to obtain the GADMM udpates for the problem in \eqref{com_admm_c1}, consider the augmented Lagrangian for the optimization problem in \eqref{com_admm_c1} as 
\begin{align}
\nonumber &\boldsymbol{\mathcal{L}}_{\rho}(\boldsymbol{\theta},\boldsymbol{\lambda})\!=\!\!\sum_{n=1}^N \!f_n(\boldsymbol{\theta}_n)\!\! +\!\! \sum_{n=1}^{N-1}\!\! \ip{\boldsymbol{\lambda}_n,\boldsymbol{\theta}_{n} \!\!-\! {\boldsymbol{\theta}}_{n+1}}\!\!\\
&+\!\! \frac{\rho}{2}  \sum_{n=1}^{N-1}\! \| \boldsymbol{\theta}_{n} \!-\! {\boldsymbol{\theta}}_{n+1}\|^2.  
\label{augmentedLag3}
\end{align}
As discussed in \cite{elgabli2020gadmm}, let ${\cal N}_h=\{\boldsymbol{\theta}_1, \boldsymbol{\theta}_3,\cdots,\boldsymbol{\theta}_{N-1}\}$\ and ${\cal N}_t=\{\boldsymbol{\theta}_2, \boldsymbol{\theta}_4,\cdots,\boldsymbol{\theta}_{N}\}$ denote the sets of head and tail workers, respectively. The GADMM updates of \cite{elgabli2020gadmm} to solve the problem in \eqref{com_admm_c1} using the augmented Lagrangian of \eqref{augmentedLag3} are given by 
\begin{align}\label{headUpdate0}
 \nonumber & {\boldsymbol{\theta}}_{n \in {\cal N}_h}^{k+1} =\underset{\theta_n}{\text{argmin}}   \Big\{
  f_n(\boldsymbol{\theta}_n) +\ip{{\boldsymbol{\lambda}_{n-1}^{k}}, {\boldsymbol{\theta}}_{n-1}^{k} - \boldsymbol{\theta}_{n}} \\
  &+\ip{{\boldsymbol{\lambda}_{n}^{k}},\boldsymbol{\theta}_{n} \!\!-\!\! {\boldsymbol{\theta}}_{n+1}^{k}} \!\! +\frac{\rho}{2} \| {\boldsymbol{\theta}}_{n-1}^{k}\!-\! \boldsymbol{\theta}_{n}\|^2\!\!+\!\!\frac{\rho}{2} \| \boldsymbol{\theta}_{n}   \!-\! {\boldsymbol{\theta}}_{n+1}^{k}\|^2 \Big\}
\end{align}
for the head workers and 
\begin{align}\label{tailUpdate0}
  \nonumber & {\boldsymbol{\theta}}_{n \in {\cal N}_t}^{k+1} =\underset{\boldsymbol{\theta}_n}{\text{argmin}}  \Big\{f_n(\boldsymbol{\theta}_n) +\ip{{\boldsymbol{\lambda}_{n-1}^{k}},{\boldsymbol{\theta}}_{n-1}^{k+1} - \boldsymbol{\theta}_{n}} \\
  & +\!\ip{{\boldsymbol{\lambda}_{n}^{k}},\boldsymbol{\theta}_{n} \!-\! {\boldsymbol{\theta}}_{n+1}^{k+1}}\!\!
+\frac{\rho}{2} \| {\boldsymbol{\theta}}_{n-1}^{k+1} \!-\! \boldsymbol{\theta}_{n}\|^2\!\!+\!\!\frac{\rho}{2} \| \boldsymbol{\theta}_{n} \!-\! {\boldsymbol{\theta}}_{n+1}^{k+1}\|^2\Big\}
\end{align}
for the tail workers. The corresponding dual update for the GADMM is written as 
\begin{align}
{\boldsymbol{\lambda}}_n^{k+1}={\boldsymbol{\lambda}}_n^{k} + \rho({{\boldsymbol{\theta}}}_{n}^{k+1} - {{\boldsymbol{\theta}}}_{n+1}^{k+1}), \forall n=1,\cdots, N-1.
\label{lambdaUpdate0}
\end{align}
As Fig.~1 illustrates, the primal variables of head workers are updated in parallel, and downloaded by their neighboring tail workers. Likewise, the primal variables of tail workers are updated in parallel, and downloaded by their neighboring head workers. Lastly, the dual variables are updated locally at each worker. GADMM algorithm shows a significant improvement in terms of the communication overhead over the existing ADMM algorithms, but still it needs to receives the latest information $\boldsymbol{\theta}$'s from the neighbors at each worker $n$. For instance, in the primal and dual GADMM updates of \eqref{headUpdate0}-\eqref{lambdaUpdate0}, each worker $n$ requires $\boldsymbol{\theta}_{n-1}$ and $\boldsymbol{\theta}_{n+1}$ to execute one iterate of GADMM algorithm. This creates a communication bottleneck; especially when the dimensions $d$ of $\boldsymbol{\theta}$ is large. We address this issue by proposing the use of stochastic quantization in which we use the quantized version of the information $\hat{\boldsymbol{\theta}}_{n-1}$ and $\hat{\boldsymbol{\theta}}_{n+1}$ to update the primal and dual variables at each worker $n$.

\renewcommand{\baselinestretch}{1}
\subsection{Stochastic Quantization} \label{stochastic_quantixation}
We now describe the overall quantization procedure of Q-GADMM. Worker $n$ in Q-GADMM at iteration $k$ quantizes the difference between the current model and the previous quantized model vector {$\boldsymbol{\theta}_n^k-\boldsymbol{\hat\theta}_n^{k-1}$\nm} as {$Q_n(\boldsymbol{\theta}_n^k- \boldsymbol{\hat\theta}_n^{k-1})$\nm. The function {$Q_n(\cdot)$\nm} is a stochastic quantization operator that depends on the quantization probability $p_{n,i}^k$ for each model vector's dimension $i\in\{1,2,\cdots,d\}$, and on $b_n^k$ bits used for representing each model vector dimension. To ensure the convergence of Q-GADMM, $p_{n,i}^k$ and $b_n^k$ should be properly chosen as detailed below.

As Fig.~1 shows, the $i$-th dimensional element {$[\boldsymbol{\hat\theta}_n^{k-1}]_i$\nm} of the previously quantized model vector is centered at the quantization range {$2 R_n^k$\nm} where $R_n^k=||\boldsymbol{\theta}_n^k - \hat{\boldsymbol{\theta}}_n^{k-1}||_{\infty}$ (the maximum difference between any two elements in $\boldsymbol{\theta}_n^k$ and $\hat{\boldsymbol{\theta}}_n^{k-1}$) and equally divided into $2^{b_n^k}-1$\nm~quantization levels, yielding the quantization step size $\Delta_n^k=2 R_n^k/(2^{b_n^k}-1)$\nm. In this coordinate, the difference between the $i$-th dimensional element {$[\boldsymbol{\theta}_n^k]_i$\nm} of the current model vector and {$[\boldsymbol{\hat\theta}_n^{k-1}]_i$\nm} is
\begin{align}\label{quantization}
[c_n(\boldsymbol\theta_n^k)]_i\!=\! \frac{1}{\Delta_n^k} \left([\boldsymbol\theta_n^k]_i-[\boldsymbol{\hat\theta}_n^{k-1}]_i\!+\!R_n^k\right)\!,
\end{align}
where adding $R_n^k$\nm~ensures the non-negativity of the quantized value. Then, {$[c_n(\boldsymbol\theta_n^k)]_i$\nm} is mapped to:
\begin{align}
[q_n(\boldsymbol\theta_n^k)]_i =\begin{cases}
\left\lceil [c_n(\boldsymbol\theta_n^k)]_i\!\right\rceil & \text{with probability $p_{n,i}^k$}\\[2pt]
\left\lfloor [c_n(\boldsymbol\theta_n^k)]_i\!\right\rfloor & \text{with probability $1-p_{n,i}^k$} \label{Eq:quant}
\end{cases}
\end{align}\nm
where $\lceil\cdot \rceil$ and $\lfloor\cdot \rfloor$ are ceiling and floor functions, respectively. 

Next, we select the probability $p_{n,i}^k$ in \eqref{Eq:quant} such that the expected quantization error is $\E{\left[\boldsymbol{\epsilon}_{n,i}^k\right]}$ is zero \ie $p_{n,i}^k$ should satisfy: 
\begin{align}\label{mean_zero}
\nonumber & p_{n,i}^k \left( [c_n(\boldsymbol\theta_n^k)]_i-\lceil[c_n(\boldsymbol\theta_n^k)]_i \!\rceil \right) \\
& + (1\!-\!p_{n,i}^k) \left( [c_n(\boldsymbol\theta_n^k)]_i -\! \lfloor[c_n(\boldsymbol\theta_n^k)]_i \!\rfloor\! \right) =\!0, 
\end{align}

Solving (8) for $p_{n,i}^k$, we obtain
\begin{align}
p_{n,i}^k = \begin{cases}\frac{\left(\! [c_n(\boldsymbol\theta_n^k)]_i-\lfloor[c_n(\boldsymbol\theta_n^k)]_i \!\rfloor\! \right)}{\left(\! \lceil[c_n(\boldsymbol\theta_n^k)]_i\rceil-\lfloor[c_n(\boldsymbol\theta_n^k)]_i \!\rfloor\! \right)}  \ \ \   \text{ if } [c_n(\boldsymbol\theta_n^k)]_i\notin {\mathbb Z}\\
	0 \ \ \   \text{ otherwise}
	\end{cases}\label{Eq:Optp0}
\end{align}
Note that $\lceil[c_n(\boldsymbol\theta_n^k)]_i\rceil$ and $\lfloor[c_n(\boldsymbol\theta_n^k)]_i \!\rfloor$ are the integer indices of the upper and lower quantization levels for the current value. Therefore, the denominator of \eqref{Eq:Optp0} which is the difference between these consecutive integers is always equal to 1. Hence, \eqref{Eq:Optp0} can be written as:
\begin{align}
p_{n,i}^k = \left(\! [c_n(\boldsymbol\theta_n^k)]_i-\lfloor[c_n(\boldsymbol\theta_n^k)]_i \!\rfloor\! \right)\label{Eq:Optp}
\end{align}

The $p_{n,i}^k$ selection in \eqref{Eq:Optp} ensures that the quantization error is unbiased, yielding the quantization error variance $\E\left[\left(\boldsymbol{\epsilon}_{n,i}^k\right)^2\right] \leq({\Delta_n^k})^2/4$ \cite{reisizadeh2019exact}. This implies that $\E\left[\norm{\boldsymbol{\epsilon}_{n}^k}^2\right]\leq d({\Delta_n^k})^2/4$.

In addition to the above, the convergence of Q-GADMM requires non-increasing quantization step sizes over iterations, \ie $\Delta_n^k \leq \Delta_n^{k-1}$ for all $k$. To satisfy this condition, $b_n^k$ is chosen~as
\begin{align}
b_n^k \geq \left\lceil \log_2\left(1 + (2^{b_n^{k-1}}-1)R_n^k/R_n^{k-1} \right) \right\rceil. \label{Eq:Optb}
\end{align}\nm
Given $p_{n,i}^k$ in \eqref{Eq:Optp} and $b_n^k$ in \eqref{Eq:Optb}, the convergence of Q-GADMM is provided in Sec.\ref{sec:convergence}. We remark that in the numerical simulations (Sec.\ref{sec:sim}), we observe that $R_n^k$ decreases over iterations, and thus $\Delta_n^k \leq \Delta_n^{k-1}$\nm~holds even when $b_n^k$ is fixed.

With the aforementioned stochastic quantization procedure, $b_n^k$\nm, $R_n^k$\nm, and $q_n(\boldsymbol{\theta}_n^k)$ suffice to represent $\boldsymbol{\hat\theta}_n^k$, where
\begin{align}
q_n(\boldsymbol{\theta}_n^k)=( [q_n(\boldsymbol{\theta}_n^k)]_1,[q_n(\boldsymbol{\theta}_n^k)]_2,\cdots,[q_n(\boldsymbol{\theta}_n^k)]_d )^\intercal,
\end{align}
\noindent which are transmitted to neighbors. After receiving these values, $\boldsymbol{\hat\theta}_n^k$\nm~can be reconstructed as follows:
{\begin{align}
\hat{\boldsymbol\theta}_n^k = \boldsymbol{\hat\theta}_n^{k-1}+ \Delta_n^k q_n(\boldsymbol\theta_n^k)-R_n^k\mathbf{1}.
 \label{recoverEq}
 \end{align}\normalsize} 
\noindent Consequently, when the full arithmetic precision uses $32$bit, every transmission payload size of Q-GADMM is $b_n^k d + (b_R + b_b)$ bits, where $b_R\leq 32$ and $b_b\leq 32$ are the required bits to represent $R_n^k$ and $b_n^k$, respectively. Compared to GADMM whose payload size is $32d$ bits, Q-GADMM can achieve a huge reduction in communication overhead, particularly for large models, \ie large $d$.

\subsection{Q-GADMM Operations}

Here we describe the updates for the proposed algorithm to solve the problem in \eqref{com_admm_c1}. Firstly, the head worker's primal variables are updated as
\begin{align}\label{headUpdate}
\nonumber  &{\boldsymbol{\theta}}_{n \in {\cal N}_h}^{k+1} =\underset{\theta_n}{\text{argmin}} \Big\{f_n(\boldsymbol{\theta}_n) +\ip{{\boldsymbol{\lambda}_{n-1}^{k}}, \hat{\boldsymbol{\theta}}_{n-1}^{k} - \boldsymbol{\theta}_{n}} \\
&+\ip{{\boldsymbol{\lambda}_{n}^{k}},\boldsymbol{\theta}_{n} \!\!-\!\! \hat{\boldsymbol{\theta}}_{n+1}^{k}} \!\!
 +\frac{\rho}{2} \| \hat{\boldsymbol{\theta}}_{n-1}^{k} - \boldsymbol{\theta}_{n}\|^2+\frac{\rho}{2} \| \boldsymbol{\theta}_{n} \!-\! \hat{\boldsymbol{\theta}}_{n+1}^{k}\|^2\Big\} 
\end{align}
for all $n \in {\cal N}_h \setminus\{1\}$. For the first worker $n=1$, $\boldsymbol{\theta}_{n-1}$ is not defined since the first head worker does not have a left neighbor, hence the update for $n=1$ is done as
\begin{align}\label{headUpdate1}
{\boldsymbol{\theta}}_{1}^{k+1} =\underset{\theta_{1}}{\text{argmin}} \big\{f_{1}(\boldsymbol{\theta}_{1}) +\ip{{\boldsymbol{\lambda}_{1}^{k}},\boldsymbol{\theta}_{1} \!\!-\!\! \hat{\boldsymbol{\theta}}_{2}^{k}} \!\! +\!\!\frac{\rho}{2} \| \boldsymbol{\theta}_{1} \!-\! \hat{\boldsymbol{\theta}}_{2}^{k}\|^2\big\}. 
\end{align}
\begin{figure}[t]\vskip-10pt

    \begin{algorithm}[H]

      \begin{algorithmic}[1]
       \STATE {\bf Input}: $N, f_n(\boldsymbol{\theta}_n)\forall n, \rho, K$, {\bf Output}: $\boldsymbol{\theta}_n, \forall n$
          \STATE {\bf Initialization}:  $\boldsymbol{\theta}_n^{(0)}=0, \boldsymbol{\lambda}_n^{(0)}=0, \forall n$
          \STATE  ${\cal N}_h=\{\boldsymbol{\theta}_1, \boldsymbol{\theta}_3,\cdots,\boldsymbol{\theta}_{n-1}\}$, ${\cal N}_t=\{\boldsymbol{\theta}_2, \boldsymbol{\theta}_4,\cdots,\boldsymbol{\theta}_{N}\}$

  \WHILE {$k \leq K$}
\STATE { \bf  Head workers ($n \in {\cal N}_h$): in Parallel}
\STATE \quad Reconstruct $\hat{\boldsymbol{\theta}}_{n-1}$ and $\hat{\boldsymbol{\theta}}_{n+1}$ via~\eqref{recoverEq}, and update $\boldsymbol{\theta}_n$ via~\eqref{headUpdate}
\STATE \quad Choose $p_{n,i}^k$ via \eqref{Eq:Optp} and $b_n^k$ via \eqref{Eq:Optb}

\STATE \quad Quantize $\boldsymbol{\theta}_n$ via \eqref{Eq:quant}
\STATE \quad Transmit~$b_n^k, R_n^k, q_n(\boldsymbol{\theta}_n)$ to its two tail neighbors

\STATE { \bf  Tail workers ($n \in {\cal N}_t$): in Parallel}
\STATE \quad Reconstruct $\hat{\boldsymbol{\theta}}_{n-1}$ and $\hat{\boldsymbol{\theta}}_{n+1}$ via~\eqref{recoverEq}, and update $\boldsymbol{\theta}_n$ via~\eqref{tailUpdate}

\STATE \quad Choose $p_{n,i}^k$ via \eqref{Eq:Optp} and $b_n^k$ via \eqref{Eq:Optb}
\STATE \quad  Quantize $\boldsymbol{\theta}_n$ via \eqref{Eq:quant}
\STATE \quad Transmit~$b_n^k, R_n^k, q_n(\boldsymbol{\theta}_n)$ to its two head neighbors
\STATE { \bf  All workers ($n \in \{1,\cdots, N\}$): in Parallel}
\STATE \quad Update $\boldsymbol{\lambda}_{n-1}^{k}$ and $\boldsymbol{\lambda}_n^{k}$ locally via~\eqref{lambdaUpdate}
\STATE $k \leftarrow k+1$
\ENDWHILE 
         \end{algorithmic}
      \caption{Quantized Group  ADMM (Q-GADMM) \label{alhead}}
          
    \end{algorithm}
  \vskip -20pt
\end{figure} 
\noindent Subsequently, each head worker transmits its quantized model to its~two tail neighbors, and the tail workers' primal variables are updated as:
\begin{align}\label{tailUpdate}
\nonumber &  {\boldsymbol{\theta}}_{n \in {\cal N}_t}^{k+1} =\underset{\boldsymbol{\theta}_n}{\text{argmin}}  \big\{f_n(\boldsymbol{\theta}_n) +\ip{{\boldsymbol{\lambda}_{n-1}^{k}},\hat{\boldsymbol{\theta}}_{n-1}^{k+1} - \boldsymbol{\theta}_{n}} \\
&+\!\ip{{\boldsymbol{\lambda}_{n}^{k}},\boldsymbol{\theta}_{n} \!-\! \hat{\boldsymbol{\theta}}_{n+1}^{k+1}}+\frac{\rho}{2} \| \hat{\boldsymbol{\theta}}_{n-1}^{k+1} \!-\! \boldsymbol{\theta}_{n}\|^2\!\!+\!\!\frac{\rho}{2} \| \boldsymbol{\theta}_{n} \!-\! \hat{\boldsymbol{\theta}}_{n+1}^{k+1}\|^2\big\}
\end{align}
for $n \in {\cal N}_t \setminus\{N\}$. For the last worker $n=N$, $\boldsymbol{\theta}_{N+1}$ is not defined. Hence, the update of the last tail worker $n=N$ is given by
\begin{align}\label{tailUpdate2}
\nonumber &{\boldsymbol{\theta}}_{N}^{k+1} =\underset{\boldsymbol{\theta}_N}{\text{argmin}}  \Big\{f_N(\boldsymbol{\theta}_N) +\ip{{\boldsymbol{\lambda}_{N-1}^{k}},\hat{\boldsymbol{\theta}}_{N-1}^{k+1} - \boldsymbol{\theta}_{N}} \\
&+ \!\! \frac{\rho}{2} \| \hat{\boldsymbol{\theta}}_{N-1}^{k+1} \!-\! \boldsymbol{\theta}_{N}\|^2\Big\}.
\end{align}\normalsize
Finally, every worker locally updates its dual variables $\boldsymbol{\lambda}_{n-1}$\nm~and $\boldsymbol{\lambda}_n$ utilizing the quantized information from its neighbors as follows
\begin{equation}
{\boldsymbol{\lambda}}_n^{k+1}={\boldsymbol{\lambda}}_n^{k} + \rho(\hat{{\boldsymbol{\theta}}}_{n}^{k+1} - \hat{{\boldsymbol{\theta}}}_{n+1}^{k+1})
\label{lambdaUpdate}
\end{equation}
for $n=1,\cdots, N-1$. The primal and the dual update procedure discussed in \eqref{headUpdate}-\eqref{lambdaUpdate} is summarized in Algorithm~\ref{alhead}.

\begin{figure*}
  \setcounter{equation}{27}
  \begin{align}
  \textsf{LB}_1&= -\sum_{n=1}^{N-1}\E\left[\langle \boldsymbol{\lambda}_n^*,  \boldsymbol{r}_{n,n+1}^{k+1} \rangle\right] \label{Eq: Lemma1_LB0}
  \\
   \textsf{UB}_1& = -\sum_{n=1}^{N-1} \E\left[\langle \boldsymbol{\lambda}_n^{k+1}, \boldsymbol{r}_{n,n+1}^{k+1} \rangle\right] - 2 \rho \sum_{n=1}^{N-1} \E\left[\langle \boldsymbol{\epsilon}_n^{k+1},\boldsymbol{\theta}_n^*-\boldsymbol{\theta}_n^{k+1} \rangle \right] +\sum_{n\in {\cal N}_h} \E\left[\langle \boldsymbol{s}_{n}^{k+1}, \boldsymbol{\theta}_n^*-\boldsymbol{\theta}_n^{k+1} \rangle \right]. \label{Eq: Lemma1_UB0}
   \end{align}
   \hrulefill
   \vspace{-0.5cm}
   \setcounter{equation}{18}
  \end{figure*}

\section{Convergence Analysis}
\label{sec:convergence}
In this section, we prove the optimality and convergence of Q-GADMM for convex functions. The necessary and sufficient optimality conditions are the primal and dual feasibility which are defined by
\begin{itemize}
\item \textbf{Primal feasibility}
\begin{align}
\boldsymbol{\theta}_n^* = \boldsymbol{\theta}_{n-1}^*, \forall n > 1, \label{primal_feas}
\end{align}
\item \textbf{Dual feasibility}
\begin{align}
&\boldsymbol{0} \in \partial f_n(\boldsymbol{\theta}_n^*) - \boldsymbol{\lambda}_{n-1}^* + \boldsymbol{\lambda}_{n}^* , ~ n \in\{2,\cdots,N-1\}, \nonumber
\\
&\boldsymbol{0} \in \partial f_n(\boldsymbol{\theta}_n^*)  + \boldsymbol{\lambda}_{n}^* , \quad\quad\quad\quad n=1, \nonumber
\\
&\boldsymbol{0} \in \partial f_n(\boldsymbol{\theta}_n^*) - \boldsymbol{\lambda}_{n-1}^*,  \quad\quad\quad n=N,
\label{eq2}
\end{align}
\end{itemize}
\noindent Firstly, note that  at iteration $k+1$, every $\boldsymbol{\theta}_{n}^{k+1}$ for $n\in {\cal N}_t \setminus\{N\}$ minimizes \eqref{tailUpdate}, which implies that
\begin{align}
\nonumber &\boldsymbol{0}\! \in \partial f_n(\boldsymbol{\theta}_{n}^{k+1}) \!-\! \boldsymbol{\lambda}_{n-1}^{k} \!\!+\!\! \boldsymbol{\lambda}_{n}^{k} \!\!+\!\! \rho (\boldsymbol{\theta}_{n}^{k+1} \!-\! \hat{\boldsymbol{\theta}}_{n-1}^{k+1})\\
&+ \rho (\boldsymbol{\theta}_{n}^{k+1} \!\!-\!\! \hat{\boldsymbol{\theta}}_{n+1}^{k+1}).
\label{eq3}
\end{align}
Since  $\boldsymbol{\epsilon}_n^{k+1}=\boldsymbol{\theta}_{n}^{k+1}- \hat{\boldsymbol{\theta}}_{n}^{k+1}$ is the quantization error at iteration $k+1$, using \eqref{lambdaUpdate}, the result in \eqref{eq3} can be re-written as
%
%
\begin{align}
\boldsymbol{0} \in \partial f_n(\boldsymbol{\theta}_{n}^{k+1}) - \boldsymbol{\lambda}_{n-1}^{k+1}+ \boldsymbol{\lambda}_{n}^{k+1}+2\rho \boldsymbol{\epsilon}_n^{k+1}
\end{align}
Similarly, the optimality condition for the last tail worker ($n=N$) is given by 
\begin{align}
\boldsymbol{0} \in \partial f_N(\boldsymbol{\theta}_{N}^{k+1}) - \boldsymbol{\lambda}_{N-1}^{k+1}+2\rho \boldsymbol{\epsilon}_N^{k+1}.
\end{align}
Secondly, every $\boldsymbol{\theta}_{n}^{k+1}$ for $n\in {\cal N}_h' \triangleq {\cal N}_h \setminus \{1\}$ minimizes \eqref{headUpdate} at iteration $k+1$. Hence, it holds that 
\begin{align}
\nonumber &\boldsymbol{0} \!\in \partial f_n(\boldsymbol{\theta}_{n}^{k+1}) \!\!-\! \boldsymbol{\lambda}_{n-1}^{k} \!\!+\!\! \boldsymbol{\lambda}_{n}^{k} \!\!+\!\! \rho (\boldsymbol{\theta}_{n}^{k+1} \!-\! \hat{\boldsymbol{\theta}}_{n-1}^{k})\\
&+ \rho (\boldsymbol{\theta}_{n}^{k+1} \!\!-\! \hat{\boldsymbol{\theta}}_{n+1}^{k}).
\label{eqAnis}
\end{align}\normalsize
We can rewrite \eqref{eqAnis} using $\boldsymbol{\epsilon}_n^{k+1}$ and ${\boldsymbol{\lambda}}_n^{k+1}$ in \eqref{lambdaUpdate} as follows
\begin{align}\label{dual_res_1}
\nonumber &\boldsymbol{0}\in \partial f_n(\boldsymbol{\theta}_{n}^{k+1}) - \boldsymbol{\lambda}_{n-1}^{k+1}+ \boldsymbol{\lambda}_{n}^{k+1}+2\rho\boldsymbol{\epsilon}_n^{k+1} \\
&+\rho(\hat{\boldsymbol{\theta}}_{n-1}^{k+1}-\hat{\boldsymbol{\theta}}_{n-1}^{k})+\rho(\hat{\boldsymbol{\theta}}_{n+1}^{k+1}-\hat{\boldsymbol{\theta}}_{n+1}^{k})
\end{align}

Next, the optimality condition for the first head worker ($n=1$) is
\begin{align}\label{dual_residual_2}
\boldsymbol{0}\in \partial f_1(\boldsymbol{\theta}_{1}^{k+1}) + \boldsymbol{\lambda}_{1}^{k+1}+2\rho\boldsymbol{\epsilon}_1^{k+1} +\rho(\hat{\boldsymbol{\theta}}_{2}^{k+1}-\hat{\boldsymbol{\theta}}_{2}^{k}).
\end{align}
In \eqref{dual_res_1} and \eqref{dual_residual_2}, let $\boldsymbol{s}_{n}^{k+1}$ be the dual residual of worker $n\in {\cal N}_h$ at iteration $k+1$ defined as 
\small
\begin{align}\label{cases}
\boldsymbol{s}_{n}^{k+1}= \begin{cases} \rho(\hat{\boldsymbol{\theta}}_{n-1}^{k+1}-\hat{\boldsymbol{\theta}}_{n-1}^{k})+ \rho(\hat{\boldsymbol{\theta}}_{n+1}^{k+1}-\hat{\boldsymbol{\theta}}_{n+1}^{k}),\ \text{for}\ \ n \in {\cal N}_h' , \\ \rho(\hat{\boldsymbol{\theta}}_{n+1}^{k+1}-\hat{\boldsymbol{\theta}}_{n+1}^{k}),\quad\quad\quad\quad\quad\quad\quad\quad \text{for}\ \ n =1. 
\end{cases}
\end{align} 
\normalsize
Consider the primal feasibility in \eqref{primal_feas} and let us define the corresponding primal residual at iteration $k+1$ as
$\boldsymbol{r}_{n-1,n}^{k+1}= {\boldsymbol{\theta}}_{n-1}^{k+1}-{\boldsymbol{\theta}}_{n}^{k+1}$ and $\boldsymbol{r}_{n,n+1}^{k+1}= {\boldsymbol{\theta}}_{n}^{k+1}-{\boldsymbol{\theta}}_{n+1}^{k+1}$. In order to show the convergence of Q-GADMM, we need to show that the optimality conditions mentioned in \eqref{primal_feas}-\eqref{eq2} are satisfied. For that, it is required to prove that the primal ($\boldsymbol{r}_{n-1,n}^{k+1}$) and dual ($\boldsymbol{s}_{n}^{k+1}$) residuals  converge to zero as $k\rightarrow\infty$. Further, we need to prove that the objective function $\sum_{n=1}^Nf_n(\boldsymbol{\theta}_{n}^{k+1})$ converges to the optimal value $\sum_{n=1}^Nf_n(\boldsymbol{\theta}_n^*)$ in the limit. Next, we state the first Lemma we need to prove the convergence of the proposed algorithm.

\begin{lemma}\label{lemma1}
\normalfont
At $k+1$ iteration of Q-GADMM, the optimality~gap satisfies
  \setcounter{equation}{29}
  \begin{equation}
  {\textsf{LB}_1 \leq \sum_{n=1}^N\E[f_n(\boldsymbol{\theta}_{n}^{k+1})-f_n(\boldsymbol{\theta}_n^*)] \leq\textsf{UB}_1\normalsize},
  \end{equation}
   where {$\textsf{LB}_1$} and {$\textsf{UB}_1$} are the lower and upper bounds, respectively, given by \eqref{Eq: Lemma1_LB0} and \eqref{Eq: Lemma1_UB0}.
\end{lemma}
 The proof of Lemma \ref{lemma1} is provided in Appendix \ref{sec:lem1}. Next, in order to prove the convergence of the proposed algorithm, let us define a Lyapunov function $V_k$ as follows 
\begin{align}
\nonumber &  V^k = \frac{1}{\rho} \sum_{n=1}^{N-1} \| \boldsymbol{\lambda}_n^{k}-\boldsymbol{\lambda}^* \|^2 + \rho \sum_{n\in{\cal N}_h'} \|\boldsymbol{\theta}_{n-1}^{k}-\boldsymbol{\theta}^*\|\\
&  + \rho \sum_{n\in{\cal N}_h} \|\boldsymbol{\theta}_{n+1}^{k}-\boldsymbol{\theta}^*\|.
\end{align}
Then, to show the convergence, we need to prove that $\E\left[V^{k}-V^{k+1}\right]$ is bounded above which is presented in Theorem \ref{theorem0} next. 
\begin{theorem}\label{theorem0}
When $f_n(\boldsymbol{\theta}_n)$ is closed, proper, and convex, and the Lagrangian $\boldsymbol{\mathcal{L}}_{0}$ has a saddle point, then the following inequality holds true at the $(k+1)$-th iteration of Q-GADMM
\small
\begin{align}
\nonumber &\E{}\left[V^k - V^{k+1} \right] + \frac{d \rho}{2}  \sum_{n\in {\cal N}_t}{\left(\Delta_n^{k+1}\right)^2}  \\
& \geq \sum_{n\in {\cal N}_h'}\E{}\left[ \rho\| \boldsymbol{\theta}_{n-1}^{k+1}-\boldsymbol{\theta}_{n-1}^k-(\boldsymbol{\epsilon}_{n-1}^{k+1}-\boldsymbol{\epsilon}_{n-1}^{k})-\boldsymbol{r}_{n-1,n}^{k+1} \|^2 \right]
\nonumber
\\
& + \sum_{n\in {\cal N}_h}\E{}\left[\rho\| (\boldsymbol{\theta}_{n+1}^{k+1}-\boldsymbol{\theta}_{n+1}^k)-(\boldsymbol{\epsilon}_{n+1}^{k+1}-\boldsymbol{\epsilon}_{n+1}^{k})+\boldsymbol{r}_{n,n+1}^{k+1} \|^2 \right].
\label{lemma3_FinalEq}
\end{align} 
\end{theorem}
\normalsize
The proof of Theorem \ref{theorem0} is detailed in Appendix \ref{sec:lem2}. Using Lemmas \ref{lemma1} and \ref{theorem0}, next we show that $\E{}[V^k]$ decreases monotonically with the iteration index $k$ and both the primal and dual residuals go to zero using Q-GADMM. We thereby present the main theorem stating the convergence and the optimality of Q-GADMM in solving \eqref{com_admm_c1} as follows.

\begin{theorem}\label{theorem}
Under the assumptions of Theorem \ref{theorem0},  and provided that the quantization step sizes is a non-increasing sequence, \ie {$\Delta_n^{k} \leq \Delta_n^{k-1}\; $} for all $k$, such that $\sum_{k=0}^{\infty}\Delta_n^{k} < \infty$, for all $n$, the primal and dual residual converges to $\boldsymbol{0}$ in the mean square sense as $k\rightarrow\infty$, i.e. , $$ \lim_{k\rightarrow\infty} \left[\|\boldsymbol{r}_{n,n+1}^{k}\|^2\right] = 0 \ \text{and} \lim_{k\rightarrow\infty} \left[\|\boldsymbol{s}_{n}^{k}\|^2\right]  = 0.$$ Furthermore, the optimality gap converges to $0$ in the mean sense, \ie $$\lim_{k\rightarrow\infty} \left[\sum_{n=1}^N f_n(\boldsymbol{\theta}_n^{k})\right]= \sum_{n=1}^Nf_n(\boldsymbol{\theta}^*).$$\nm
\end{theorem}
The proof of Theorem \ref{theorem} is deferred to Appendix \ref{sec:thm2}.

\section{Numerical Results}
\label{sec:sim}
To validate the performance of Q-GADMM and to gauge its communication efficiency compared to GADMM, we conduct numerical experiments for distributed machine learning tasks for the convex problem of linear regression. We also evaluated the performance of its stochastic version (Q-SGADMM) for the stochastic and non-convex problem of image classification using deep neural networks (DNN). We compare Q-GADMM and Q-SGADMM with the following benchmark algorithms.

\begin{itemize}
    \item \textbf{GADMM}~\cite{elgabli2020gadmm} and \textbf{SGADMM}: The original full precision GADMM in which $32$bits are used to transmit each model's element at every iteration, and its stochastic version
    \item \textbf{GD} and \textbf{SGD}: Distributed gradient descent algorithm in which at every iteration, each worker computes the gradient of its local function with respect to a global model and uploads it to the parameter server which updates the global model using a one global gradient descent step, and its stochastic version
    \item \textbf{QGD} and \textbf{QSGD}: Quantized version of GD, and its stochastic version.
    \item \textbf{ADIANA}~\cite{li2020acceleration} (Accelerated DIANA): ADIANA enjoys faster convergence compared to GD/SGD with less number of transmitted bits since  every worker quantizes its model. However, every worker needs to share more variables per iteration (the gradient vector with respect to the current model, and the gradient vector with respect to another variable that is introduced to accelerate the convergence).  
\end{itemize}

\begin{figure*}[t]
\centering
\includegraphics[width=\textwidth]{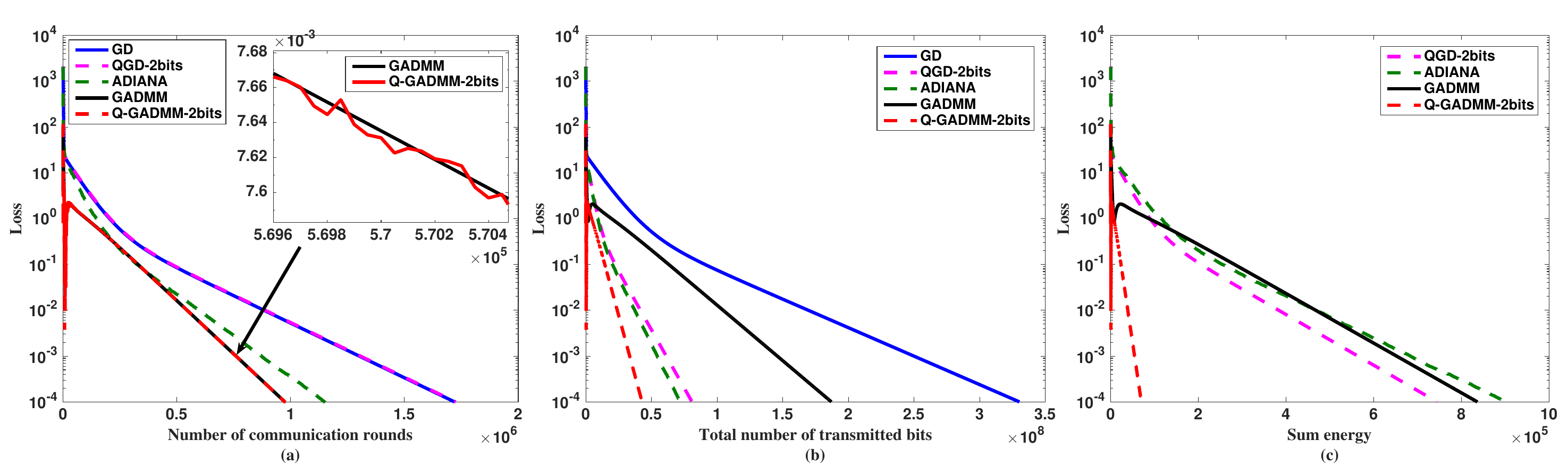}
\caption{\emph{Linear regression} results showing: (a) loss $(|F-F^*|)$ w.r.t. \# number of communication rounds; (b) loss w.r.t. number of transmitted bits; and (c) energy efficiency (loss w.r.t consumed energy).}
\label{fig:gdLR} \vskip -10pt
\end{figure*}

\begin{figure*}[t]
\centering
\includegraphics[width=\textwidth]{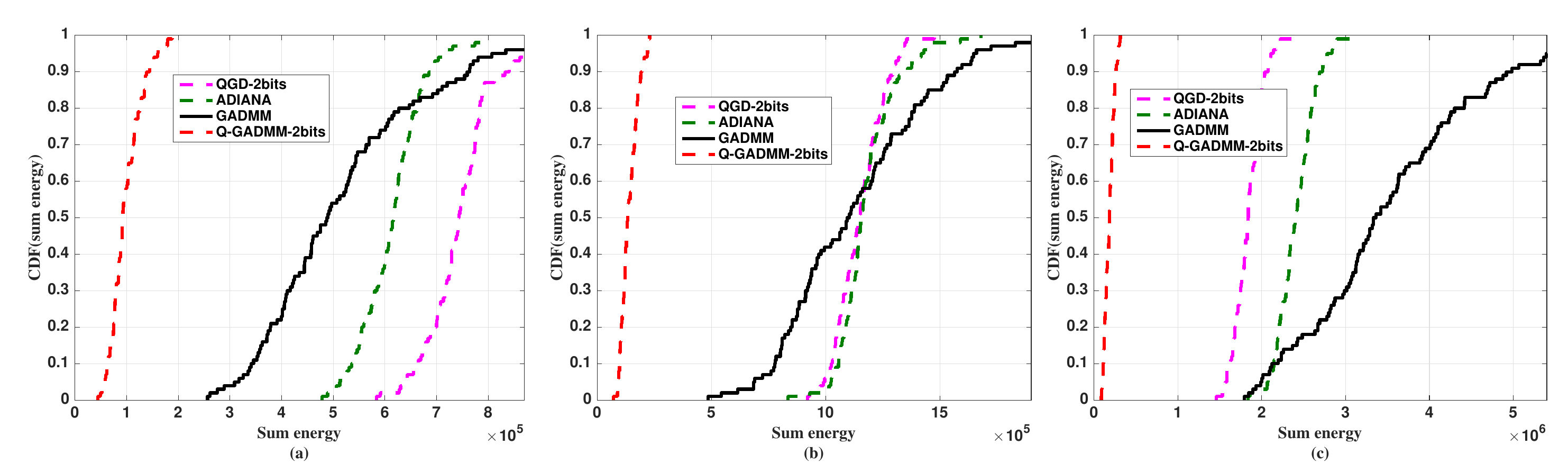}
\caption{\emph{Linear regression} results showing: CDF of the consumed energy to achieve the loss value of $10^{-4}$, (a) system bandwidth is 10MHz, (b) system bandwidth is 2MHz, and (c) system bandwidth is 1MHz}
\label{fig:enLR} \vskip -10pt
\end{figure*}

\subsection{Numerical results  for the convex problem of linear regression}

\subsubsection{Simulation settings} We evaluate the performance of Q-GADMM for decentralized linear regression using the California Housing dataset \cite{Torgo:14} which consists of $20,000$ samples, each has six features, i.e., the model size $d=6$. We uniformly distribute the samples across $50$ workers. Note that, in GD, QGD, and ADIANA, each iteration involves $N+1$ communication rounds. i.e., N uploads ($N$ workers upload their gradient vectors) and one download from the parameter server (PS). On the other hand, for GADMM and QGADMM, each iteration involves $N$ communication rounds since each worker needs to broadcast its updated model.  Finally, for all quantized algorithms, we assume that the quantizer resolution is equal to 2 (4 quantization levels), and it remains constant over iterations and across workers (i.e., $b_n^k=2\forall n,k$). Finally, we choose $\rho=24$.

We drop $50$ workers randomly in a 250x250$m^2$ grid, and for the PS-based algorithms, we choose the worker with the minimum sum distance to all workers as the PS. For the decentralized algorithms (GADMM/QGADMM), we implement the heuristic described in \cite{elgabli2020gadmm} to find the neighbors of each worker.  We assume that the total system bandwidth is $2$MHz equally divided across workers. Therefore, the available bandwidth to the $n$-th worker ($B_n$) at every communication round when utilizing GADMM and Q-GADMM is $(4/N)$MHz since only half of the workers are transmitting at each communication round. On the other hand, the available bandwidth to each worker when using GD, QGD, and A-DIANA  is $(2/N)$MHz. The power spectral density ($N_0$) is $10^{-6}$W$/$Hz, and each upload/download transmission time ($\tau$) is $1ms$. We assume a free space model, and each worker needs to transmit at a power level that allows uploading the gradient vector in one communication round. For example, using GD, each worker needs to find the transmission power that achieves the transmission rate $R=(32d/1ms)$ bits/sec. Therefore, using Shannon capacity, the corresponding transmission power can be calculated as $P=D^2\cdot N_0 \cdot B_n \cdot(2^{(R/B_n)}-1)$, and the consumed energy will be $E=P \cdot \tau$.

\subsubsection{Result discussion} Figure~\ref{fig:gdLR}(a) plots the loss function which is defined as the absolute difference between the objective function at communication round $k$ and the optimal solution of the problem. Clearly, Figure~\ref{fig:gdLR}(a) verifies Theorem~\ref{theorem}, and shows the convergence of Q-GADMM for convex loss functions. Moreover, it shows that both GADMM and Q-GADMM-2bits achieve the loss of $10^{-4}$ at almost the same number of communication rounds outperforming the considered baselines. 

We plot  in Figure~\ref{fig:gdLR}(b) the loss value versus the total number of transmitted bits. Figure~\ref{fig:gdLR}(b) shows that the quantized algorithms require less number of bits to achieve the loss value of $10^{-4}$ as compared to GADMM and GD. Moreover, Q-GADMM is shown to achieve the minimum number of transmitted bits since it enjoys the fastest convergence speed and transmits the lower number of bits per iteration. Note that in A-DIANA, each worker transmits two variables at every iteration, namely $32+2\cdot d\cdot 2$ bits versus $32+d\cdot 2$ bits for Q-GADMM. Moreover, in A-DIANA and QGD, the PS transmits $32\cdot d$ bits at every iteration. 

To draw a comparison  in terms of the actual consumed energy in the system, in Figure~\ref{fig:gdLR}(c), we plot the loss versus the total sum energy (all workers' uplink energy $+$ server's downlink energy). 
Figure~\ref{fig:gdLR}(c) shows significant reduction in the total consumed energy for Q-GADMM compared to all baselines, thanks to the decentralization in which each worker communicates with only nearby neighbors, the fast convergence inherited from GADMM, and the stochastic quantization which reduces the number of transmitted bits at every iteration significantly while ensuring convergence.

In Fig.~\ref{fig:enLR}, we plot the CDF of the total consumed energy of all algorithms after conducting $100$ experiments, where at each experiment, we randomly drop $50$ workers and run all algorithms. We plot the results in Figs.~\ref{fig:enLR}(a)-(c) for three choices of the system bandwidth $10$, $2$, and $1$ MHz respectively. We make two observations from Figs.~\ref{fig:enLR}(a)-(c). First, at high bandwidth regime, even GADMM, which relies only on topology decentralization, can outperform the quantized PS-based algorithms since each worker has enough bandwidth and requires significantly less transmission power to communicate with its neighbors compared to communicating to a central PS. Second, Q-GADMM achieves the minimum energy cost in all experiments and maintains the same performance gap compared to all PS-based baselines in all bandwidth regimes.


\subsection{Numerical results for the non-convex problem of Image classification using DNN} 
\label{Q-SGADMM}

 \begin{figure*}
\centering
\includegraphics[trim=0.3in 0.2in 0.2in 0in, clip,  width=\textwidth]{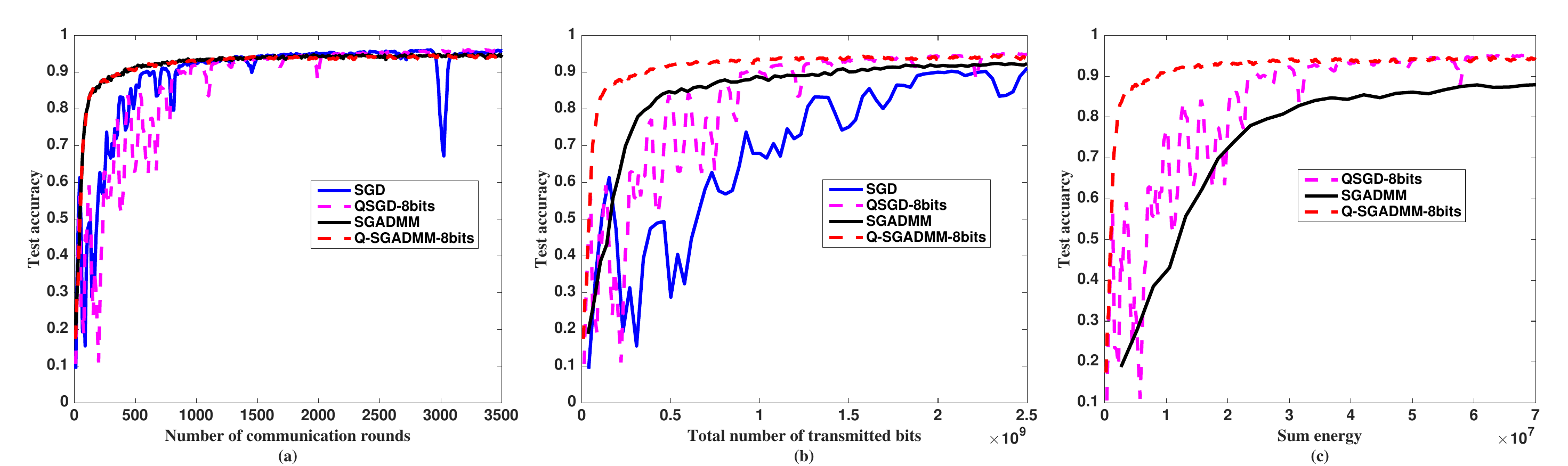} 
\caption{\emph{Image classification} results showing: (a) test accuracy w.r.t. \# number of communication rounds; (b) test accuracy w.r.t. number of transmitted bits; and (c) energy efficiency (test accuracy w.r.t consumed energy) when the system bandwidth is 40MHz.}
\label{fig:dnn} \vskip -10pt
\end{figure*}
\begin{figure*}[t]
\centering
\includegraphics[trim=0.1in 0.2in 0.2in 0in, clip,  width=\textwidth]{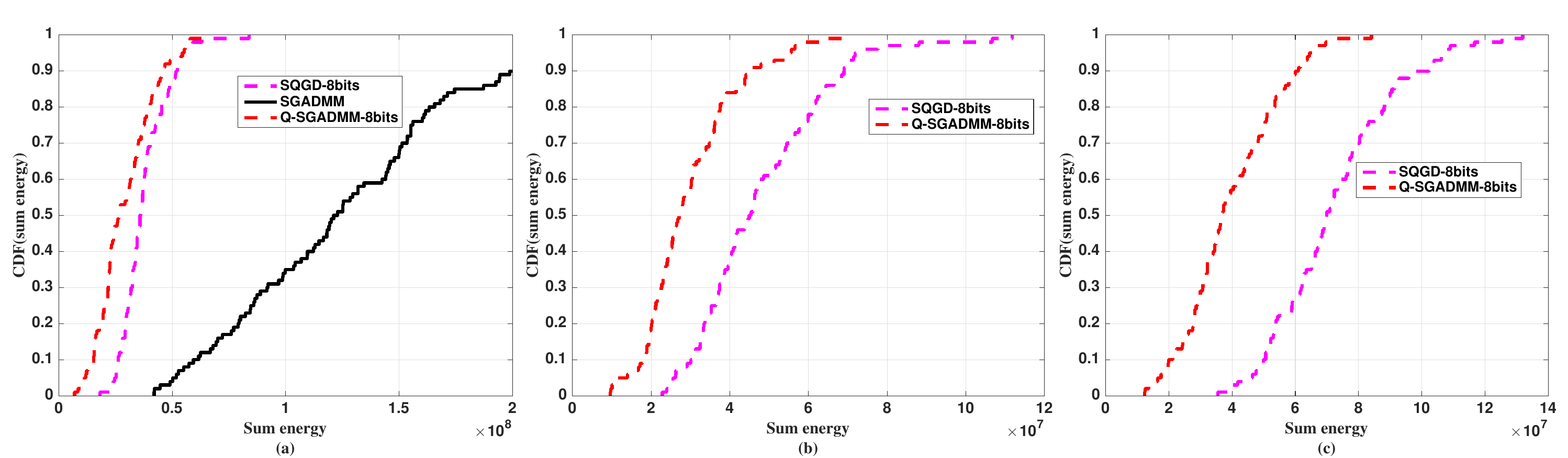}
\caption{\emph{Image classification} results showing: CDF of the consumed energy to achieve $90\%$ accuracy, (a) system bandwidth is 400MHz, system bandwidth is 100MHz, and (c) system bandwidth is 40MHz}
\label{fig:dnnCDF} \vskip -10pt
\end{figure*}  

\subsubsection{Simulation settings} To evaluate the performance of the stochastic version of Q-GADMM in stochastic and non-convex functions, we consider an image classification task using deep neural networks (DNN). Specifically, we consider the MNIST dataset comprising $60,000$ 28x28 pixel images that represent hand-written 0-9 digits. The dataset is divided into training ($70\%$) and testing ($30\%$) and then the training data is distributed uniformly across $10$ workers. Our DNN architecture is a multi-layer perceptron (MLP) that comprises three fully connected layers having $128$, $64$, and $10$ neurons, respectively. Note that the model size using this architecture is $109184$ which significantly larger than the model size of the linear regression model described above. Following the standard settings, we consider that the input image is flattened into a vector of size $784 \times 1$. The rectified linear unit (ReLu) activation function is used for each neuron, and the soft-max function is applied after the output layer. The loss function is cross-entropy defined as $-\sum_i y_i log(y^\prime_i)$, where $y_i$ is the vector label, and $y^\prime_i$ is the DNN output vector. We use Adam optimizer with a learning rate $0.001$ and ten iterations when solving the local problem at each worker. Moreover, we use a quantization resolution of $8$ bits ($256$ quantization levels). We assume that the transmission time is $100$ ms and the system bandwidth is $40MHz$. Finally, we choose $\rho=20$. 

Under the aforementioned deep NN architecture, the loss function is non-convex. In this case, it is well-known that stochastic sampling of training data mini-batches enables not only low computing complexity but also escaping saddle points. For this reason, we consider the stochastic variants of GADMM and Q-GADMM, coined stochastic GADMM (SGADMM) and quantized stochastic GADMM (Q-SGADMM), respectively, in which a mini-batch of $100$ samples are randomly selected per iteration. It is worth mentioning in order for both SGADMM and Q-SGADMM to converge for this non-convex problem, the updating step of the dual variable needs to be modified as ${\boldsymbol{\lambda}}_n^{k+1}={\boldsymbol{\lambda}}_n^{k} + \alpha\rho(\hat{{\boldsymbol{\theta}}}_{n}^{k+1} - \hat{{\boldsymbol{\theta}}}_{n+1}^{k+1})$, where $\alpha$ is a small constant, chosen to be equal to 0.01 in our simulation experiments. 

\subsubsection{Result discussion} Figs.~\ref{fig:dnn} and \ref{fig:dnnCDF}  show the results of the classification task using DNN. The results confirm the findings of the linear regression results. Namely, both Q-SGADMM and SGADMM have the same convergence speed and achieve the accuracy after the same number of iterations, outperforming both baselines (SGD and QSGD). However, as shown in Fig.~\ref{fig:dnn}-(b), Q-SGADMM-8bits achieves the accuracy of $90\%$ at one-fifth of the number of transmitted bits compared to full precision SGADMM. 



\subsection{Sensitivity Analysis}
\label{sensAnalysis}

In this section, we study the impact of the number of workers and the hyperparameters on the convergence speed and accuracy of Q-GADMM and QSGADMM. 

\begin{figure*}[t]
\centering
\includegraphics[width=0.95\textwidth]{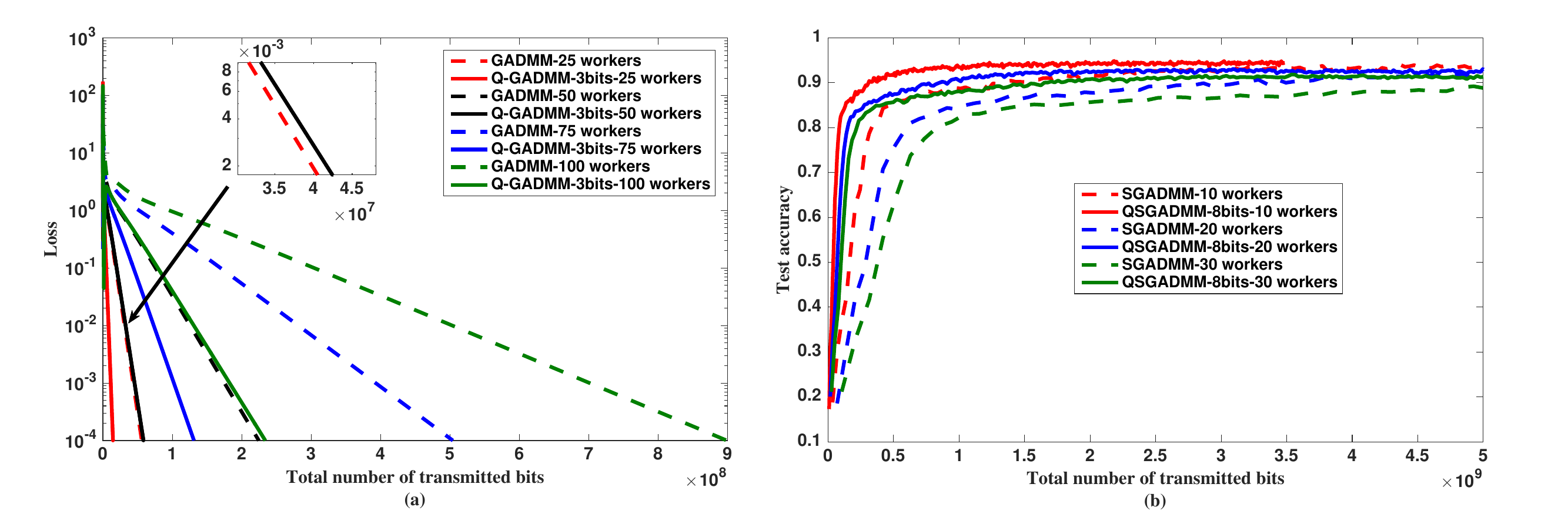}
\caption{Impact of the number of workers in (a) linear regression and (b) image classification using DNNs.}
\label{fig:noWorkersFig} \vskip -10pt
\end{figure*} 
\begin{figure*}[t]
\centering
\includegraphics[width=0.95\textwidth]{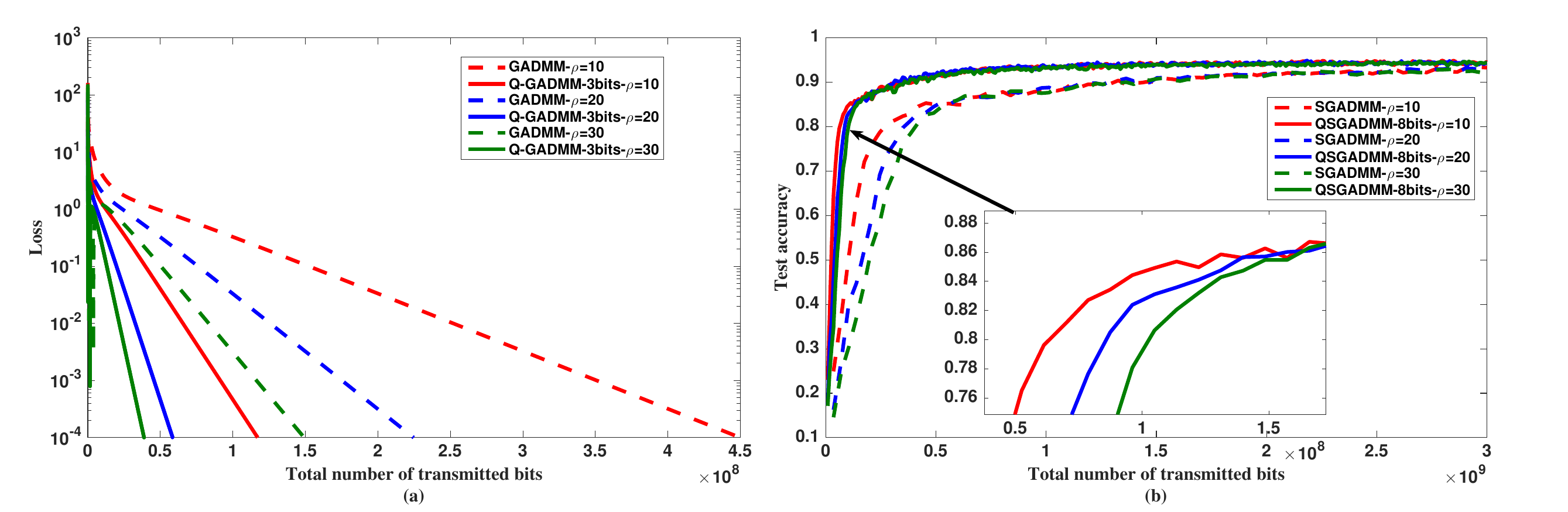}
\caption{Impact of the \emph{disagreement penalty weight $\rho$} in (a) linear regression and (b) image classification using DNNs.}
\label{rhoFig} \vskip -10pt
\end{figure*} 

\paragraph{Impact of the number of workers}
We conducted a number of linear regression and image classification experiments where for varying number of workers, we plot the loss/accuracy. In particular, Fig.~\ref{fig:noWorkersFig}-(a) plots the loss versus the total transmitted bits for linear regression problem under different choices of the number of workers. The figure reflects the scalability of Q-GADMM with respect to increasing number of workers. As we can see from the figure, the communication cost in terms of the total number of transmitted bits to achieve the loss value of $10^{-4}$ is increasing linearly and maintains the same $3.5$x less number of transmitted bits compared to GADMM. Finally, Fig.~\ref{fig:noWorkersFig}-(b) shows similar results and confirms the findings of the linear regression problem for the stochastic and non-convex problem of image classification using Q-SGADMM.

 \paragraph{Impact of $\rho$} The penalty weight $\rho$ adjusts the degree of disagreement between local and neighbor models in both linear regression and image classification tasks. In linear regression, Fig.~\ref{rhoFig}(a) shows that a larger $\rho$ leads to faster convergence for both Q-GADMM and GADMM. 
 
 In image classification, on the other hand, Fig.~\ref{rhoFig}(b) shows that a smaller $\rho$ reaches the highest test accuracy faster. For small~$\rho$, the disagreement penalty with other workers is not large. Therefore, every worker is likely to be biased towards its local optima. Hence, if the local optima is very close to the global one (the datasets of the workers are highly dependent), lower value of $\rho$ yields faster convergence. On the other hand, if the local optima is far away from the global one (the datasets are highly independent), large~$\rho$ can reduce the model disagreement and push all workers towards minimizing the disagreement in their model updates at every iteration which may lead to a faster convergence.

\subsection{Computation time}
\label{compTime}

\begin{figure*}[t]
\centering
\includegraphics[width=0.95\textwidth]{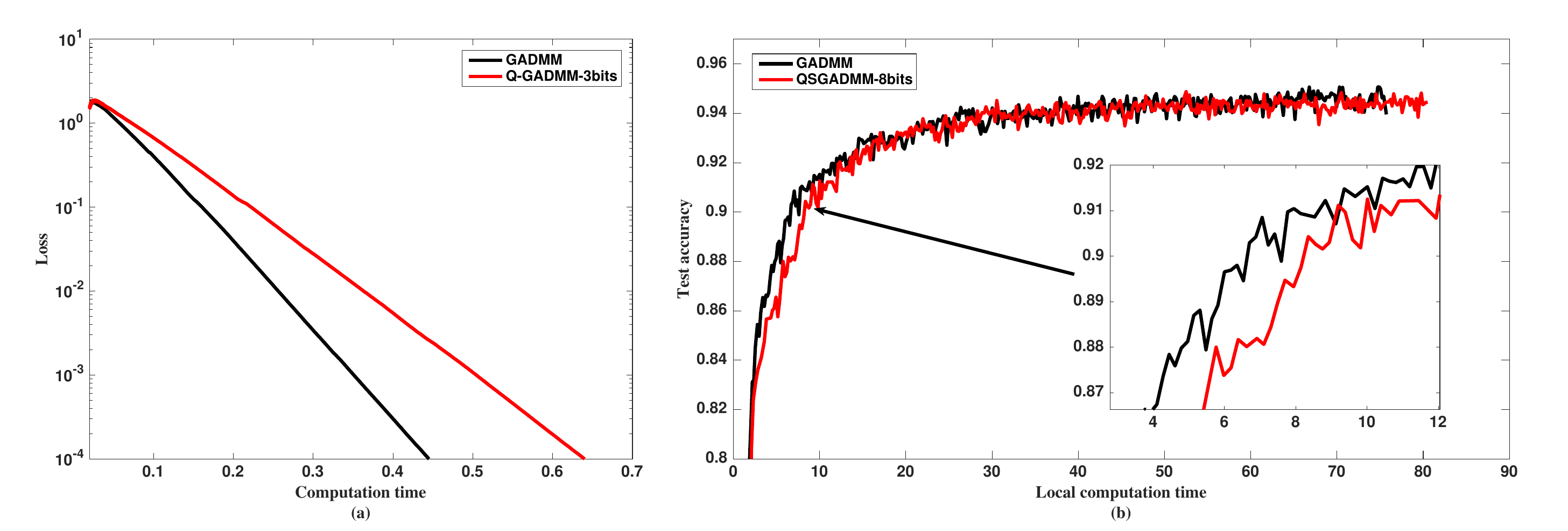}
\caption{The computation time (a) Q-GADMM vs. GADMM and (b) Q-SGADMM vs. SGADMM }
\label{fig:compTimeFig} \vskip -10pt
\end{figure*} 

Though our focus is on minimizing the communication energy given the scarce bandwidth, in this subsection, we show the extra cost encountered using Q-GADMM due to the quantization step. Fig.~\ref{fig:compTimeFig} plots the loss/accuracy versus the local computation time for both linear regression and image classification tasks; we ignore the communication time in this comparison. Since both GADMM and QGADMM and their stochastic versions encounter the same number of communication rounds to achieve the same loss/accuracy, the running time gap boils down to the per iteration consumed time. It is worth mentioning that linear regression task is performed using MATLAB R2015b and image classification task is performed using Python 3.2 and TensorFlow 2.0, and both run in a CPU based MacBook Air notebook. Fig.~\ref{fig:compTimeFig}-(a) shows that QGADMM requires $40\%$ more computation time to achieve the accuracy of $10^{-4}$ for the linear regression problem. However, for the image classification problem, even though the model size is significantly larger, Q-GADMM reduces the computation time gap compared to GADMM for high accuracy targets. That is due to the stochastic nature of the problem, introducing errors that can result in oscillation in the achieved accuracy.     

Finally, we conclude the evaluation section by mentioning that with as low as 2-bit quantizer resolution (4 quantization levels), Q-GADMM achieves the same accuracy and convergence speed of GADMM for convex loss functions, which confirms the statement of Theorem \ref{theorem}. For the stochastic and non-convex problem of image classification using a DNN, we observe that as we reduce the quantizer resolution, we see a very slight drop on the accuracy of Q-SGADMM compared to SGADMM. Theoretically studying the performance of Q-SGADMM for stochastic and non-convex problems could be an interesting topic for future work.

\section{Conclusion}
This article proposed a communication-efficient decentralized ML algorithm, Q-GADMM. Compared to the original GADMM with full precision, Q-GADMM enjoys the same convergence rate, but at significantly lower communication overhead. Utilizing stochastic quantization and re-defining the quantization range at every iteration are key features that make Q-GADMM robust to errors while ensuring its convergence. Numerical results in a convex linear regression task for Q-GADMM and stochastic non-convex image classification task for Q-SGADMM using deep learning corroborate the advantages of Q-GADMM over GADMM, GD, QGD, and A-DIANA. Theoretically studying the feasibility of Q-SGADMM for stochastic and non-convex functions, studying the impact of time-varying network topologies are interesting topics for future research.

\section{Appendices}
\subsection{Proof of Lemma \ref{lemma1}}
\label{sec:lem1}
Given that $f = \sum_nf_n(\boldsymbol{\theta}_n)$ is closed, proper, and convex, then the augmented Lagrangian $\boldsymbol{\mathcal{L}}_{\rho}$ is subdifferentiable. Since $\boldsymbol{\theta}_{n\in {\cal N}_h}^{k+1}$ minimizes $\boldsymbol{\mathcal{L}}_{\rho}( \boldsymbol{\theta}_{n\in {\cal N}_h},\boldsymbol{\theta}_{n\in {\cal N}_t}^{k}, \boldsymbol{\lambda}_{n}^{k})$, for each $n\in {\cal N}_h'$ at each iteration $k+1$, from \eqref{dual_res_1} and \eqref{cases}, it holds that 
\begin{align}\label{lemma_1_1}
\boldsymbol{0} \in \partial f_n(\boldsymbol{\theta}_{n}^{k+1}) - \boldsymbol{\lambda}_{n-1}^{k+1}+ \boldsymbol{\lambda}_{n}^{k+1} +2 \rho\boldsymbol{\epsilon}_n^{k+1}+ \boldsymbol{s}_{n}^{k+1}.
\end{align}
Without loss of generality, we use $\boldsymbol{\lambda}_{0}^{k+1} = \boldsymbol{\lambda}_{N}^{k+1} = 0 $ for the rest of the proof.
From \eqref{lemma_1_1}, we note that $\boldsymbol{\theta}_{n\in {\cal N}_h}^{k+1}$ minimizes the function
\begin{align}
f_n(\boldsymbol{\theta}_{n}) + \langle - \boldsymbol{\lambda}_{n-1}^{k+1}+ \boldsymbol{\lambda}_{n}^{k+1} +2 \rho \boldsymbol{\epsilon}_n^{k+1} + \boldsymbol{s}_{n \in {\cal N}_h}^{k+1}, \boldsymbol{\theta}_n \rangle.
\end{align}
Next, we substitute $\boldsymbol{\theta}_n=\boldsymbol{\theta}_n^{k+1}$ and $\boldsymbol{\theta}_n=\boldsymbol{\theta}_n^\star$ to obtain 
\small
\begin{align}
&\E{\left[f_n(\boldsymbol{\theta}_{n}^{k+1}) + \langle - \boldsymbol{\lambda}_{n-1}^{k+1} + \boldsymbol{\lambda}_{n}^{k+1} +2 \rho \boldsymbol{\epsilon}_n^{k+1} + \boldsymbol{s}_{n \in {\cal N}_h}^{k+1},\boldsymbol{\theta}_n^{k+1}\rangle \right]}\nonumber
\\
&\leq \E{\left[f_n(\boldsymbol{\theta}_n^*) + \langle - \boldsymbol{\lambda}_{n-1}^{k+1} + \boldsymbol{\lambda}_{n}^{k+1} + 2 \rho\boldsymbol{\epsilon}_n^{k+1} + \boldsymbol{s}_{n \in {\cal N}_h}^{k+1}, \boldsymbol{\theta}_n^*\rangle \right]}
\label{unbiased_prop}
\end{align}
\normalsize
for $n\in {\cal N}_h$. The second equality in \eqref{unbiased_prop} holds since the error introduced due to the stochastic quantization is zero mean. 
Similarly, for $n\in {\cal N}_t$, it holds that
\begin{align}
\nonumber &\E{\left[f_n(\boldsymbol{\theta}_{n}^{k+1}) + \langle - \boldsymbol{\lambda}_{n-1}^{k+1} + \boldsymbol{\lambda}_{n}^{k+1} + 2 \rho \boldsymbol{\epsilon}_n^{k+1}, \boldsymbol{\theta}_n^{k+1}\rangle\right]}\\
& \leq \E{\left[f_n(\boldsymbol{\theta}_n^*) + \langle - \boldsymbol{\lambda}_{n-1}^{k+1}+ \boldsymbol{\lambda}_{n}^{k+1}+2 \rho\boldsymbol{\epsilon}_n^{k+1}, \boldsymbol{\theta}_n^* \rangle \right]}.\label{unbiased_prop2} 
\end{align}
Adding \eqref{unbiased_prop} and \eqref{unbiased_prop2}, summing over all workers, and rearranging terms yields
\begin{align}
\nonumber &\sum_{n=1}^{N} \mathbb{E}[f_n(\boldsymbol{\theta}_{n}^{k+1})-f_n(\boldsymbol{\theta}_n^*)]
\\ \nonumber&\leq \E{}\left[\sum_{n=1}^N \langle - \boldsymbol{\lambda}_{n-1}^{k+1}+ \boldsymbol{\lambda}_{n}^{k+1}, \boldsymbol{\theta}_n^* -\boldsymbol{\theta}_n^{k+1} \rangle \right. \\
& \left. + \sum_{n\in {\cal N}_h}\langle \boldsymbol{s}_{n \in {\cal N}_h}^{k+1} , \boldsymbol{\theta}_n^*-\boldsymbol{\theta}_n^{k+1} \rangle - 2 \rho\sum_{n=1}^{N}  \langle \boldsymbol{\epsilon}_n^{k+1},\boldsymbol{\theta}_n^*- \boldsymbol{\theta}_n^{k+1} \rangle \right].\label{main_term}
\end{align}
Using the definition of $\boldsymbol{r}_{n,n+1}^{k+1}= \boldsymbol{\theta}_{n}^{k+1}-\boldsymbol{\theta}_{n+1}^{k+1}$, we can write
\begin{align}
\langle \boldsymbol{\lambda}_{n}^{k+1}, \boldsymbol{\theta}_n^* -\boldsymbol{\theta}_n^{k+1} \rangle =  \langle \boldsymbol{\lambda}_{n}^{k+1}, \boldsymbol{\theta}_n^* -\boldsymbol{\theta}_{n+1}^{k+1} \rangle - \langle \boldsymbol{\lambda}_{n}^{k+1}, \boldsymbol{r}_{n,n+1}^{k+1} \rangle.
\end{align}
There are three summation terms in the right hand side of \eqref{main_term}. Let us analyze each of them separately. Consider the following term
\begin{align}
&\sum_{n=1}^N \langle - \boldsymbol{\lambda}_{n-1}^{k+1}, \boldsymbol{\theta}_n^* - \boldsymbol{\theta}_n^{k+1} \rangle + \langle \boldsymbol{\lambda}_{n}^{k+1}, \boldsymbol{\theta}_n^* - \boldsymbol{\theta}_{n+1}^{k+1} \rangle\nonumber
\\
\nonumber &= - \langle \boldsymbol{\lambda}_{0}^{k+1}, \boldsymbol{\theta}_1^* - \boldsymbol{\theta}_1^{k+1} \rangle + \langle \boldsymbol{\lambda}_{1}^{k+1}, \boldsymbol{\theta}_1^* - \boldsymbol{\theta}_{2}^{k+1} \rangle \\
\nonumber &- \langle \boldsymbol{\lambda}_{1}^{k+1}, \boldsymbol{\theta}_2^* - \boldsymbol{\theta}_2^{k+1} \rangle + \langle \boldsymbol{\lambda}_{2}^{k+1}, \boldsymbol{\theta}_2^* - \boldsymbol{\theta}_{3}^{k+1} \rangle + \dots 
\\
& + \langle \boldsymbol{\lambda}_{N-1}^{k+1}, \boldsymbol{\theta}_{N-1}^* - \boldsymbol{\theta}_N^{k+1} \rangle - \langle \boldsymbol{\lambda}_{N-1}^{k+1}, \boldsymbol{\theta}_N^* - \boldsymbol{\theta}_{N}^{k+1} \rangle  {=} 0,\label{here}
\end{align}
where we have used $\boldsymbol{\theta}_n^*-\boldsymbol{\theta}_{n+1}^*=0$ for all $n$ and $\boldsymbol{\lambda}_0 = \boldsymbol{\lambda}_N = \boldsymbol{0}$ in \eqref{here}. Therefore, we can write \eqref{main_term} as 
\begin{align}
\nonumber &\sum_{n=1}^{N} \mathbb{E}[f_n(\boldsymbol{\theta}_{n}^{k+1})-f_n(\boldsymbol{\theta}_n^*)] 
\\ \nonumber & \leq \E\left[-\sum_{n=1}^{N-1} \langle \boldsymbol{\lambda}_n^{k+1}, \boldsymbol{r}_{n,n+1}^{k+1} \rangle 
  - 2 \rho \sum_{n=1}^{N} \langle \boldsymbol{\epsilon}_n^{k+1}, \boldsymbol{\theta}_n^*-\boldsymbol{\theta}_n^{k+1} \rangle \right.\\
& \left. + \sum_{n\in {\cal N}_h} \langle \boldsymbol{s}_{n \in {\cal N}_h}^{k+1}, \boldsymbol{\theta}_n^*-\boldsymbol{\theta}_n^{k+1} \rangle \right]
\label{eq2a}
\end{align}
which is the upper bound stated in Lemma \ref{lemma1}. For the lower bound, consider the definition of the augmented Lagrangian using $\boldsymbol{\theta}_{n+1}$ rather than $\hat{\boldsymbol{\theta}}_{n+1}$ as follows
\begin{align}
\boldsymbol{\mathcal{L}}_{\rho}\!=\!\!\sum_{n=1}^N \!f_n(\boldsymbol{\theta}_n)\!\! +\!\! \sum_{n=1}^{N-1}\!\! \ip{\boldsymbol{\lambda}_n,\boldsymbol{\theta}_{n} \!\!-\! \boldsymbol{\theta}_{n+1}}\!\!+\!\! \frac{\rho}{2}  \sum_{n=1}^{N-1}\! \| \boldsymbol{\theta}_{n} \!-\! \boldsymbol{\theta}_{n+1}\|^2.  
\label{augmentedLag}
\end{align}
Given that $(\boldsymbol{\theta}_n^*,\boldsymbol{\lambda}_n^*)$ is a saddle point for  $\boldsymbol{\mathcal{L}}_{0}(\boldsymbol{\theta}_n, \boldsymbol{\lambda}_n)$, it holds that
$\boldsymbol{\mathcal{L}}_{0}( \boldsymbol{\theta}_n^*, \boldsymbol{\lambda}_n^*) \leq \boldsymbol{\mathcal{L}}_{0}( \boldsymbol{\theta}_n^{k+1}, \boldsymbol{\lambda}_n^*)$.
Therefore, we can write
\begin{align}
\nonumber &\E{} \left[\sum_{n=1}^{N} f_n(\boldsymbol{\theta}_n^*)+ \sum_{n=1}^{N} \langle  \boldsymbol{\lambda}_n^*,\boldsymbol{\theta}_{n}^{*} - \boldsymbol{\theta}_{n+1}^{*}  \rangle \right] \\
& \leq \E{}\left[\sum_{n=1}^{N} f_n(\boldsymbol{\theta}_n^{k+1}) + \sum_{n=1}^{N-1}\langle \boldsymbol{\lambda}_n^*, \boldsymbol{\theta}_{n}^{k+1} - \boldsymbol{\theta}_{n+1}^{k+1} \rangle \right].
\label{eq3a}
\end{align}

Note that  $ \boldsymbol{\theta}_{n}^{*} - \boldsymbol{\theta}_{n+1}^{*} = 0$ for all $n$ and using $ \boldsymbol{r}_{n,n+1}^{k+1}= \boldsymbol{\theta}_{n}^{k+1} - \boldsymbol{\theta}_{n+1}^{k+1}$, we obtain can write  \eqref{eq3a} as 
\begin{align}
&\sum_{n=1}^N \E{[f_n(\boldsymbol{\theta}_n^{k+1})-f_n(\boldsymbol{\theta}_n^*)]} \geq -\sum_{n=1}^{N-1} \E{}[\langle \boldsymbol{\lambda}^*_n, \boldsymbol{r}_{n,n+1}^{k+1}\rangle]\label{last}
\end{align}
which concludes the proof of Lemma \ref{lemma1}.\hfill $\blacksquare$

\subsection{Proof of Theorem 1}
\label{sec:lem2}
To proceed with the analysis, we add \eqref{eq2a} and \eqref{last}, and then multiplying by 2 to get
\small
\begin{align}
\nonumber &\underbrace{2\sum_{n=1}^{N-1}\E{}\left[\langle \boldsymbol{\lambda}_n^{k+1}-\boldsymbol{\lambda}_n^*, \boldsymbol{r}_{n,n+1}^{k+1} \rangle \right]}_{(I)} 
 + \underbrace{2\sum_{n\in {\cal N}_h}\E{}\left[\boldsymbol{s}_{n \in {\cal N}_h}^{k+1}, \boldsymbol{\theta}_n^{k+1}-\boldsymbol{\theta}_n^* \rangle \right]}_{(II)} \\
 & + \underbrace{4 \rho \sum_{n=1}^N \E{}\left[ \langle \boldsymbol{\epsilon}_{n}^{k+1},\boldsymbol{\theta}_n^{k+1} -\boldsymbol{\theta}_n^*\rangle \right]}_{(III)} \leq 0.
\label{eq4a2}
\end{align}
\normalsize
There are three summation terms on the left hand side of \eqref{eq4a2}. We analyze each of them separately in order to obtain the final result as follows. From the dual update in \eqref{lambdaUpdate}, we have $\boldsymbol{\lambda}_n^{k+1}=\boldsymbol{\lambda}_n^{k}+\rho \boldsymbol{r}_{n,n+1}^{k+1}- \rho \boldsymbol{\epsilon}_n^{k+1} + \rho \boldsymbol{\epsilon}_{n+1}^{k+1}$. Therefore, the first term $(I)$ on the left hand side of \eqref{eq4a2}, can be re-written as
\begin{align}
\nonumber & I =\sum_{n=1}^{N-1} \left\{2 \E{}\left[\langle \boldsymbol{\lambda}_n^{k}-\boldsymbol{\lambda}_n^*, \boldsymbol{r}_{n,n+1}^{k+1} \rangle \right] + 2 \rho \E{}\|\boldsymbol{r}_{n,n+1}^{k+1}\|^2 \right. \\ \label{eq7a}
&\left. -2\rho\E{}\left[ \langle \boldsymbol{\epsilon}_n^{k+1}, \boldsymbol{r}_{n,n+1}^{k+1} \rangle \right] + 2 \rho\E{}\left[ \langle \boldsymbol{\epsilon}_{n+1}^{k+1}, \boldsymbol{r}_{n,n+1}^{k+1}\rangle \right] \right\}.
\end{align}

We rewrite \eqref{eq7a} as

\begin{align}
\nonumber & I =\sum_{n=1}^{N-1} \left\{2 \E{}\left[\langle \boldsymbol{\lambda}_n^{k}-\boldsymbol{\lambda}_n^*, \boldsymbol{r}_{n,n+1}^{k+1} \rangle \right] + \rho \E{}\|\boldsymbol{r}_{n,n+1}^{k+1}\|^2 \right. \\ \label{eq7aa}
&\left. -2\rho\E{}\left[ \langle \boldsymbol{\epsilon}_n^{k+1}, \boldsymbol{r}_{n,n+1}^{k+1} \rangle \right] + 2 \rho\E{}\left[ \langle \boldsymbol{\epsilon}_{n+1}^{k+1}, \boldsymbol{r}_{n,n+1}^{k+1}\rangle \right] \right. \nonumber\\
&\left. +\rho \E{}\|\boldsymbol{r}_{n,n+1}^{k+1}\|^2\right\}
\end{align}

Replacing $\boldsymbol{r}_{n,n+1}^{k+1}$ with its expression $\frac{1}{\rho} (\boldsymbol{\lambda}_n^{k+1}-\boldsymbol{\lambda}_n^k) + \boldsymbol{\epsilon}_n^{k+1}- \boldsymbol{\epsilon}_{n+1}^{k+1}$ in the first four terms and re-arranging terms in Eq. (\ref{eq7aa}), we get

\begin{align}
\nonumber & I =\sum_{n=1}^{N-1} \left\{2 \E{}\left[\langle \boldsymbol{\lambda}_n^{k}-\boldsymbol{\lambda}_n^*, \frac{1}{\rho} (\boldsymbol{\lambda}_n^{k+1}-\boldsymbol{\lambda}_n^k) + \boldsymbol{\epsilon}_n^{k+1}- \boldsymbol{\epsilon}_{n+1}^{k+1} \rangle \right]\right.\\
&\left. + \rho \E{}\|\frac{1}{\rho} (\boldsymbol{\lambda}_n^{k+1}-\boldsymbol{\lambda}_n^k) + \boldsymbol{\epsilon}_n^{k+1}- \boldsymbol{\epsilon}_{n+1}^{k+1}\|^2 \right. \nonumber \\ \label{eq7aaa}
&\left. + 2 \rho\E{}\left[ \langle \boldsymbol{\epsilon}_{n+1}^{k+1}-\boldsymbol{\epsilon}_{n+1}^{k}, \frac{1}{\rho} (\boldsymbol{\lambda}_n^{k+1}-\boldsymbol{\lambda}_n^k) + \boldsymbol{\epsilon}_n^{k+1}- \boldsymbol{\epsilon}_{n+1}^{k+1}\rangle \right] \right. \nonumber\\
&\left. +\rho \E{}\|\boldsymbol{r}_{n,n+1}^{k+1}\|^2\right\}
\end{align}

Writing $\boldsymbol{\lambda}_n^{k+1}-\boldsymbol{\lambda}_n^k$ as $(\boldsymbol{\lambda}_n^{k+1}-\boldsymbol{\lambda}_n^*) - (\boldsymbol{\lambda}_n^{k}-\boldsymbol{\lambda}_n^*)$, expanding the square term, and re-arranging terms, we get

\begin{align}
I  & = \frac{2}{\rho} \E{}\left[ \langle \boldsymbol{\lambda}_n^{k}-\boldsymbol{\lambda}_n^*,\boldsymbol{\lambda}_n^{k+1}-\boldsymbol{\lambda}_n^* \rangle \right] - \frac{2}{\rho} \E{}\left[\|\boldsymbol{\lambda}_n^k - \boldsymbol{\lambda}_n^*\|^2\right]\nonumber \\
 \nonumber &
+ 2\E{}\left[ \langle \boldsymbol{\lambda}_n^{k}-\boldsymbol{\lambda}_n^*, \boldsymbol{\epsilon}_n^{k+1} - \boldsymbol{\epsilon}_{n+1}^{k+1}\rangle \right] 
 \\
\nonumber &  + \frac{1}{\rho} \E{} \left[\|\boldsymbol{\lambda}_n^{k+1} - \boldsymbol{\lambda}_n^*\|^2\right] + \frac{1}{\rho} \E{} \left[\|\boldsymbol{\lambda}_n^{k} - \boldsymbol{\lambda}_n^*\|^2\right]\\
 \nonumber & + 2\E{}\left[ \langle \boldsymbol{\lambda}_n^{k+1}-\boldsymbol{\lambda}_n^*, \boldsymbol{\epsilon}_n^{k+1} - \boldsymbol{\epsilon}_{n+1}^{k+1}\rangle \right]
\\
\nonumber &  + 2\E{}\left[ \langle \boldsymbol{\lambda}_n^{*}-\boldsymbol{\lambda}_n^{k}, \boldsymbol{\epsilon}_n^{k+1} - \boldsymbol{\epsilon}_{n+1}^{k+1}\rangle \right] \\
\nonumber & + \frac{2}{\rho} \E{}\left[ \langle \boldsymbol{\lambda}_n^{k+1}-\boldsymbol{\lambda}_n^*,\boldsymbol{\lambda}_n^{*}-\boldsymbol{\lambda}_n^{k} \rangle \right] \\
\nonumber &   + 2 \E{} \left[ \langle \boldsymbol{\epsilon}_{n+1}^{k+1} - \boldsymbol{\epsilon}_{n}^{k+1}, \boldsymbol{\lambda}_n^{k+1}-\boldsymbol{\lambda}_n^{*}\rangle \right]
 \\
   \nonumber  &+ 2 \E{} \left[ \langle \boldsymbol{\epsilon}_{n+1}^{k+1} - \boldsymbol{\epsilon}_{n}^{k+1}, \boldsymbol{\lambda}_n^{*}-\boldsymbol{\lambda}_n^{k}\rangle \right]\\
  & + \rho \E{} \left[\|\boldsymbol{\epsilon}_{n}^{k+1} - \boldsymbol{\epsilon}_{n+1}^{k+1} \|^2\right]- 2 \rho \E{} \left[\|\boldsymbol{\epsilon}_{n}^{k+1} - \boldsymbol{\epsilon}_{n+1}^{k+1} \|^2\right]\nonumber\\
  &+ \rho \E{}\left[\| \boldsymbol{r}_{n,n+1}^{k+1} \|^2 \right].
\label{eq6a}
\end{align}
After removing the terms that cancel each other, expanding the term $\|\boldsymbol{\epsilon}_{n}^{k+1} - \boldsymbol{\epsilon}_{n+1}^{k+1} \|^2$ and re-arranging the expressions in \eqref{eq7a}, we can write
\begin{align}
\nonumber &I= \frac{1}{\rho} \sum_{n=1}^{N-1} \E{}\left[\| \boldsymbol{\lambda}_n^{k+1}-\boldsymbol{\lambda}_n^* \|^2 \right] - \frac{1}{\rho} \sum_{n=1}^{N-1} \E{}\left[\| \boldsymbol{\lambda}_n^{k}-\boldsymbol{\lambda}_n^* \|^2\right] \\
\nonumber & + \rho \sum_{n=1}^{N-1}  \E{}\left[\| \boldsymbol{r}_{n,n+1}^{k+1} \|^2\right]
- \rho \sum_{n=1}^{N-1} \E{}\left[\| \boldsymbol{\epsilon}_{n}^{k+1} \|^2 \right] \\
& - \rho \sum_{n=1}^{N-1}\E{}\left[\| \boldsymbol{\epsilon}_{n+1}^{k+1} \|^2 \right] + 2 \rho \sum_{n=1}^{N-1} \E{}\left[\langle \boldsymbol{\epsilon}_{n}^{k+1}, \boldsymbol{\epsilon}_{n+1}^{k+1} \rangle \right].
\label{term1}
\end{align}
Using the fact that
\small
\begin{align}
\nonumber & -\sum_{n=1}^{N-1} \left\{ \E{} \left[\| \boldsymbol{\epsilon}_{n}^{k+1} \|^2 \right] + \E{} \left[ \| \boldsymbol{\epsilon}_{n+1}^{k+1} \|^2 \right] \right\}\\
&= - 2 \sum_{n=1}^{N-1} \E{} \left[\| \boldsymbol{\epsilon}_{n}^{k+1} \|^2 \right] - \E{} \left[\| \boldsymbol{\epsilon}_{N}^{k+1} \|^2 \right] + \E{} \left[\| \boldsymbol{\epsilon}_{1}^{k+1} \|^2 \right],
\end{align}
\normalsize
we can write \eqref{term1} as 
\begin{align}
\nonumber &I = \frac{1}{\rho} \sum_{n=1}^{N-1} \E{}\left[\| \boldsymbol{\lambda}_n^{k+1}-\boldsymbol{\lambda}_n^* \|^2 \right] - \frac{1}{\rho} \sum_{n=1}^{N-1} \E{}\left[\| \boldsymbol{\lambda}_n^{k}-\boldsymbol{\lambda}_n^* \|^2\right]
\\
 \nonumber &+ \rho \sum_{n=1}^{N-1}  \E{}\left[\| \boldsymbol{r}_{n,n+1}^{k+1} \|^2\right] - 2 \rho \sum_{n=1}^{N-1}  \E{}\left[\| \boldsymbol{\epsilon}_{n}^{k+1} \|^2 \right]\\
 &+2 \rho \sum_{n=1}^{N-1} \E{}\left[\langle \boldsymbol{\epsilon}_{n}^{k+1}, \boldsymbol{\epsilon}_{n+1}^{k+1} \rangle \right]  - \E{} \left[\| \boldsymbol{\epsilon}_{N}^{k+1} \|^2 \right] + \E{} \left[\| \boldsymbol{\epsilon}_{1}^{k+1} \|^2 \right]. \label{term2}
\end{align}
Next, expanding the sum over nodes $\rho \sum\limits_{n=1}^{N-1} \E{}\left[\| \boldsymbol{r}_{n,n+1}^{k+1} \|^2\right]$ into the summation over $n\in{\cal N}_h'$ and $n\in{\cal N}_h $, we obtain
\begin{align}
\nonumber &I
= \frac{1}{\rho} \sum_{n=1}^{N-1} \E{}\left[\| \boldsymbol{\lambda}_n^{k+1}-\boldsymbol{\lambda}_n^* \|^2 \right] - \frac{1}{\rho} \sum_{n=1}^{N-1} \E{}\left[\| \boldsymbol{\lambda}_n^{k}-\boldsymbol{\lambda}_n^* \|^2\right]\\
\nonumber & + \rho \sum_{n\in{\cal N}_h' } \E{}\left[\| \boldsymbol{r}_{n-1,n}^{k+1} \|^2\right]+  \rho \sum_{n\in{\cal N}_h} \E{}\left[\| \boldsymbol{r}_{n,n+1}^{k+1} \|^2\right]
\\
\nonumber &- 2 \rho \sum_{n=1}^{N-1} \left\{ \E{}\left[\| \boldsymbol{\epsilon}_{n}^{k+1} \|^2 \right] - \E{}\left[\langle \boldsymbol{\epsilon}_{n}^{k+1}, \boldsymbol{\epsilon}_{n+1}^{k+1} \rangle \right] \right\} \\
& + \rho \E{} \left[\| \boldsymbol{\epsilon}_{1}^{k+1} \|^2 \right] - \rho \E{} \left[\| \boldsymbol{\epsilon}_{N}^{k+1} \|^2 \right]. 
\label{term2_final}
\end{align}
Next, we consider the second term on the left side of \eqref{eq4a2}. Since  $\boldsymbol{s}_{n}^{k+1} = \rho(\hat{\boldsymbol{\theta}}_{n-1}^{k+1}-\hat{\boldsymbol{\theta}}_{n-1}^{k})+\rho(\hat{\boldsymbol{\theta}}_{n+1}^{k+1}-\hat{\boldsymbol{\theta}}_{n+1}^{k})$ for $n\in {\cal N}_h'$, it holds that
\begin{align}
\nonumber & II=2 \rho \sum_{n\in {\cal N}_h' }\E{}\left[ \langle \hat{\boldsymbol{\theta}}_{n-1}^{k+1}-\hat{\boldsymbol{\theta}}_{n-1}^{k}, \boldsymbol{\theta}_{n}^{k+1}-\boldsymbol{\theta}_{n}^*\rangle \right]\\ &+2 \rho \sum_{n\in {\cal N}_h}\E{}\left[ \langle \hat{\boldsymbol{\theta}}_{n+1}^{k+1}-\hat{\boldsymbol{\theta}}_{n+1}^{k}, \boldsymbol{\theta}_{n}^{k+1}-\boldsymbol{\theta}_{n}^*\rangle \right].
\label{term2_eq1}
\end{align}
Note that for $n \in \{2, \dots, N-1\}$, we have  $\hat{\boldsymbol{\theta}}_{n-1}^{k}=\boldsymbol{\theta}_{n-1}^{k}-\boldsymbol{\epsilon}_{n-1}^{k}$ and $\hat{\boldsymbol{\theta}}_{n+1}^{k}=\boldsymbol{\theta}_{n+1}^{k}-\boldsymbol{\epsilon}_{n+1}^{k}$ for all $k$. Hence, \eqref{term2_eq1} becomes 
\small
\begin{align}
 \nonumber &II =2 \rho \sum_{n\in {\cal N}_h'} \E{}\left[\langle \boldsymbol{\theta}_{n-1}^{k+1}-\boldsymbol{\theta}_{n-1}^{k}-\boldsymbol{\epsilon}_{n-1}^{k+1}+\boldsymbol{\epsilon}_{n-1}^{k}, \boldsymbol{\theta}_{n}^{k+1}-\boldsymbol{\theta}_{n}^* \rangle \right]
\\
 &+ 2 \rho \sum_{n\in {\cal N}_h} \E{}\left[\langle \boldsymbol{\theta}_{n+1}^{k+1}-\boldsymbol{\theta}_{n+1}^{k}-\boldsymbol{\epsilon}_{n+1}^{k+1}+\boldsymbol{\epsilon}_{n+1}^{k}, \boldsymbol{\theta}_{n}^{k+1}-\boldsymbol{\theta}_{n}^* \rangle \right].
\label{term2_eq2}
\end{align}
\normalsize
Further, for $n \in \{2, \dots, N-1\}$, using  $\boldsymbol{\theta}_n^{k+1}=-\boldsymbol{r}_{n-1,n}^{k+1}+\boldsymbol{\theta}_{n-1}^{k+1} = \boldsymbol{r}_{n,n+1}^{k+1}+\boldsymbol{\theta}_{n+1}^{k+1}$ and $\boldsymbol{\theta}_n^*=\boldsymbol{\theta}_{n-1}^*=\boldsymbol{\theta}_{n+1}^*=\boldsymbol{\theta}^*$, and expanding the inner product terms,  we can rewrite \eqref{term2_eq2} as 
\begin{align}
\nonumber &II= 2 \rho \sum_{n\in {\cal N}_h'}\E{}\left[-\langle \boldsymbol{\theta}_{n-1}^{k+1}-\boldsymbol{\theta}_{n-1}^{k}, \boldsymbol{r}_{n-1,n}^{k+1} \rangle  \right. \\
\nonumber &+ \langle \boldsymbol{\epsilon}_{n-1}^{k+1}-\boldsymbol{\epsilon}_{n-1}^{k}, \boldsymbol{r}_{n-1,n}^{k+1} \rangle +  \langle \boldsymbol{\theta}_{n-1}^{k+1}-\boldsymbol{\theta}_{n-1}^{k},\boldsymbol{\theta}_{n-1}^{k+1}-\boldsymbol{\theta}^* \rangle
\\
\nonumber &
 \left. -\langle \boldsymbol{\epsilon}_{n-1}^{k+1}-\boldsymbol{\epsilon}_{n-1}^{k}, \boldsymbol{\theta}_{n-1}^{k+1}-\boldsymbol{\theta}^* \rangle \right] \\
\nonumber &+ 2 \rho \sum_{n\in {\cal N}_h} \E{} \left[ \langle \boldsymbol{\theta}_{n+1}^{k+1}-\boldsymbol{\theta}_{n+1}^{k}, \boldsymbol{r}_{n,n+1}^{k+1} \rangle \right. \\
\nonumber & -\langle \boldsymbol{\epsilon}_{n+1}^{k+1}-\boldsymbol{\epsilon}_{n+1}^{k}, \boldsymbol{r}_{n,n+1}^{k+1} \rangle + \langle \boldsymbol{\theta}_{n+1}^{k+1}-\boldsymbol{\theta}_{n+1}^{k}, \boldsymbol{\theta}_{n+1}^{k+1}-\boldsymbol{\theta}^* \rangle
 \nonumber
 \\
 &\left. - \langle \boldsymbol{\epsilon}_{n+1}^{k+1}-\boldsymbol{\epsilon}_{n+1}^{k}, \boldsymbol{\theta}_{n+1}^{k+1}-\boldsymbol{\theta}^* \rangle \right].
\label{term2_eq3}
\end{align}
Given the equalities
\begin{align}
 &\boldsymbol{\theta}_{n-1}^{k+1}-\boldsymbol{\theta}^* = (\boldsymbol{\theta}_{n-1}^{k+1}-\boldsymbol{\theta}_{n-1}^k)+(\boldsymbol{\theta}_{n-1}^{k}-\boldsymbol{\theta}^*), n\in {\cal N}_h',
 \nonumber\\&\boldsymbol{\theta}_{n+1}^{k+1}-\boldsymbol{\theta}^* = (\boldsymbol{\theta}_{n+1}^{k+1}-\boldsymbol{\theta}_{n+1}^k)+(\boldsymbol{\theta}_{n+1}^{k}-\boldsymbol{\theta}^*), n\in {\cal N}_h 
 \label{theta_eq}
\end{align}
we can rewrite the Eq. \eqref{term2_eq3} as
\begin{align}
\nonumber & II =2\rho \sum_{n\in {\cal N}_h'}\E{}\big[-\langle \boldsymbol{\theta}_{n-1}^{k+1}-\boldsymbol{\theta}_{n-1}^{k}, \boldsymbol{r}_{n-1,n}^{k+1} \rangle \\
\nonumber & + \langle \boldsymbol{\epsilon}_{n-1}^{k+1}-\boldsymbol{\epsilon}_{n-1}^{k}, \boldsymbol{r}_{n-1,n}^{k+1} \rangle + \|\boldsymbol{\theta}_{n-1}^{k+1}-\boldsymbol{\theta}_{n-1}^{k}\|^2
\\ \nonumber &+ \langle \boldsymbol{\theta}_{n-1}^{k+1}-\boldsymbol{\theta}_{n-1}^{k}, \boldsymbol{\theta}_{n-1}^{k}-\boldsymbol{\theta}^* \rangle \\
\nonumber & -\langle \boldsymbol{\epsilon}_{n-1}^{k+1}-\boldsymbol{\epsilon}_{n-1}^{k},\boldsymbol{\theta}_{n-1}^{k+1}-\boldsymbol{\theta}_{n-1}^k \rangle \\
\nonumber & -\langle \boldsymbol{\epsilon}_{n-1}^{k+1}-\boldsymbol{\epsilon}_{n-1}^{k}, \boldsymbol{\theta}_{n-1}^{k}-\boldsymbol{\theta}^* \rangle \big]
\\ \nonumber & + 2 \rho \sum_{n\in {\cal N}_h} \E{} \left[ \langle \boldsymbol{\theta}_{n+1}^{k+1}-\boldsymbol{\theta}_{n+1}^{k}, \boldsymbol{r}_{n,n+1}^{k+1} \rangle  \right.\\
\nonumber & 
-\langle \boldsymbol{\epsilon}_{n+1}^{k+1}-\boldsymbol{\epsilon}_{n+1}^{k}, \boldsymbol{r}_{n,n+1}^{k+1} \rangle
+\|\boldsymbol{\theta}_{n+1}^{k+1}-\boldsymbol{\theta}_{n+1}^{k}\|^2
\\ \nonumber &+ \langle \boldsymbol{\theta}_{n+1}^{k+1}-\boldsymbol{\theta}_{n+1}^{k}, \boldsymbol{\theta}_{n+1}^{k}-\boldsymbol{\theta}^* \rangle \\
\nonumber &  - \langle \boldsymbol{\epsilon}_{n+1}^{k+1}-\boldsymbol{\epsilon}_{n+1}^{k}, \boldsymbol{\theta}_{n+1}^{k+1}-\boldsymbol{\theta}_{n+1}^k \rangle \\
&- \left. \langle \boldsymbol{\epsilon}_{n+1}^{k+1}-\boldsymbol{\epsilon}_{n+1}^{k}, \boldsymbol{\theta}_{n+1}^{k}-\boldsymbol{\theta}^* \rangle\right]. \label{term2_eq4}
\end{align}
Further using the equalities,
\begin{align}
 &\boldsymbol{\theta}_{n-1}^{k+1}-\boldsymbol{\theta}_{n-1}^{k} = (\boldsymbol{\theta}_{n-1}^{k+1}-\boldsymbol{\theta}^*)-(\boldsymbol{\theta}_{n-1}^{k}-\boldsymbol{\theta}^*), n\in {\cal N}_h',
 \nonumber\\&\boldsymbol{\theta}_{n+1}^{k+1}-\boldsymbol{\theta}_{n+1}^k = (\boldsymbol{\theta}_{n+1}^{k+1}-\boldsymbol{\theta}^*)-(\boldsymbol{\theta}_{n+1}^{k}-\boldsymbol{\theta}^*), n\in {\cal N}_h,
 \label{theta_eq2}
\end{align}
we write \eqref{term2_eq4} as 
\begin{align}
\nonumber &II  =  2\rho \sum_{n\in {\cal N}_h' } \E{}\left[ \langle -\boldsymbol{\theta}_{n-1}^{k+1}-\boldsymbol{\theta}_{n-1}^{k}, \boldsymbol{r}_{n-1,n}^{k+1} \rangle \right. \\
\nonumber &+ \langle \boldsymbol{\epsilon}_{n-1}^{k+1}-\boldsymbol{\epsilon}_{n-1}^{k}, \boldsymbol{r}_{n-1,n}^{k+1} \rangle + \|\boldsymbol{\theta}_{n-1}^{k+1}-\boldsymbol{\theta}_{n-1}^{k}\|^2 \\
\nonumber &+\langle \boldsymbol{\theta}_{n-1}^{k+1}-\boldsymbol{\theta}^*, \boldsymbol{\theta}_{n-1}^{k}-\boldsymbol{\theta}^* \rangle -\|\boldsymbol{\theta}_{n-1}^{k}-\boldsymbol{\theta}^*\|^2 \\ \nonumber
&
-\langle \boldsymbol{\epsilon}_{n-1}^{k+1}-\boldsymbol{\epsilon}_{n-1}^{k}, \boldsymbol{\theta}_{n-1}^{k+1}-\boldsymbol{\theta}_{n-1}^k \rangle\\
\nonumber &-\langle \left. \boldsymbol{\epsilon}_{n-1}^{k+1}-\boldsymbol{\epsilon}_{n-1}^{k}, \boldsymbol{\theta}_{n-1}^{k}-\boldsymbol{\theta}^* \rangle \right]
\\ \nonumber
&
+\sum_{n\in {\cal N}_h} \E{} \left[ \langle \boldsymbol{\theta}_{n+1}^{k+1}-\boldsymbol{\theta}_{n+1}^{k}, \boldsymbol{r}_{n,n+1}^{k+1} \rangle \right. \\ \nonumber
&
- \langle \boldsymbol{\epsilon}_{n+1}^{k+1}-\boldsymbol{\epsilon}_{n+1}^{k}, \boldsymbol{r}_{n,n+1}^{k+1} \rangle + \|\boldsymbol{\theta}_{n+1}^{k+1}-\boldsymbol{\theta}_{n+1}^{k}\|^2 
\\ \nonumber &+ \langle \boldsymbol{\theta}_{n+1}^{k+1}-\boldsymbol{\theta}^*, \boldsymbol{\theta}_{n+1}^{k}-\boldsymbol{\theta}^* \rangle -\|\boldsymbol{\theta}_{n+1}^{k}-\boldsymbol{\theta}^*\|^2 \\ \nonumber
&
-\langle \boldsymbol{\epsilon}_{n+1}^{k+1}-\boldsymbol{\epsilon}_{n+1}^{k}, \boldsymbol{\theta}_{n+1}^{k+1}-\boldsymbol{\theta}_{n+1}^k \rangle \\
& \left. - \langle \boldsymbol{\epsilon}_{n+1}^{k+1}-\boldsymbol{\epsilon}_{n+1}^{k}, \boldsymbol{\theta}_{n+1}^{k}-\boldsymbol{\theta}^* \rangle \right].\label{term2_eq5}
\end{align}
Rearranging the terms, we can write
\begin{align}
\nonumber &II \!=\! \!\!\sum_{n\in {\cal N}_h'}\!\!\E{}\! \left [ -2\rho \langle \boldsymbol{\theta}_{n-1}^{k+1}\!-\!\boldsymbol{\theta}_{n-1}^{k}, \boldsymbol{r}_{n-1,n}^{k+1} \rangle \right. \\ \nonumber &
\!\!+\!\! 2\rho \langle \boldsymbol{\epsilon}_{n-1}^{k+1}\!-\!\boldsymbol{\epsilon}_{n-1}^{k}, \boldsymbol{r}_{n-1,n}^{k+1} \rangle\!\! +\!\! \rho \|\boldsymbol{\theta}_{n-1}^{k+1}-\boldsymbol{\theta}_{n-1}^{k}\|^2\!\! \\
\nonumber & +\!\! \rho \|(\boldsymbol{\theta}_{n-1}^{k+1}\!-\!\boldsymbol{\theta}^*)\!-\!(\boldsymbol{\theta}_{n-1}^{k}\!-\!\boldsymbol{\theta}^*)\|^2 
\\ \nonumber
&
+ 2\rho \langle \boldsymbol{\theta}_{n-1}^{k+1}\!-\!\boldsymbol{\theta}^*,\boldsymbol{\theta}_{n-1}^{k}\!-\!\boldsymbol{\theta}^* \rangle \!-\!2\rho \|\boldsymbol{\theta}_{n-1}^{k}\!-\!\boldsymbol{\theta}^*\|^2 \\ \nonumber & -2\rho\langle \boldsymbol{\epsilon}_{n-1}^{k+1}\!-\!\boldsymbol{\epsilon}_{n-1}^{k}, \boldsymbol{\theta}_{n-1}^{k+1}\!-\!\boldsymbol{\theta}_{n-1}^k \rangle \\ \nonumber & \left.
\!-\!2\rho \langle \boldsymbol{\epsilon}_{n-1}^{k+1}\!-\!\boldsymbol{\epsilon}_{n-1}^{k},\boldsymbol{\theta}_{n-1}^{k}\!-\!\boldsymbol{\theta}^* \rangle \right]
\\ \nonumber
&+ \sum_{n\in {\cal N}_h}\E{} \left[ 2\rho \langle \boldsymbol{\theta}_{n+1}^{k+1}-\boldsymbol{\theta}_{n+1}^{k}, \boldsymbol{r}_{n,n+1}^{k+1} \rangle \right. \\ \nonumber &
-2\rho \langle \boldsymbol{\epsilon}_{n+1}^{k+1}-\boldsymbol{\epsilon}_{n+1}^{k}, \boldsymbol{r}_{n,n+1}^{k+1} \rangle
+\rho\|\boldsymbol{\theta}_{n+1}^{k+1}-\boldsymbol{\theta}_{n+1}^{k}\|^2
\\ \nonumber
&+\rho\|(\boldsymbol{\theta}_{n+1}^{k+1}-\boldsymbol{\theta}^*)-(\boldsymbol{\theta}_{n+1}^{k}-\boldsymbol{\theta}^*)\|^2 \\ \nonumber &
+2\rho \langle \boldsymbol{\theta}_{n+1}^{k+1}-\boldsymbol{\theta}^*,\boldsymbol{\theta}_{n+1}^{k}-\boldsymbol{\theta}^* \rangle -2\rho\|\boldsymbol{\theta}_{n+1}^{k}-\boldsymbol{\theta}^*\|^2
\\ \nonumber
&-2\rho \langle \boldsymbol{\epsilon}_{n+1}^{k+1}-\boldsymbol{\epsilon}_{n+1}^{k}, \boldsymbol{\theta}_{n+1}^{k+1}-\boldsymbol{\theta}_{n+1}^k \rangle \\ &
 \left. -2\rho \langle \boldsymbol{\epsilon}_{n+1}^{k+1}-\boldsymbol{\epsilon}_{n+1}^{k}, \boldsymbol{\theta}_{n+1}^{k}-\boldsymbol{\theta}^* \rangle \right].
\label{term2_eq6}
\end{align}
Finally, expanding the square terms $\|(\boldsymbol{\theta}_{n-1}^{k+1}-\boldsymbol{\theta}^*)-(\boldsymbol{\theta}_{n-1}^{k}-\boldsymbol{\theta}^*)\|^2$ and $\|(\boldsymbol{\theta}_{n+1}^{k+1}-\boldsymbol{\theta}^*)-(\boldsymbol{\theta}_{n+1}^{k}-\boldsymbol{\theta}^*)\|^2$ in \eqref{term2_eq6}, we get
\begin{align}
\nonumber &II = \sum_{n\in {\cal N}_h'}\E{} \left [ -2\rho \langle \boldsymbol{\theta}_{n-1}^{k+1}-\boldsymbol{\theta}_{n-1}^{k}, \boldsymbol{r}_{n-1,n}^{k+1} \rangle \right. \\
\nonumber & + 2\rho \langle \boldsymbol{\epsilon}_{n-1}^{k+1}-\boldsymbol{\epsilon}_{n-1}^{k}, \boldsymbol{r}_{n-1,n}^{k+1} \rangle+ \rho \|\boldsymbol{\theta}_{n-1}^{k+1}-\boldsymbol{\theta}_{n-1}^{k}\|^2
 \\ \nonumber &
+\rho\big(\| \boldsymbol{\theta}_{n-1}^{k+1}-\boldsymbol{\theta}^* \|^2 - \| \boldsymbol{\theta}_{n-1}^{k}-\boldsymbol{\theta}^* \|^2\big)\\
\nonumber &-2\rho\langle \boldsymbol{\epsilon}_{n-1}^{k+1}-\boldsymbol{\epsilon}_{n-1}^{k}, \boldsymbol{\theta}_{n-1}^{k+1}-\boldsymbol{\theta}_{n-1}^k \rangle \\
\nonumber & \left.
-2\rho \langle \boldsymbol{\epsilon}_{n-1}^{k+1}-\boldsymbol{\epsilon}_{n-1}^{k},\boldsymbol{\theta}_{n-1}^{k}-\boldsymbol{\theta}^* \rangle \right]
\\
\nonumber &+ \sum_{n\in {\cal N}_h}\E{} \left[2\rho \langle \boldsymbol{\theta}_{n+1}^{k+1}-\boldsymbol{\theta}_{n+1}^{k}, \boldsymbol{r}_{n,n+1}^{k+1} \rangle \right. \\
\nonumber &-2\rho \langle \boldsymbol{\epsilon}_{n+1}^{k+1}-\boldsymbol{\epsilon}_{n+1}^{k}, \boldsymbol{r}_{n,n+1}^{k+1} \rangle
+\rho\|\boldsymbol{\theta}_{n+1}^{k+1}-\boldsymbol{\theta}_{n+1}^{k}\|^2
\\
\nonumber &+\rho\big(\| \boldsymbol{\theta}_{n+1}^{k+1}\!-\!\boldsymbol{\theta}^* \|^2\!-\!\| \boldsymbol{\theta}_{n+1}^{k}\!-\!\boldsymbol{\theta}^*\|^2\big) \\
\nonumber &-2\rho \langle \boldsymbol{\epsilon}_{n+1}^{k+1}-\boldsymbol{\epsilon}_{n+1}^{k}, \boldsymbol{\theta}_{n+1}^{k+1}-\boldsymbol{\theta}_{n+1}^k \rangle \\
 &\left. -2\rho \langle \boldsymbol{\epsilon}_{n+1}^{k+1}-\boldsymbol{\epsilon}_{n+1}^{k}, \boldsymbol{\theta}_{n+1}^{k}-\boldsymbol{\theta}^* \rangle \right]. \label{term2_eq7}
\end{align}
Substituting \eqref{term2_final} and \eqref{term2_eq7} into \eqref{eq4a2} and using the following Lyapunov function
\small
\begin{align}
\nonumber & V^k = \frac{1}{\rho} \sum_{n=1}^{N-1} \mathbb{E}\left[\| \boldsymbol{\lambda}_n^{k}-\boldsymbol{\lambda}^* \|^2\right] + \rho \sum_{n\in{\cal N}_h'} \mathbb{E}\left[\|\boldsymbol{\theta}_{n-1}^{k}-\boldsymbol{\theta}^*\|^2\right]\\
& +  \rho \sum_{n\in{\cal N}_h} \mathbb{E}\left[\|\boldsymbol{\theta}_{n+1}^{k}-\boldsymbol{\theta}^*\|^2\right],
\end{align}
\normalsize
we obtain
\begin{align}
\nonumber &V_{k+1}- V_{k} - 2 \rho \sum_{n=1}^{N-1} \left\{ \E{}\left[\| \boldsymbol{\epsilon}_{n}^{k+1} \|^2 \right] - \E{}\left[\langle \boldsymbol{\epsilon}_{n}^{k+1}, \boldsymbol{\epsilon}_{n+1}^{k+1} \rangle \right] \right\}\\
\nonumber & +\sum_{n\in {\cal N}_h'}\E{} \left [\rho\| \boldsymbol{r}_{n-1,n}^{k+1} \|^2 -2\rho \langle \boldsymbol{\theta}_{n-1}^{k+1}-\boldsymbol{\theta}_{n-1}^{k}, \boldsymbol{r}_{n-1,n}^{k+1} \rangle \right.
\\
\nonumber &
+ 2\rho \langle \boldsymbol{\epsilon}_{n-1}^{k+1}-\boldsymbol{\epsilon}_{n-1}^{k}, \boldsymbol{r}_{n-1,n}^{k+1} \rangle + \rho \|\boldsymbol{\theta}_{n-1}^{k+1}-\boldsymbol{\theta}_{n-1}^{k}\|^2 \\
\nonumber & 
-2\rho\langle \boldsymbol{\epsilon}_{n-1}^{k+1}-\boldsymbol{\epsilon}_{n-1}^{k}, \boldsymbol{\theta}_{n-1}^{k+1}-\boldsymbol{\theta}_{n-1}^k \rangle \\
\nonumber & \left. -2\rho \langle \boldsymbol{\epsilon}_{n-1}^{k+1}-\boldsymbol{\epsilon}_{n-1}^{k},\boldsymbol{\theta}_{n-1}^{k}-\boldsymbol{\theta}^* \rangle \right]
 \\
\nonumber & + \sum_{n\in {\cal N}_h}\E{} \left [\rho\| \boldsymbol{r}_{n,n+1}^{k+1} \|^2 +2\rho \langle \boldsymbol{\theta}_{n+1}^{k+1}-\boldsymbol{\theta}_{n+1}^{k}, \boldsymbol{r}_{n,n+1}^{k+1} \rangle \right. \\
\nonumber &-2\rho \langle \boldsymbol{\epsilon}_{n+1}^{k+1}-\boldsymbol{\epsilon}_{n+1}^{k}, \boldsymbol{r}_{n,n+1}^{k+1} \rangle
+\rho\|\boldsymbol{\theta}_{n+1}^{k+1}-\boldsymbol{\theta}_{n+1}^{k}\|^2
\\
\nonumber &-2\rho \langle \boldsymbol{\epsilon}_{n+1}^{k+1}-\boldsymbol{\epsilon}_{n+1}^{k}, \boldsymbol{\theta}_{n+1}^{k+1}-\boldsymbol{\theta}_{n+1}^k \rangle\\
\nonumber & \left. -2\rho \langle \boldsymbol{\epsilon}_{n+1}^{k+1}-\boldsymbol{\epsilon}_{n+1}^{k}, \boldsymbol{\theta}_{n+1}^{k}-\boldsymbol{\theta}^* \rangle \right] - \rho \E{} \left[\| \boldsymbol{\epsilon}_{N}^{k+1} \|^2 \right] 
 \\
 &  + \rho \E{} \left[\| \boldsymbol{\epsilon}_{1}^{k+1} \|^2 \right]+ 4 \rho \sum_{n=1}^N \E{}\left[ \langle \boldsymbol{\epsilon}_{n}^{k+1},\boldsymbol{\theta}_n^{k+1}-\boldsymbol{\theta}_n^* \rangle \right] \leq 0.
\label{combined_terms}
\end{align}
Adding and subtracting the terms $\rho \| \boldsymbol{\epsilon}_{n-1}^{k+1}-\boldsymbol{\epsilon}_{n-1}^{k}\|^2$ and $\rho \| \boldsymbol{\epsilon}_{n+1}^{k+1}-\boldsymbol{\epsilon}_{n+1}^{k}\|^2$ to \eqref{combined_terms}, we obtain
\begin{align}
\nonumber &V_{k+1}\!-\! V_{k}
 \!-\! 2 \rho \sum_{n=1}^{N-1} \!\left\{ \E{}\left[\| \boldsymbol{\epsilon}_{n}^{k+1} \|^2 \right] \!-\! \E{}\left[\langle \boldsymbol{\epsilon}_{n}^{k+1}, \boldsymbol{\epsilon}_{n+1}^{k+1} \rangle \right] \right\}\!\\
\nonumber &+\!\!\sum_{n\in {\cal N}_h'}\!\!\E{}\!\left[\!\!  \right.
Z_{n-1,n}\!-\!\rho\| \boldsymbol{\epsilon}_{n-1}^{k+1}\!-\!\boldsymbol{\epsilon}_{n-1}^{k}\|^2\! \\
\nonumber & -\!2\rho \langle \boldsymbol{\epsilon}_{n-1}^{k+1}\!-\!\boldsymbol{\epsilon}_{n-1}^{k},\boldsymbol{\theta}_{n-1}^{k}-\boldsymbol{\theta}^* \rangle
 \left.\right]
\\ \nonumber
&+ \sum_{n\in {\cal N}_h}\E{}\left[Z_{n
+1,n} \right.
-\rho \| \boldsymbol{\epsilon}_{n+1}^{k+1}-\boldsymbol{\epsilon}_{n+1}^{k}\|^2 \\
\nonumber & -2\rho \langle \boldsymbol{\epsilon}_{n+1}^{k+1}-\boldsymbol{\epsilon}_{n+1}^{k}, \boldsymbol{\theta}_{n+1}^{k}-\boldsymbol{\theta}^* \rangle 
 \left. \right] - \rho \E{} \left[\| \boldsymbol{\epsilon}_{N}^{k+1} \|^2 \right] 
\nonumber
\\
& + \rho \E{} \left[\| \boldsymbol{\epsilon}_{1}^{k+1} \|^2 \right]  + 4 \rho \sum_{n=1}^N \E{}\left[ \langle \boldsymbol{\epsilon}_{n}^{k+1},\boldsymbol{\theta}_n^{k+1}-\boldsymbol{\theta}_n^* \rangle \right] \leq 0.
\label{combined_terms2}
\end{align}
where 
\small
\begin{align}
Z_{n-1,n} &\triangleq \E{}\left[ \rho\| (\boldsymbol{\theta}_{n-1}^{k+1}-\boldsymbol{\theta}_{n-1}^k)-(\boldsymbol{\epsilon}_{n-1}^{k+1}-\boldsymbol{\epsilon}_{n-1}^{k})-\boldsymbol{r}_{n-1,n}^{k+1} \|^2 \right],\\
Z_{n+1,n} &\triangleq\E{}\left[\rho\| (\boldsymbol{\theta}_{n+1}^{k+1}-\boldsymbol{\theta}_{n+1}^k)-(\boldsymbol{\epsilon}_{n+1}^{k+1}-\boldsymbol{\epsilon}_{n+1}^{k})+\boldsymbol{r}_{n,n+1}^{k+1} \|^2 \right].
\end{align}
\normalsize
Furthermore, since every tail worker is summed twice, i.e., when we sum over $n-1$ and $n+1$ for the head worker $n$, we can write the following identities
\begin{align}
\nonumber & -\rho\sum_{n\in {\cal N}_h'} \| \boldsymbol{\epsilon}_{n-1}^{k+1}-\boldsymbol{\epsilon}_{n-1}^{k}\|^2 -\rho \sum_{n\in {\cal N}_h} \| \boldsymbol{\epsilon}_{n+1}^{k+1}-\boldsymbol{\epsilon}_{n+1}^{k}\|^2 \\
& = - 2 \rho \sum_{n\in {\cal N}_t} \| \boldsymbol{\epsilon}_{n}^{k+1}-\boldsymbol{\epsilon}_{n}^{k}\|^2\label{iden}
\end{align}
and
\begin{align}\label{iden2}
\nonumber &-2\rho \sum_{n\in {\cal N}_h'} \langle \boldsymbol{\epsilon}_{n-1}^{k+1}-\boldsymbol{\epsilon}_{n-1}^{k}, \boldsymbol{\theta}_{n-1}^{k}-\boldsymbol{\theta}^* \rangle \\
\nonumber & - 2 \rho \sum_{n\in {\cal N}_h} \langle \boldsymbol{\epsilon}_{n+1}^{k+1}-\boldsymbol{\epsilon}_{n+1}^{k},\boldsymbol{\theta}_{n+1}^{k}-\boldsymbol{\theta}^* \rangle\\
& = -4 \rho \sum_{n\in {\cal N}_t} \langle \boldsymbol{\epsilon}_{n}^{k+1}-\boldsymbol{\epsilon}_{n}^{k},\boldsymbol{\theta}_{n}^{k}-\boldsymbol{\theta}^* \rangle.
\end{align}
Using the identities of \eqref{iden} and \eqref{iden2} into \eqref{combined_terms2}, we get 
\begin{align}
\nonumber & V_{k+1}- V_{k}
 - 2 \rho \sum_{n=1}^{N-1} \left\{ \E{}\left[\| \boldsymbol{\epsilon}_{n}^{k+1} \|^2 \right] - \E{}\left[\langle \boldsymbol{\epsilon}_{n}^{k+1}, \boldsymbol{\epsilon}_{n+1}^{k+1} \rangle \right] \right\} \\
 \nonumber &+\sum_{n\in {\cal N}_h}Z_{n-1,n}
+ \sum_{n\in {\cal N}_h}Z_{n+1,n}+ \rho \E{} \left[\| \boldsymbol{\epsilon}_{1}^{k+1} \|^2 \right]   
\\
\nonumber &- \rho \E{} \left[\| \boldsymbol{\epsilon}_{N}^{k+1} \|^2 \right] 
 + 4 \rho \sum_{n\in {\cal N}_h}\E{}\left[ \langle \boldsymbol{\epsilon}_{n}^{k+1},\boldsymbol{\theta}_n^{k+1}-\boldsymbol{\theta}_n^* \rangle \right]  \\
 \nonumber & + 4 \rho \sum_{n\in {\cal N}_t} \E{}\left[ \langle \boldsymbol{\epsilon}_{n}^{k+1},\boldsymbol{\theta}_n^{k+1}-\boldsymbol{\theta}_n^* \rangle \right] 
\\
\nonumber & -4 \rho \sum_{n\in {\cal N}_t} \E{} \left[ \langle \boldsymbol{\epsilon}_{n}^{k+1}-\boldsymbol{\epsilon}_{n}^{k},\boldsymbol{\theta}_{n}^{k}-\boldsymbol{\theta}^* \rangle  \right] \\
& - 2 \rho \sum_{n\in {\cal N}_t} \E{} \left[\| \boldsymbol{\epsilon}_{n}^{k+1}-\boldsymbol{\epsilon}_{n}^{k}\|^2\right]   \leq 0.
\label{combined_terms3}
\end{align}
Using the equality $\boldsymbol{\theta}_n^{k+1}-\boldsymbol{\theta}^*=(\boldsymbol{\theta}_n^{k+1}-\boldsymbol{\theta}_n^{k})+(\boldsymbol{\theta}_n^{k}-\boldsymbol{\theta}^*)$ in the term $4\rho\sum_{n\in {\cal N}_t}\E{}\left[\langle \boldsymbol{\epsilon}_{n}^{k+1}, \boldsymbol{\theta}_n^{k+1}-\boldsymbol{\theta}^* \rangle \right]$ gives
\begin{align}
  \nonumber & V_{k+1}- V_{k}
 - 2 \rho \sum_{n=1}^{N-1} \left\{ \E{}\left[\| \boldsymbol{\epsilon}_{n}^{k+1} \|^2 \right] - \E{}\left[\langle \boldsymbol{\epsilon}_{n}^{k+1}, \boldsymbol{\epsilon}_{n+1}^{k+1} \rangle \right] \right\} \\
 \nonumber &+\!\!\sum_{n\in {\cal N}_h' }Z_{n-1,n}+ \sum_{n\in {\cal N}_h}Z_{n+1,n}- \rho \E{} \left[\| \boldsymbol{\epsilon}_{N}^{k+1} \|^2 \right] 
  \\
  \nonumber & + \rho \E{} \left[\| \boldsymbol{\epsilon}_{1}^{k+1} \|^2 \right] + 4 \rho \sum_{n\in {\cal N}_h}\E{}\left[ \langle \boldsymbol{\epsilon}_{n}^{k+1},\boldsymbol{\theta}_n^{k+1}-\boldsymbol{\theta}_n^* \rangle \right] \\
 \nonumber &+ 4 \rho \sum_{n\in {\cal N}_t} \E{}\left[ \langle \boldsymbol{\epsilon}_{n}^{k+1},\boldsymbol{\theta}_n^{k+1}-\boldsymbol{\theta}_n^k \rangle \right] 
 \\
  \nonumber &+ 4 \rho \sum_{n\in {\cal N}_t} \E{} \left[\langle \boldsymbol{\epsilon}_{n}^{k},\boldsymbol{\theta}_{n}^{k}-\boldsymbol{\theta}^* \rangle  \right] \\
 & - 2 \rho \sum_{n\in {\cal N}_t} \E{} \left[\| \boldsymbol{\epsilon}_{n}^{k+1}-\boldsymbol{\epsilon}_{n}^{k}\|^2\right]  \leq 0.
\label{combined_terms4}
\end{align}
Let us define a sigma algebra $\mathcal{F}_k$ that collect the algorithm history until instant $k$, which implies that $\mathbb{E}[\hat{\boldsymbol{\theta}}_{n}^{k}~|~\mathcal{F}_k]={\boldsymbol{\theta}}_{n}^{k}$ or alternatively $\mathbb{E}[{\epsilon}_{n}^{k}~|~\mathcal{F}_k]=0$. Similarly, this implies that $\mathbb{E}[\hat{\boldsymbol{\theta}}_{n}^{k+1}~|~\mathcal{F}_{k+1}]={\boldsymbol{\theta}}_{n}^{k+1}$, $\E{}\left[\langle \boldsymbol{\epsilon}_{n}^{k}, \boldsymbol{\epsilon}_{n}^{k+1} \rangle ~|~\mathcal{F}_{k+1}\right]=\E{}\left[\langle \boldsymbol{\epsilon}_{n}^{k},\mathbb{E}[ \boldsymbol{\epsilon}_{n}^{k+1} ~|~\mathcal{F}_{k+1}] \rangle\right]=0$, $ \E{}\left[\langle \boldsymbol{\epsilon}_{n}^{k}, \boldsymbol{\theta}^* \rangle \right] =0$. 
Now, let us consider the following inner product terms from \eqref{combined_terms4}, 
\begin{align}
\nonumber &\E{}\left[ \langle \boldsymbol{\epsilon}_{n}^{k+1},\boldsymbol{\theta}_n^{k+1}-\boldsymbol{\theta}_n^* \rangle \right]\\
\nonumber &=\E{}\left[ \langle \boldsymbol{\epsilon}_{n}^{k+1},\boldsymbol{\theta}_n^{k+1}-\boldsymbol{\theta}_n^* \rangle~|~\mathcal{F}_{k+1} \right]\\
&= \langle \E{}[\boldsymbol{\epsilon}_{n}^{k+1}~|~\mathcal{F}_{k+1}],\boldsymbol{\theta}_n^{k+1}-\boldsymbol{\theta}_n^* \rangle=0,\label{one}
\end{align}
Using a similar argument, we can write
\begin{align}
\E{}\left[ \langle \boldsymbol{\epsilon}_{n}^{k+1},\boldsymbol{\theta}_n^{k+1}-\boldsymbol{\theta}_n^k \rangle \right] &= 0, \label{two} \\
\E{} \left[\langle \boldsymbol{\epsilon}_{n}^{k},\boldsymbol{\theta}_{n}^{k}-\boldsymbol{\theta}^* \rangle  \right] &= 0. \label{three}
\end{align}
Using \eqref{one}-\eqref{three} into \eqref{combined_terms4}, we obtain
\begin{align}
\nonumber &V_{k+1}- V_k
- 2 \rho \sum_{n=1}^{N-1} \left\{ \E{}\left[\| \boldsymbol{\epsilon}_{n}^{k+1} \|^2 \right] - \E{}\left[\langle \boldsymbol{\epsilon}_{n}^{k+1}, \boldsymbol{\epsilon}_{n+1}^{k+1} \rangle \right] \right\}\\
\nonumber &+\sum_{n\in {\cal N}_h'}Z_{n-1,n}+ \sum_{n\in {\cal N}_h}Z_{n+1,n} - \rho \E{} \left[\| \boldsymbol{\epsilon}_{N}^{k+1} \|^2 \right] 
\\
\nonumber & + \rho \E{} \left[\| \boldsymbol{\epsilon}_{1}^{k+1} \|^2 \right] 
   + 4 \rho \sum_{n\in {\cal N}_h}\E{}\left[ \| \boldsymbol{\epsilon}_{n}^{k+1} \|^2 \right] \\
&+ 2 \rho \sum_{n\in {\cal N}_t} \left\{ \E{} \left[\| \boldsymbol{\epsilon}_{n}^{k+1} \|^2 \right] +  \E{} \left[\| \boldsymbol{\epsilon}_{n}^{k}\|^2\right] \right\}   \leq 0.
\label{combined_terms5}
\end{align}
Using the following inequality
\begin{align}
\nonumber &-2\rho \sum_{n=1}^{N-1}\E{}\left[\| \boldsymbol{\epsilon}_{n}^{k+1}\|^2\right] -2 \rho \E{}\left[\| \boldsymbol{\epsilon}_{N}^{k+1}\|^2\right]\\
&= -2\rho\sum_{n\in {\cal N}_h}\E{}\left[\| \boldsymbol{\epsilon}_{n}^{k+1}\|^2\right]-2\rho\sum_{n\in {\cal N}_t}\E{}\left[\| \boldsymbol{\epsilon}_{n}^{k+1}\|^2\right],
\end{align}
Eq. \eqref{combined_terms5} can be expressed as 
\begin{align}
\nonumber &V_{k+1}- V_k
+ 2 \rho \sum_{n=1}^{N-1}  \E{}\left[\langle \boldsymbol{\epsilon}_{n}^{k+1}, \boldsymbol{\epsilon}_{n+1}^{k+1} \rangle \right] +\sum_{n\in {\cal N}_h'}Z_{n-1,n} \\
&+ \sum_{n\in {\cal N}_h}Z_{n+1,n}+ \rho \E{} \left[\| \boldsymbol{\epsilon}_{N}^{k+1} \|^2 \right] 
 + \rho \E{} \left[\| \boldsymbol{\epsilon}_{1}^{k+1} \|^2 \right] 
\nonumber
\\
&+ 2 \rho \sum_{n\in {\cal N}_h}\E{}\left[ \| \boldsymbol{\epsilon}_{n}^{k+1} \|^2 \right] 
 + 2 \rho \sum_{n\in {\cal N}_t} \E{} \left[\| \boldsymbol{\epsilon}_{n}^{k}\|^2\right]    \leq 0.
\label{combined_terms66}
\end{align}
From the result in \eqref{combined_terms66}, it holds that 
\begin{align}
\nonumber &V^k - V^{k+1}  \\
\nonumber &\geq 2 \rho \sum_{n=1}^{N-1}  \E{}\left[\langle \boldsymbol{\epsilon}_{n}^{k+1}, \boldsymbol{\epsilon}_{n+1}^{k+1} \rangle \right]  + 
 \sum_{n\in {\cal N}_h'}Z_{n-1,n}\\
 \nonumber &+ \sum_{n\in {\cal N}_h}Z_{n+1,n}
 + \rho \E{} \left[\| \boldsymbol{\epsilon}_{1}^{k+1} \|^2 \right] 
  + \rho \E{} \left[\| \boldsymbol{\epsilon}_{N}^{k+1} \|^2 \right] 
\nonumber
  \\
  &+ 2 \rho \sum_{n\in {\cal N}_h}\E{}\left[ \| \boldsymbol{\epsilon}_{n}^{k+1} \|^2 \right] 
 + 2 \rho \sum_{n\in {\cal N}_t} \E{} \left[\| \boldsymbol{\epsilon}_{n}^{k}\|^2\right].
\label{combined_terms6}
\end{align}
Finally, using the following equality,
\begin{align}
\nonumber &2\rho\sum_{n=1}^{N-1}\E{}\left[\langle \boldsymbol{\epsilon}_{n}^{k+1}, \boldsymbol{\epsilon}_{n+1}^{k+1} \rangle\right]\\
\nonumber &=\rho \sum_{n=1}^{N-1}\left\{\E{}\left[\| \boldsymbol{\epsilon}_{n}^{k+1} + \boldsymbol{\epsilon}_{n+1}^{k+1} \|^2 \right] -\E{}\left[\|\boldsymbol{\epsilon}_{n}^{k+1}\|^2 \right]\right\} \\
&-\rho \sum_{n=1}^{N-1} \E{}\left[ \| \boldsymbol{\epsilon}_{n+1}^{k+1} \|^2\right] -2 \rho \sum_{n\in {\cal N}_t} \E{}\left[\| \boldsymbol{\epsilon}_{n}^{k+1}\|^2\right]
\nonumber
\\
&  + \rho \E{}\left[\| \boldsymbol{\epsilon}_{N}^{k+1}\|^2\right] + \rho \E{}\left[\| \boldsymbol{\epsilon}_{1}^{k+1}\|^2\right],
\label{combined_terms7}
\end{align}
we can rewrite \eqref{combined_terms6} as 
\begin{align}
\nonumber V^k - V^{k+1}  &\geq
\sum_{n\in {\cal N}_h'}Z_{n-1,n}+ \sum_{n\in {\cal N}_h}Z_{n+1,n} \\
&- 2 \rho \sum_{n\in {\cal N}_t} \E{} \left[\| \boldsymbol{\epsilon}_{n}^{k+1}\|^2\right].
\label{combined_terms8}
\end{align}
where we drop the positive terms $2 \rho \E{} \left[\| \boldsymbol{\epsilon}_{1}^{k+1} \|^2 \right] $, $2 \rho \E{} \left[\| \boldsymbol{\epsilon}_{N}^{k+1} \|^2 \right] $, $2 \rho \sum_{n\in {\cal N}_t}\E{}\left[ \| \boldsymbol{\epsilon}_{n}^{k} \|^2 \right] $, and \\$\rho \sum_{n=1}^{N-1} \E{}\left[\| \boldsymbol{\epsilon}_{n}^{k+1} + \boldsymbol{\epsilon}_{n+1}^{k+1}\|^2 \right]$ from the right hand side. 
Now, we have $\E{}\left[\|  \boldsymbol{\epsilon}_{n}^{k+1}\|^2\right] \leq d \left(\Delta_n^{k+1}\right)^2/4$, which implies that 
\begin{align}
&- 2\rho\sum_{n\in {\cal N}_t}\E{}\left[\| \boldsymbol{\epsilon}_{n}^{k+1}\|^2\right] \geq - \frac{d \rho}{2}  \sum_{n\in {\cal N}_t}{\left(\Delta_n^{k+1}\right)^2}.
\label{combined_terms9}
\end{align}
Hence, the inequality \eqref{combined_terms8} can be rewritten as
\begin{align}
\nonumber V^k - V^{k+1} &\geq \sum_{n\in {\cal N}_h'}Z_{n-1,n}+ \sum_{n\in {\cal N}_h}Z_{n+1,n}\\
& - \frac{d \rho}{2}  \sum_{n\in {\cal N}_t}{\left(\Delta_n^{k+1}\right)^2},
\label{combined_terms10}
\end{align}
\noindent which finalizes the proof. \hfill $\blacksquare$
\subsection{Proof of Theorem 2}
\label{sec:thm2}
\noindent From \eqref{lemma3_FinalEq}, we can write
\small
\begin{align}
\nonumber & \E{}\left[V^k - V^{k+1} \right] + \frac{d \rho}{2}  \sum_{n\in {\cal N}_t}{\left(\Delta_n^{k+1}\right)^2}\\
& \geq
 \sum_{n\in {\cal N}_h'}\E{}\left[ \rho\| \boldsymbol{\theta}_{n-1}^{k+1}-\boldsymbol{\theta}_{n-1}^k-(\boldsymbol{\epsilon}_{n-1}^{k+1}-\boldsymbol{\epsilon}_{n-1}^{k})-\boldsymbol{r}_{n-1,n}^{k+1} \|^2 \right]
\nonumber
\\
& + \sum_{n\in {\cal N}_h}\E{}\left[\rho\| \boldsymbol{\theta}_{n+1}^{k+1}-\boldsymbol{\theta}_{n+1}^k-(\boldsymbol{\epsilon}_{n+1}^{k+1}-\boldsymbol{\epsilon}_{n+1}^{k})+\boldsymbol{r}_{n,n+1}^{k+1} \|^2 \right].
\label{combined_terms11}
\end{align}
\normalsize
Taking the sum over $k=0, \dots, K$, and using the fact that $V^k$ is a positive decreasing sequence, we can write
\small
\begin{align}
\nonumber & \E{}\left[V^0\right] + \frac{d \rho}{2} \sum_{n\in {\cal N}_t} \sum_{k=0}^K {\left(\Delta_n^{k+1}\right)^2}\\
\nonumber &\geq
 \sum_{k=0}^K \sum_{n\in {\cal N}_h'}\E{}\left[ \rho\| \boldsymbol{\theta}_{n-1}^{k+1}-\boldsymbol{\theta}_{n-1}^k-(\boldsymbol{\epsilon}_{n-1}^{k+1}-\boldsymbol{\epsilon}_{n-1}^{k})-\boldsymbol{r}_{n-1,n}^{k+1} \|^2 \right]
\nonumber
\\
& + \sum_{k=0}^K \sum_{n\in {\cal N}_h}\E{}\left[\rho\| \boldsymbol{\theta}_{n+1}^{k+1}-\boldsymbol{\theta}_{n+1}^k-(\boldsymbol{\epsilon}_{n+1}^{k+1}-\boldsymbol{\epsilon}_{n+1}^{k})+\boldsymbol{r}_{n,n+1}^{k+1} \|^2 \right].
\end{align}
\normalsize
Taking the limit as $k \rightarrow \infty$, and using the assumption $\sum_{k=0}^{\infty}\Delta_n^{k} < \infty$, we conclude that the left hand side is convergent and as a consequence, we can write that
\small
\begin{align}
&\underset{k \rightarrow \infty}{\lim} \sum_{n\in {\cal N}_h'} \E{}\left[\| \boldsymbol{\theta}_{n-1}^{k+1}-\boldsymbol{\theta}_{n-1}^k-(\boldsymbol{\epsilon}_{n-1}^{k+1}-\boldsymbol{\epsilon}_{n-1}^{k})-\boldsymbol{r}_{n-1,n}^{k+1} \|^2 \right] = 0, \label{ff1} \\
&\underset{k \rightarrow \infty}{\lim} \sum_{n\in {\cal N}_h} \E{}\left[\| \boldsymbol{\theta}_{n+1}^{k+1}-\boldsymbol{\theta}_{n+1}^k-(\boldsymbol{\epsilon}_{n+1}^{k+1}-\boldsymbol{\epsilon}_{n+1}^{k})+\boldsymbol{r}_{n,n+1}^{k+1} \|^2 \right] = 0. \label{ff2}
\end{align}
\normalsize
Expanding the term in \eqref{ff1}, we can write
\small
\begin{align}
\nonumber &\E{}\left[\| \boldsymbol{\theta}_{n-1}^{k+1}-\boldsymbol{\theta}_{n-1}^k-(\boldsymbol{\epsilon}_{n-1}^{k+1}-\boldsymbol{\epsilon}_{n-1}^{k})-\boldsymbol{r}_{n-1,n}^{k+1} \|^2 \right]\\
\nonumber & = \E{}\left[\| \boldsymbol{\theta}_{n-1}^{k+1}-\boldsymbol{\theta}_{n-1}^k-\boldsymbol{r}_{n-1,n}^{k+1} \|^2 \right] + \E{}\left[\|\boldsymbol{\epsilon}_{n-1}^{k+1}-\boldsymbol{\epsilon}_{n-1}^{k}\|^2\right]\\
& - 2 \E{}\left[ \langle  \boldsymbol{\theta}_{n-1}^{k+1}-\boldsymbol{\theta}_{n-1}^k-\boldsymbol{r}_{n-1,n}^{k+1} ,\boldsymbol{\epsilon}_{n-1}^{k+1}-\boldsymbol{\epsilon}_{n-1}^{k}\rangle \right].\nonumber
\end{align}
\normalsize
Similarly, from \eqref{ff2}, we have
\small
\begin{align}
\nonumber &\E{}\left[\| \boldsymbol{\theta}_{n+1}^{k+1}-\boldsymbol{\theta}_{n+1}^k-(\boldsymbol{\epsilon}_{n+1}^{k+1}-\boldsymbol{\epsilon}_{n+1}^{k})+\boldsymbol{r}_{n,n+1}^{k+1} \|^2 \right]\\
\nonumber  & = \E{}\left[\| \boldsymbol{\theta}_{n+1}^{k+1}-\boldsymbol{\theta}_{n+1}^k+\boldsymbol{r}_{n,n+1}^{k+1} \|^2 \right] + \E{}\left[\|\boldsymbol{\epsilon}_{n+1}^{k+1}-\boldsymbol{\epsilon}_{n+1}^{k}\|^2\right] \\
&- 2 \E{}\left[ \langle  \boldsymbol{\theta}_{n+1}^{k+1}-\boldsymbol{\theta}_{n+1}^k+\boldsymbol{r}_{n,n+1}^{k+1} ,\boldsymbol{\epsilon}_{n+1}^{k+1}-\boldsymbol{\epsilon}_{n+1}^{k}\rangle \right].\nonumber
\end{align}
\normalsize
Using the following bounds
\small
\begin{align}
& \E{}\left[\|\boldsymbol{\epsilon}_{n-1}^{k+1}-\boldsymbol{\epsilon}_{n-1}^{k}\|^2\right] \leq 2 \left(\E{}\left[\|\boldsymbol{\epsilon}_{n-1}^{k+1}\|^2\right] + \E{}\left[\|\boldsymbol{\epsilon}_{n-1}^{k}\|^2\right]\right), \label{bn1}\\
\nonumber & \left|\E{}\left[ \langle  \boldsymbol{\theta}_{n-1}^{k+1}-\boldsymbol{\theta}_{n-1}^k-\boldsymbol{r}_{n-1,n}^{k+1}, \boldsymbol{\epsilon}_{n-1}^{k+1}-\boldsymbol{\epsilon}_{n-1}^{k}\rangle \right]\right| \\
&\leq \E{}\left[\|\boldsymbol{\theta}_{n-1}^{k+1}-\boldsymbol{\theta}_{n-1}^k-\boldsymbol{r}_{n-1,n}^{k+1}\|^2\right]^{\frac{1}{2}} \E{}\left[\|\boldsymbol{\epsilon}_{n-1}^{k+1}-\boldsymbol{\epsilon}_{n-1}^{k}\|^2\right]^{\frac{1}{2}} , \label{bn2}
\end{align}
\normalsize
where in the second bound, we have used Jensen and Cauchy-Schwarz inequalities. Since we have $E\left[\|\boldsymbol{\epsilon}_{n}^k\|^2\right]\leq d ({\Delta_n^k})^2/4$ and $\underset{k \rightarrow \infty}{\lim} \Delta_n^k = 0$, then $\underset{k \rightarrow \infty}{\lim} \E{}[\|\boldsymbol{\epsilon}_{n}^k\|^2] = 0$. Combining this with the bounds \eqref{bn1} and \eqref{bn2}, we obtain
\begin{align}
&\underset{k \rightarrow \infty}{\lim} \E{}\left[\|\boldsymbol{\epsilon}_{n-1}^{k+1}-\boldsymbol{\epsilon}_{n-1}^{k}\|^2\right] = 0, \label{qe3} \\
&\underset{k \rightarrow \infty}{\lim} \E{}\left[ \langle  \boldsymbol{\theta}_{n-1}^{k+1}-\boldsymbol{\theta}_{n-1}^k-\boldsymbol{r}_{n-1,n}^{k+1}, \boldsymbol{\epsilon}_{n-1}^{k+1}-\boldsymbol{\epsilon}_{n-1}^{k}\rangle \right]  = 0. 
\end{align}
As a consequence, we get from \eqref{ff1}, that
\begin{align}
\underset{k \rightarrow \infty}{\lim} \sum_{n\in {\cal N}_h \setminus\{1\}} \E{}\left[\| \boldsymbol{\theta}_{n-1}^{k+1}-\boldsymbol{\theta}_{n-1}^k-\boldsymbol{r}_{n-1,n}^{k+1} \|^2 \right] = 0.
\end{align}
Using similar arguments, we get
\begin{align}
\underset{k \rightarrow \infty}{\lim} \sum_{n\in {\cal N}_h} \E{}\left[\| \boldsymbol{\theta}_{n+1}^{k+1}-\boldsymbol{\theta}_{n+1}^k+\boldsymbol{r}_{n,n+1}^{k+1} \|^2 \right] = 0.
\end{align}
Therefore, we can write
\begin{align}\label{ch0}
\nonumber & \underset{k \rightarrow \infty}{\lim} \left\{\sum_{n\in {\cal N}_h \setminus\{1\}} \E{}\left[\| \boldsymbol{\theta}_{n-1}^{k+1}-\boldsymbol{\theta}_{n-1}^k-\boldsymbol{r}_{n-1,n}^{k+1} \|^2 \right] \right. \\
& \left. + \sum_{n\in {\cal N}_h} \E{}\left[\| \boldsymbol{\theta}_{n+1}^{k+1}-\boldsymbol{\theta}_{n+1}^k+\boldsymbol{r}_{n,n+1}^{k+1} \|^2 \right]\right\} = 0.
\end{align}
Expanding the terms inside the limit, we get
\begin{align}
\nonumber & \sum_{n\in {\cal N}_h'} \E{}\left[\| \boldsymbol{\theta}_{n-1}^{k+1}-\boldsymbol{\theta}_{n-1}^k-\boldsymbol{r}_{n-1,n}^{k+1} \|^2 \right] \\
\nonumber &  + \sum_{n\in {\cal N}_h} \E{}\left[\| \boldsymbol{\theta}_{n+1}^{k+1}-\boldsymbol{\theta}_{n+1}^k+\boldsymbol{r}_{n,n+1}^{k+1} \|^2 \right]\\
\nonumber &= \sum_{n\in {\cal N}_h'} \E{}\left[\| \boldsymbol{\theta}_{n-1}^{k+1}-\boldsymbol{\theta}_{n-1}^k\|^2 \right] + \sum_{n\in {\cal N}_h'} \E{}\left[\|\boldsymbol{r}_{n-1,n}^{k+1} \|^2 \right]\\
\nonumber &+ \sum_{n\in {\cal N}_h} \E{}\left[\| \boldsymbol{\theta}_{n+1}^{k+1}-\boldsymbol{\theta}_{n+1}^k\|^2 \right] + \sum_{n\in {\cal N}_h} \E{}\left[\|\boldsymbol{r}_{n,n+1}^{k+1} \|^2 \right] \\
\nonumber & -2 \sum_{n\in {\cal N}_h'} \E{}\left[\langle \boldsymbol{\theta}_{n-1}^{k+1}-\boldsymbol{\theta}_{n-1}^k, \boldsymbol{r}_{n-1,n}^{k+1} \rangle\right]\\
& + 2 \sum_{n\in {\cal N}_h} \E{}\left[\langle  \boldsymbol{\theta}_{n+1}^{k+1}-\boldsymbol{\theta}_{n+1}^k, \boldsymbol{r}_{n,n+1}^{k+1} \rangle\right]. \label{finale}
\end{align}
Since $\boldsymbol{\theta}_{n \in {\cal N}_t}^{k+1}$ minimizes $f_n(\boldsymbol{\theta}_{n}) + \langle -\boldsymbol{\lambda}_{n-1}^{k+1} +  \boldsymbol{\lambda}_{n}^{k+1} + 2 \rho \boldsymbol{\epsilon}_{n}^{k+1},  \boldsymbol{\theta}_{n}\rangle$,  then we can write
\begin{align}
\nonumber &f_n(\boldsymbol{\theta}_{n}^{k+1}) + \langle -\boldsymbol{\lambda}_{n-1}^{k+1} +  \boldsymbol{\lambda}_{n}^{k+1} + 2 \rho \boldsymbol{\epsilon}_{n}^{k+1},  \boldsymbol{\theta}_{n}^{k+1}\rangle \\
&\leq f_n(\boldsymbol{\theta}_{n}^{k}) + \langle -\boldsymbol{\lambda}_{n-1}^{k+1} +  \boldsymbol{\lambda}_{n}^{k+1} + 2 \rho \boldsymbol{\epsilon}_{n}^{k+1},  \boldsymbol{\theta}_{n}^{k}\rangle.
\label{eqch1}
\end{align} 
Similarly, $\boldsymbol{\theta}_{n \in {\cal N}_t}^k$ minimizes $f_n(\boldsymbol{\theta}_{n}) + \langle -\boldsymbol{\lambda}_{n-1}^{k} +  \boldsymbol{\lambda}_{n}^{k} + 2 \rho \boldsymbol{\epsilon}_{n}^{k},  \boldsymbol{\theta}_{n}\rangle$, then we can write
\begin{align}
\nonumber &f_n(\boldsymbol{\theta}_{n}^{k}) + \langle -\boldsymbol{\lambda}_{n-1}^{k} +  \boldsymbol{\lambda}_{n}^{k} + 2 \rho \boldsymbol{\epsilon}_{n}^{k},  \boldsymbol{\theta}_{n}^{k}\rangle \\
& \leq f_n(\boldsymbol{\theta}_{n}^{k+1}) + \langle -\boldsymbol{\lambda}_{n-1}^{k} +  \boldsymbol{\lambda}_{n}^{k} + 2 \rho \boldsymbol{\epsilon}_{n}^{k},  \boldsymbol{\theta}_{n}^{k+1}\rangle.
\label{eqch2}
\end{align} 
Adding \eqref{eqch1} and \eqref{eqch2} and using the fact that $\rho \boldsymbol{r}_{n,n+1}^{k+1} = \boldsymbol{\lambda}_{n}^{k+1} - \boldsymbol{\lambda}_{n}^{k} -\rho \boldsymbol{\epsilon}_{n}^{k+1} + \rho \boldsymbol{\epsilon}_{n+1}^{k+1}$ and $\rho \boldsymbol{r}_{n-1,n}^{k+1} = \boldsymbol{\lambda}_{n-1}^{k+1} - \boldsymbol{\lambda}_{n-1}^{k} -\rho \boldsymbol{\epsilon}_{n-1}^{k+1} + \rho \boldsymbol{\epsilon}_{n}^{k+1}$,  we get that $\forall n \in {\cal N}_t$
\small
\begin{align}
\rho \langle \boldsymbol{r}_{n-1,n}^{k+1} - \boldsymbol{r}_{n,n+1}^{k+1} + \boldsymbol{\epsilon}_{n+1}^{k+1} +\boldsymbol{\epsilon}_{n-1}^{k+1}-2\boldsymbol{\epsilon}_{n}^{k}, \boldsymbol{\theta}_{n}^{k+1} - \boldsymbol{\theta}_{n}^{k} \rangle \geq 0,
\end{align}
\normalsize
then, we can easily show that
\begin{align}\label{mb1}
\nonumber &\sum_{n\in {\cal N}_h} \E{}\left[\langle \boldsymbol{\theta}_{n+1}^{k+1}-\boldsymbol{\theta}_{n+1}^k, \boldsymbol{r}_{n,n+1}^{k+1} \rangle \right] \\
\nonumber & \geq \sum_{n\in {\cal N}_h' } \E{}\left[\langle \boldsymbol{\theta}_{n-1}^{k+1}-\boldsymbol{\theta}_{n-1}^k, \boldsymbol{r}_{n-1,n}^{k+1}\rangle\right]\\
\nonumber & - \sum_{n\in {\cal N}_t} \E{}\left[\langle \boldsymbol{\theta}_{n}^{k+1}-\boldsymbol{\theta}_{n}^k, \boldsymbol{\epsilon}_{n+1}^{k+1} + \boldsymbol{\epsilon}_{n-1}^{k+1} \rangle\right]\\
& + 2 \sum_{n\in {\cal N}_t} \E{}\left[\langle \boldsymbol{\theta}_{n}^{k+1}-\boldsymbol{\theta}_{n}^k, \boldsymbol{\epsilon}_{n}^{k} \rangle\right] .
\end{align}
We also have
\begin{align}\label{mb2}
\nonumber &\sum_{n\in {\cal N}_h'} \E{}\left[\|\boldsymbol{r}_{n-1,n}^{k+1} \|^2 \right] + \sum_{n\in {\cal N}_h} \E{}\left[\|\boldsymbol{r}_{n,n+1}^{k+1} \|^2 \right] \\
& = \sum_{n=1}^{N-1} \E{}\left[\|\boldsymbol{r}_{n,n+1}^{k+1} \|^2\right].
\end{align}
Then, from \eqref{mb1}, \eqref{mb2} and \eqref{finale}, we can write
\begin{align}
\nonumber &\sum_{n\in {\cal N}_h'} \E{}\left[\| \boldsymbol{\theta}_{n-1}^{k+1}-\boldsymbol{\theta}_{n-1}^k\|^2 \right] + \sum_{n\in {\cal N}_h} \E{}\left[\| \boldsymbol{\theta}_{n+1}^{k+1}-\boldsymbol{\theta}_{n+1}^k\|^2 \right]\\
\nonumber & + \sum_{n=1}^{N-1} \E{}\left[\|\boldsymbol{r}_{n,n+1}^{k+1} \|^2 \right] + 4 \sum_{n\in {\cal N}_t} \E{}\left[\langle \boldsymbol{\theta}_{n}^{k+1}-\boldsymbol{\theta}_{n}^k, \boldsymbol{\epsilon}_{n}^{k} \rangle\right] \\
\nonumber &\leq \sum_{n\in {\cal N}_h \setminus\{1\}} \E{}\left[\| \boldsymbol{\theta}_{n-1}^{k+1}-\boldsymbol{\theta}_{n-1}^k-\boldsymbol{r}_{n-1,n}^{k+1} \|^2 \right] \\
\nonumber & + \sum_{n\in {\cal N}_h} \E{}\left[\| \boldsymbol{\theta}_{n+1}^{k+1}-\boldsymbol{\theta}_{n+1}^k+\boldsymbol{r}_{n,n+1}^{k+1} \|^2 \right] \\
&+ 2 \sum_{n\in {\cal N}_t} \E{}\left[\langle \boldsymbol{\theta}_{n}^{k+1}-\boldsymbol{\theta}_{n}^k, \boldsymbol{\epsilon}_{n+1}^{k+1} + \boldsymbol{\epsilon}_{n-1}^{k+1} \rangle\right].
\end{align}
Using Jensen and Cauchy-Schwarz inequalities, we can write
\small
\begin{align}
\nonumber  &\left|\E{}\left[ \langle  \boldsymbol{\theta}_{n}^{k+1}-\boldsymbol{\theta}_{n}^k, \boldsymbol{\epsilon}_{n+1}^{k+1}+\boldsymbol{\epsilon}_{n-1}^{k+1}\rangle \right]\right|^2\\
& \leq 2 \E{}\left[\|\boldsymbol{\theta}_{n}^{k+1}-\boldsymbol{\theta}_{n}^k\|^2\right] \left(\E{}\left[\|\boldsymbol{\epsilon}_{n+1}^{k+1}\|^2\right] + \E{}\left[\|\boldsymbol{\epsilon}_{n-1}^{k+1}\|^2\right] \right), \\
&\left|\E{}\left[ \langle  \boldsymbol{\theta}_{n}^{k+1}-\boldsymbol{\theta}_{n}^k, \boldsymbol{\epsilon}_{n}^{k} \rangle \right]\right|^2 \leq \E{}\left[\|\boldsymbol{\theta}_{n}^{k+1}-\boldsymbol{\theta}_{n}^k\|^2\right]\E{}\left[\|\boldsymbol{\epsilon}_{n}^{k}\|^2\right]. 
\end{align}
\normalsize
Since we have $\underset{k \rightarrow \infty}{\lim} \E{}[\|\boldsymbol{\epsilon}_{n}^k\|^2] = 0$, then
\begin{align}
&\underset{k \rightarrow \infty}{\lim} \E{}\left[ \langle  \boldsymbol{\theta}_{n}^{k+1}-\boldsymbol{\theta}_{n}^k, \boldsymbol{\epsilon}_{n+1}^{k+1}+\boldsymbol{\epsilon}_{n-1}^{k+1}\rangle \right] = 0, \label{ag1} \\
&\underset{k \rightarrow \infty}{\lim} \E{}\left[ \langle  \boldsymbol{\theta}_{n}^{k+1}-\boldsymbol{\theta}_{n}^k, \boldsymbol{\epsilon}_{n}^{k} \rangle \right] = 0.\label{ag2}
\end{align}
Therefore, from \eqref{ch0}, \eqref{ag1}, and \eqref{ag2},  we obtain
\small
\begin{align}
\nonumber & \underset{k \rightarrow \infty}{\lim} \left\{ \sum_{n\in {\cal N}_h \setminus\{1\}} \E{}\left[\| \boldsymbol{\theta}_{n-1}^{k+1}-\boldsymbol{\theta}_{n-1}^k\|^2 \right] \right.
\\ & \left. + \sum_{n\in {\cal N}_h} \E{}\left[\| \boldsymbol{\theta}_{n+1}^{k+1}-\boldsymbol{\theta}_{n+1}^k\|^2 \right] + \sum_{n=1}^{N-1} \E{}\left[\|\boldsymbol{r}_{n,n+1}^{k+1} \|^2 \right] \right\} = 0.
\end{align}
\normalsize
As a consequence, we get that the primal residual converges to $\boldsymbol{0}$ in the mean square sense 
\begin{align}\label{pres}
\underset{k \rightarrow \infty}{\lim} \E{}\left[\|\boldsymbol{r}_{n,n+1}^{k+1} \|^2 \right] = 0.
\end{align}
Furthermore, we get
\begin{align}
&\underset{k \rightarrow \infty}{\lim} \E{}\left[\| \boldsymbol{\theta}_{n-1}^{k+1}-\boldsymbol{\theta}_{n-1}^k\|^2 \right] = 0, \label{qe1}\\
&\underset{k \rightarrow \infty}{\lim} \E{}\left[\| \boldsymbol{\theta}_{n+1}^{k+1}-\boldsymbol{\theta}_{n+1}^k\|^2 \right] = 0. \label{qe2}
\end{align}
Using the bound on the dual residual
\begin{align}\label{dres}
\nonumber & \E{}\left[\|\boldsymbol{S}_{n,n+1}^{k+1} \|^2\right] \\
\nonumber & \leq 4 \rho^2 \left(\E{}\left[\|\boldsymbol{\theta}_{n-1}^{k+1}-\boldsymbol{\theta}_{n-1}^k\|^2\right] + \E{}\left[\|\boldsymbol{\theta}_{n+1}^{k+1}-\boldsymbol{\theta}_{n+1}^k\|^2\right] \right.\\
& \left. + \E{}\left[\|\boldsymbol{\epsilon}_{n-1}^{k+1}-\boldsymbol{\epsilon}_{n-1}^{k}\|^2\right] + \E{}\left[\|\boldsymbol{\epsilon}_{n+1}^{k+1}-\boldsymbol{\epsilon}_{n+1}^{k}\|^2\right] \right).
\end{align}
Using Eqs. \eqref{qe3}, \eqref{qe1}, and \eqref{qe2}, we get that the dual residual converges to $\boldsymbol{0}$ in the mean square sense 
\begin{align}
\underset{k \rightarrow \infty}{\lim} \E{}\left[\|\boldsymbol{S}_{n,n+1}^{k+1} \|^2 \right] = 0.
\end{align}
Using the lower and upper bounds derived in Lemma \ref{lemma1} and using Jensen and Cauchy–Schwarz inequalities combined with \eqref{pres} and \eqref{dres}, we get that
\begin{align}
\lim_{k\rightarrow\infty} \E{}\left[\sum_{n=1}^N f_n(\boldsymbol{\theta}_n^{k})\right]= \E{}\left[\sum_{n=1}^Nf_n(\boldsymbol{\theta}^*)\right].
\end{align}
 \hfill $\blacksquare$


\bibliographystyle{IEEEtran}


\begin{thebibliography}{10}
	\providecommand{\url}[1]{#1}
	\csname url@samestyle\endcsname
	\providecommand{\newblock}{\relax}
	\providecommand{\bibinfo}[2]{#2}
	\providecommand{\BIBentrySTDinterwordspacing}{\spaceskip=0pt\relax}
	\providecommand{\BIBentryALTinterwordstretchfactor}{4}
	\providecommand{\BIBentryALTinterwordspacing}{\spaceskip=\fontdimen2\font plus
	\BIBentryALTinterwordstretchfactor\fontdimen3\font minus
	  \fontdimen4\font\relax}
	\providecommand{\BIBforeignlanguage}[2]{{%
	\expandafter\ifx\csname l@#1\endcsname\relax
	\typeout{** WARNING: IEEEtran.bst: No hyphenation pattern has been}%
	\typeout{** loaded for the language `#1'. Using the pattern for}%
	\typeout{** the default language instead.}%
	\else
	\language=\csname l@#1\endcsname
	\fi
	#2}}
	\providecommand{\BIBdecl}{\relax}
	\BIBdecl
	
	\bibitem{Google:FL19}
	P.~Kairouz, H.~B. McMahan, B.~Avent, A.~Bellet, M.~Bennis \emph{et~al.},
	  ``Advances and open problems in federated learning,'' \emph{arXiv preprint
	  arXiv:1912.04977}, 2019.
	
	\bibitem{park2020extreme}
	J.~Park, S.~Samarakoon, H.~Shiri, M.~K. Abdel-Aziz, T.~Nishio, A.~Elgabli, and
	  M.~Bennis, ``Extreme {URLLC}: Vision, challenges, and key enablers,''
	  \emph{arXiv preprint arXiv:2001.09683}, 2020.
	
	\bibitem{Girgis20:SPAWC}
	A.~M. Girgis, J.~Park, C.-F. Liu, and M.~Bennis, ``Predictive control and
	  communication co-design: A gaussian process regression approach,''
	  \emph{arXiv preprint arXiv: 2003.00243}, 2020.
	
	\bibitem{Shiri:CL20}
	H.~Shiri, J.~Park, and M.~Bennis, ``Remote uav online path planning via neural
	  network based opportunistic control,'' \emph{to appear in IEEE Communications
	  Letters}.
	
	\bibitem{park2018wireless}
	J.~Park, S.~Samarakoon, M.~Bennis, and M.~Debbah, ``Wireless network
	  intelligence at the edge,'' \emph{to appear in Proceedings of the IEEE
	  [Online]. Early access is available at:
	  https://ieeexplore.ieee.org/document/8865093}, November 2019.
	
	\bibitem{Park:2019:FLlet}
	J.~Park, S.~Wang, A.~Elgabli, S.~Oh, E.~Jeong, H.~Cha, H.~Kim, S.-L. Kim, and
	  M.~Bennis, ``Distilling on-device intelligence at the network edge,''
	  \emph{Arxiv preprint}, vol. abs/1908.05895, August 2019.
	
	\bibitem{park2020communicationefficient}
	J.~Park, S.~Samarakoon, A.~Elgabli, J.~Kim, M.~Bennis, S.-L. Kim, and
	  M.~Debbah, ``Communication-efficient and distributed learning over wireless
	  networks: Principles and applications,'' \emph{arXiv preprint
	  arXiv:2008.02608}, 2020.
	
	\bibitem{hosseinalipour2020multi}
	S.~Hosseinalipour, S.~S. Azam, C.~G. Brinton, N.~Michelusi, V.~Aggarwal, D.~J.
	  Love, and H.~Dai, ``Multi-stage hybrid federated learning over large-scale
	  wireless fog networks,'' \emph{arXiv preprint arXiv:2007.09511}, 2020.
	
	\bibitem{mcmahan2017federated}
	B.~McMahan and D.~Ramage, ``Federated learning: Collaborative machine learning
	  without centralized training data,'' \emph{Google Research Blog}, vol.~3,
	  2017.
	
	\bibitem{Smith:FLSurvey}
	T.~Li, A.~K. Sahu, A.~Talwalkar, and V.~Smith, ``Federated learning:
	  Challenges, methods, and future directions,'' \emph{IEEE Signal Processing
	  Magazine}, vol.~37, no.~3, pp. 50--60, 2020.
	
	\bibitem{Bernstein:2018aa}
	J.~Bernstein, Y.-X. Wang, K.~Azizzadenesheli, and A.~Anandkumar, ``{SignSGD}:
	  Compressed optimisation for non-convex problems,'' \emph{In Proc. Intl. Conf.
	  Machine Learn., Stockholm, Sweden}, July 2018.
	
	\bibitem{sun2019communication}
	J.~Sun, T.~Chen, G.~Giannakis, and Z.~Yang, ``Communication-efficient
	  distributed learning via lazily aggregated quantized gradients,'' in
	  \emph{Advances in Neural Information Processing Systems}, 2019, pp.
	  3370--3380.
	
	\bibitem{pap:jakub16}
	\BIBentryALTinterwordspacing
	J.~Konecny, H.~B. McMahan, F.~X. Yu, P.~Richtarik, A.~T. Suresh, and D.~Bacon,
	  ``Federated learning: strategies for improving communication efficiency,'' in
	  \emph{Proc. of NIPS Wksp. PMPML}, Barcelona, Spain, December 2016. [Online].
	  Available: \url{https://arxiv.org/abs/1610.05492}
	\BIBentrySTDinterwordspacing
	
	\bibitem{Jeong18}
	\BIBentryALTinterwordspacing
	E.~Jeong, S.~Oh, H.~Kim, J.~Park, M.~Bennis, and S.-L. Kim,
	  ``Communication-efficient on-device machine learning: Federated distillation
	  and augmentation under non-iid private data,'' \emph{presented at Neural
	  Information Processing Systems Workshop on Machine Learning on the Phone and
	  other Consumer Devices (MLPCD), Montr\'{e}al, Canada}, 2018. [Online].
	  Available: \url{http://arxiv.org/abs/1811.11479}
	\BIBentrySTDinterwordspacing
	
	\bibitem{Ahn:2019aa}
	J.-H. Ahn, O.~Simeone, and J.~Kang, ``Wireless federated distillation for
	  distributed edge learning with heterogeneous data,'' pp. 1--6, 2019.
	
	\bibitem{Wang:2019aa}
	S.~{Wang}, T.~{Tuor}, T.~{Salonidis}, K.~K. {Leung}, C.~{Makaya}, T.~{He}, and
	  K.~{Chan}, ``Adaptive federated learning in resource constrained edge
	  computing systems,'' \emph{IEEE Journal on Selected Areas in Communications},
	  vol.~37, no.~6, pp. 1205--1221, June 2019.
	
	\bibitem{chen2018lag}
	T.~Chen, G.~Giannakis, T.~Sun, and W.~Yin, ``Lag: Lazily aggregated gradient
	  for communication-efficient distributed learning,'' in \emph{Advances in
	  Neural Information Processing Systems}, 2018, pp. 5055--5065.
	
	\bibitem{FL_Nishio}
	\BIBentryALTinterwordspacing
	T.~Nishio and R.~Yonetani, ``Client selection for federated learning with
	  heterogeneous resources in mobile edge,'' \emph{In Proc. Int'l Conf. Commun.
	  (ICC), Shanghai, China}, May 2019. [Online]. Available:
	  \url{http://arxiv.org/abs/1804.08333}
	\BIBentrySTDinterwordspacing
	
	\bibitem{YangQuekPoor:2019aa}
	H.~H. Yang, Z.~Liu, T.~Q. Quek, and H.~V. Poor, ``Scheduling policies for
	  federated learning in wireless networks,'' \emph{IEEE Transactions on
	  Communications}, vol.~68, no.~1, pp. 317--333, 2019.
	
	\bibitem{Chen:20019aa}
	M.~Chen, Z.~Yang, W.~Saad, C.~Yin, H.~V. Poor, and S.~Cui, ``A joint learning
	  and communications framework for federated learning over wireless networks,''
	  \emph{arXiv preprint arXiv: 1909.07972}, 20019.
	
	\bibitem{Liu:2019aa}
	W.~Liu, L.~Chen, Y.~Chen, and W.~Zhang, ``Accelerating federated learning via
	  momentum gradient descent,'' \emph{IEEE Transactions on Parallel and
	  Distributed Systems}, vol.~31, no.~8, pp. 1754--1766, 2020.
	
	\bibitem{Yu:2019aa}
	H.~Yu, R.~Jin, and S.~Yang, ``On the linear speedup analysis of communication
	  efficient momentum {SGD} for distributed non-convex optimization,'' \emph{In
	  Proc. Intl. Conf. Machine Learn., Long Beach, CA, USA}, June 2019.
	
	\bibitem{elgabli2020gadmm}
	A.~Elgabli, J.~Park, A.~S. Bedi, M.~Bennis, and V.~Aggarwal, ``Gadmm: Fast and
	  communication efficient framework for distributed machine learning.''
	  \emph{Journal of Machine Learning Research}, vol.~21, no.~76, pp. 1--39,
	  2020.
	
	\bibitem{boyd2011distributed}
	S.~Boyd, N.~Parikh, E.~Chu, B.~Peleato, J.~Eckstein \emph{et~al.},
	  ``Distributed optimization and statistical learning via the alternating
	  direction method of multipliers,'' \emph{Foundations and
	  Trends{\textregistered} in Machine learning}, vol.~3, no.~1, pp. 1--122,
	  2011.
	
	\bibitem{li2020acceleration}
	Z.~Li, D.~Kovalev, X.~Qian, and P.~Richt{\'a}rik, ``Acceleration for compressed
	  gradient descent in distributed and federated optimization,'' \emph{arXiv
	  preprint arXiv:2002.11364}, 2020.
	
	\bibitem{elgabli2019qgadmm}
	A.~Elgabli, J.~Park, A.~S. Bedi, M.~Bennis, and V.~Aggarwal, ``Q-gadmm:
	  Quantized group admm for communication efficient decentralized machine
	  learning,'' in \emph{ICASSP 2020-2020 IEEE International Conference on
	  Acoustics, Speech and Signal Processing (ICASSP)}.\hskip 1em plus 0.5em minus
	  0.4em\relax IEEE, 2020, pp. 8876--8880.
	
	\bibitem{jakovetic2014fast}
	D.~Jakoveti{\'c}, J.~Xavier, and J.~M. Moura, ``Fast distributed gradient
	  methods,'' \emph{IEEE Transactions on Automation and Control Automa.
	  Control}, vol.~59, no.~5, pp. 1131--1146, 2014.
	
	\bibitem{nedic2014distributed}
	A.~Nedi{\'c} and A.~Olshevsky, ``Distributed optimization over time-varying
	  directed graphs,'' \emph{IEEE Trans. Automa. Control}, vol.~60, no.~3, pp.
	  601--615, 2014.
	
	\bibitem{nedic2009distributed}
	A.~Nedi\'{c} and A.~Ozdaglar, ``Distributed subgradient methods for multi-agent
	  optimization,'' \emph{IEEE Transactions on Automation and Control}, vol.~54,
	  no.~1, pp. 48--61, 2009.
	
	\bibitem{shi2015proximal}
	W.~Shi, Q.~Ling, G.~Wu, and W.~Yin, ``A proximal gradient algorithm for
	  decentralized composite optimization,'' \emph{IEEE Transactions on Signal
	  Processing}, vol.~63, no.~22, pp. 6013--6023, 2015.
	
	\bibitem{chang2014multi}
	T.-H. Chang, M.~Hong, and X.~Wang, ``Multi-agent distributed optimization via
	  inexact consensus admm,'' \emph{IEEE Transactions on Signal Processing},
	  vol.~63, no.~2, pp. 482--497, 2014.
	
	\bibitem{koppel2017proximity}
	A.~Koppel, B.~M. Sadler, and A.~Ribeiro, ``Proximity without consensus in
	  online multiagent optimization,'' \emph{IEEE Transactions on Signal
	  Processing}, vol.~65, no.~12, pp. 3062--3077, 2017.
	
	\bibitem{8624463}
	A.~S. Bedi, A.~Koppel, and R.~Ketan, ``Asynchronous saddle point algorithm for
	  stochastic optimization in heterogeneous networks,'' \emph{IEEE Transactions
	  on Signal Processing}, vol.~67, no.~7, pp. 1742--1757, 2019.
	
	\bibitem{glowinski1975approximation}
	R.~Glowinski and A.~Marroco, ``Sur l'approximation, par {\'e}l{\'e}ments finis
	  d'ordre un, et la r{\'e}solution, par p{\'e}nalisation-dualit{\'e} d'une
	  classe de probl{\`e}mes de dirichlet non lin{\'e}aires,'' \emph{ESAIM:
	  Mathematical Modelling and Numerical Analysis-Mod{\'e}lisation
	  Math{\'e}matique et Analyse Num{\'e}rique}, vol.~9, no.~R2, pp. 41--76, 1975.
	
	\bibitem{gabay1975dual}
	D.~Gabay and B.~Mercier, \emph{A dual algorithm for the solution of non linear
	  variational problems via finite element approximation}.\hskip 1em plus 0.5em
	  minus 0.4em\relax Institut de recherche d'informatique et d'automatique,
	  1975.
	
	\bibitem{jaggi2014communication}
	M.~Jaggi, V.~Smith, M.~Tak{\'a}c, J.~Terhorst, S.~Krishnan, T.~Hofmann, and
	  M.~I. Jordan, ``Communication-efficient distributed dual coordinate ascent,''
	  \emph{Advances in Neural Information Processing Systems}, vol.~27, pp.
	  3068--3076, 2014.
	
	\bibitem{ma2017distributed}
	C.~Ma, J.~Kone{\v{c}}n{\`y}, M.~Jaggi, V.~Smith, M.~I. Jordan,
	  P.~Richt{\'a}rik, and M.~Tak{\'a}{\v{c}}, ``Distributed optimization with
	  arbitrary local solvers,'' \emph{Optimization Methods and Software}, vol.~32,
	  no.~4, pp. 813--848, 2017.
	
	\bibitem{deng2017parallel}
	W.~Deng, M.-J. Lai, Z.~Peng, and W.~Yin, ``Parallel multi-block admm with
	  $o(1/k)$ convergence,'' \emph{Journal of Scientific Computing}, vol.~71,
	  no.~2, pp. 712--736, 2017.
	
	\bibitem{zhang2012communication}
	Y.~Zhang, M.~J. Wainwright, and J.~C. Duchi, ``Communication-efficient
	  algorithms for statistical optimization,'' in \emph{Advances in Neural
	  Information Processing Systems}, 2012, pp. 1502--1510.
	
	\bibitem{liu2019communication}
	Y.~Liu, W.~Xu, G.~Wu, Z.~Tian, and Q.~Ling, ``Communication-censored {ADMM} for
	  decentralized consensus optimization,'' \emph{IEEE Transactions on Signal
	  Processing}, vol.~67, no.~10, pp. 2565--2579, 2019.
	
	\bibitem{nandan2019}
	N.~Sriranga, C.~R. Murthy, and V.~Aggarwal, ``A method to improve consensus
	  averaging using quantized admm,'' in \emph{2019 IEEE International Symposium
	  on Information Theory (ISIT)}.\hskip 1em plus 0.5em minus 0.4em\relax IEEE,
	  2019.
	
	\bibitem{suresh2017distributed}
	A.~T. Suresh, F.~X. Yu, S.~Kumar, and H.~B. McMahan, ``Distributed mean
	  estimation with limited communication,'' in \emph{Proceedings of the 34th
	  International Conference on Machine Learning-Volume 70}.\hskip 1em plus 0.5em
	  minus 0.4em\relax JMLR. org, 2017, pp. 3329--3337.
	
	\bibitem{Zhu:2016aa}
	S.~Zhu, M.~Hong, and B.~Chen, ``Quantized consensus {ADMM} for multi-agent
	  distributed optimization,'' \emph{In Proceedings of International Conference
	  on Acoustics, Speech, and Signal Processing, Shanghai, China}, March 2016.
	
	\bibitem{Wang:2018aa}
	S.~Wang, T.~Tuor, T.~Salonidis, K.~K. Leung, C.~Makaya, T.~He, and K.~Chan,
	  ``Adaptive federated learning in resource constrained edge computing
	  systems,'' \emph{IEEE Journal on Selected Areas in Communications}, vol.~37,
	  no.~6, pp. 1205--1221, 2019.
	
	\bibitem{chen2016direct}
	C.~Chen, B.~He, Y.~Ye, and X.~Yuan, ``The direct extension of admm for
	  multi-block convex minimization problems is not necessarily convergent,''
	  \emph{Mathematical Programming}, vol. 155, no. 1-2, pp. 57--79, 2016.
	
	\bibitem{Nedic:09}
	A.~{Nedic} and A.~{Ozdaglar}, ``Distributed subgradient methods for multi-agent
	  optimization,'' \emph{IEEE Transactions on Automatic Control}, vol.~54,
	  no.~1, pp. 48--61, 2009.
	
	\bibitem{nedic2018network}
	A.~Nedi{\'c}, A.~Olshevsky, and M.~G. Rabbat, ``Network topology and
	  communication-computation tradeoffs in decentralized optimization,''
	  \emph{Proceedings of the IEEE}, vol. 106, no.~5, pp. 953--976, 2018.
	
	\bibitem{Lian:17}
	X.~Lian, C.~Zhang, H.~Zhang, C.-J. Hsieh, W.~Zhang, and J.~Liu, ``Can
	  decentralized algorithms outperform centralized algorithms? a case study for
	  decentralized parallel stochastic gradient descent,'' in \emph{Proc. of
	  NIPS}, Long Beach, CA, USA, December 2017.
	
	\bibitem{koloskova2019decentralized}
	A.~Koloskova, T.~Lin, S.~U. Stich, and M.~Jaggi, ``Decentralized deep learning
	  with arbitrary communication compression,'' \emph{arXiv preprint
	  arXiv:1907.09356}, 2019.
	
	\bibitem{gao2020adaptive}
	H.~Gao and H.~Huang, ``Adaptive serverless learning,'' \emph{arXiv preprint
	  arXiv:2008.10422}, 2020.
	
	\bibitem{Stich:18}
	S.~U. Stich, J.-B. Cordonnier, and M.~Jaggi, ``Sparsified sgd with memory,'' in
	  \emph{Advances in Neural Information Processing Systems}, 2018, pp.
	  4447--4458.
	
	\bibitem{chen2020communicationefficient}
	Y.~Chen, A.~Hashemi, and H.~Vikalo, ``Communication-efficient algorithms for
	  decentralized optimization over directed graphs,'' \emph{arXiv preprint
	  arXiv:2005.13189}, 2020.
	
	\bibitem{singh2019sparqsgd}
	N.~Singh, D.~Data, J.~George, and S.~Diggavi, ``Sparq-sgd: Event-triggered and
	  compressed communication in decentralized stochastic optimization,''
	  \emph{arXiv preprint arXiv:1910.14280}, 2019.
	
	\bibitem{xing2020decentralized}
	H.~Xing, O.~Simeone, and S.~Bi, ``Decentralized federated learning via sgd over
	  wireless d2d networks,'' \emph{arXiv preprint arXiv:2002.12507}, 2020.
	
	\bibitem{wen:2017_6749}
	W.~Wen, C.~Xu, F.~Yan, C.~Wu, Y.~Wang, Y.~Chen, and H.~Li, ``Terngrad: Ternary
	  gradients to reduce communication in distributed deep learning,'' in
	  \emph{Advances in neural information processing systems}, 2017, pp.
	  1509--1519.
	
	\bibitem{NIPS2017_Alistarh}
	D.~Alistarh, D.~Grubic, J.~Li, R.~Tomioka, and M.~Vojnovic, ``{QSGD}:
	  Communication-efficient {SGD} via gradient quantization and encoding,'' in
	  \emph{Proc. of Advances in NIPS 30}, Long Beach, CA, December 2017, pp.
	  1709--1720.
	
	\bibitem{Reisizadeh:2019aa}
	A.~Reisizadeh, A.~Mokhtari, H.~Hassani, A.~Jadbabaie, and R.~Pedarsani,
	  ``{FedPAQ}: A communication-efficient federated learning method with periodic
	  averaging and quantization,'' \emph{arXiv preprint arXiv: 1909.13014}, 2019.
	
	\bibitem{Horvath:19}
	S.~Horvath, C.-Y. Ho, L.~Horvath, A.~N. Sahu, M.~Canini, and P.~Richtarik,
	  ``Natural compression for distributed deep learning,'' \emph{arXiv preprint
	  arXiv: 1905.10988}.
	
	\bibitem{reisizadeh2019exact}
	A.~Reisizadeh, A.~Mokhtari, H.~Hassani, and R.~Pedarsani, ``An exact quantized
	  decentralized gradient descent algorithm,'' \emph{IEEE Transactions on Signal
	  Processing}, vol.~67, no.~19, pp. 4934--4947, 2019.
	
	\bibitem{Torgo:14}
	\BIBentryALTinterwordspacing
	L.~Torgo, ``Regression datasets,'' 2014. [Online]. Available:
	  \url{https://www.dcc.fc.up.pt/~ltorgo/Regression/DataSets.html}
	\BIBentrySTDinterwordspacing
	
	\end{thebibliography}

\begin{IEEEbiography}
	 [{\includegraphics[width=1in,height=1.25in,clip,keepaspectratio]{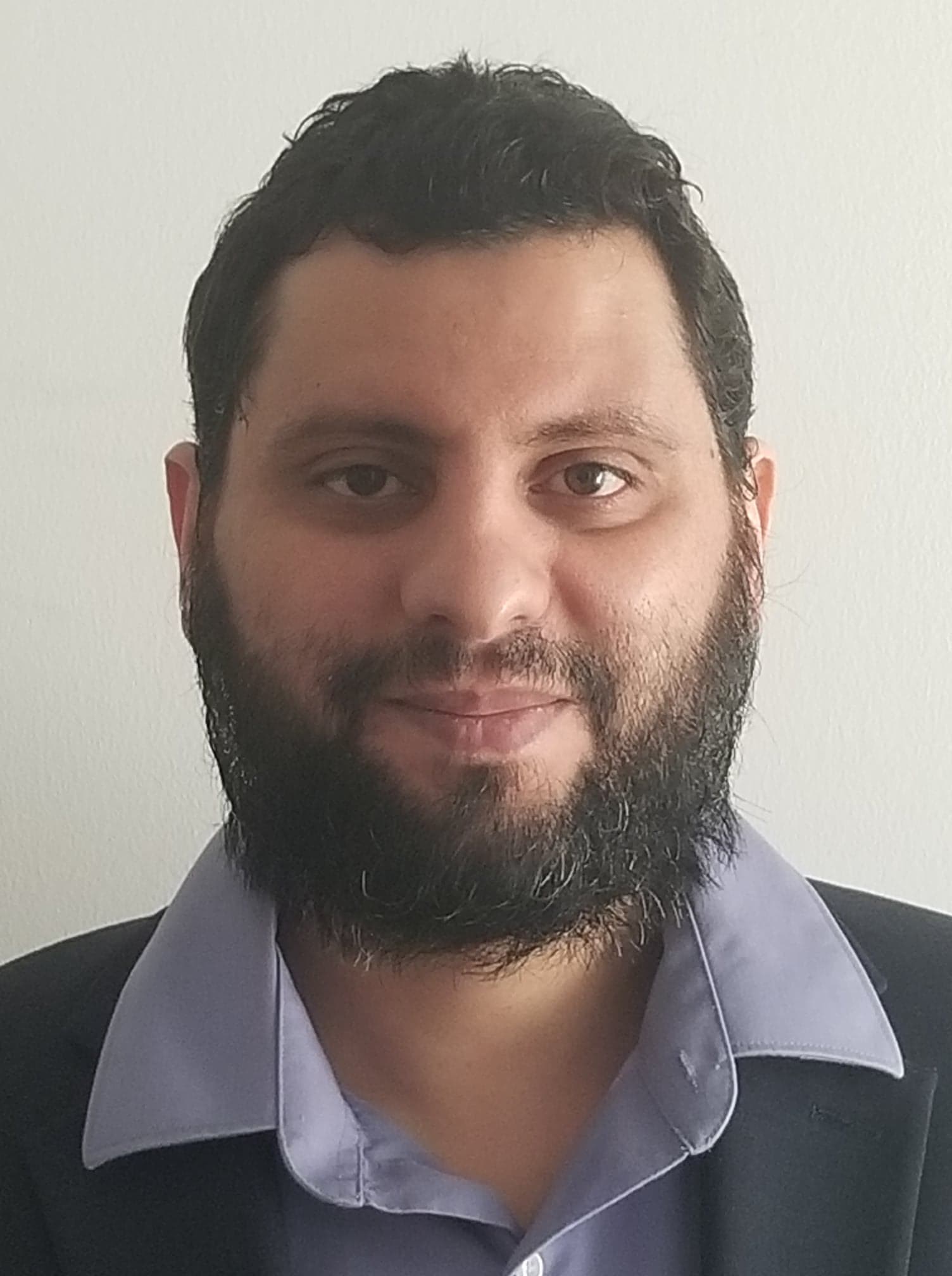}}]
	{Anis Elgabli} is a postdoctoral researcher at the Centre for Wireless Communications, University of Oulu. He received the B.Sc. degree in electrical and electronic engineering from the University of Tripoli, Libya, in 2004, the M.Eng. degree from UKM, Malaysia, in 2007, and MSc and PhD from the department of electrical and computer engineering, Purdue university, Indiana, USA  in 2015 and 2018 respectively. His main research interests are in heterogeneous networks, radio resource management, vehicular communication, video streaming, and distributed machine learning. He was the recipient of the best paper award in HotSpot workshop, 2018 (Infocom 2018).
\end{IEEEbiography}

\begin{IEEEbiography}
	[{\includegraphics[width=1in,height=1.25in,clip,keepaspectratio]{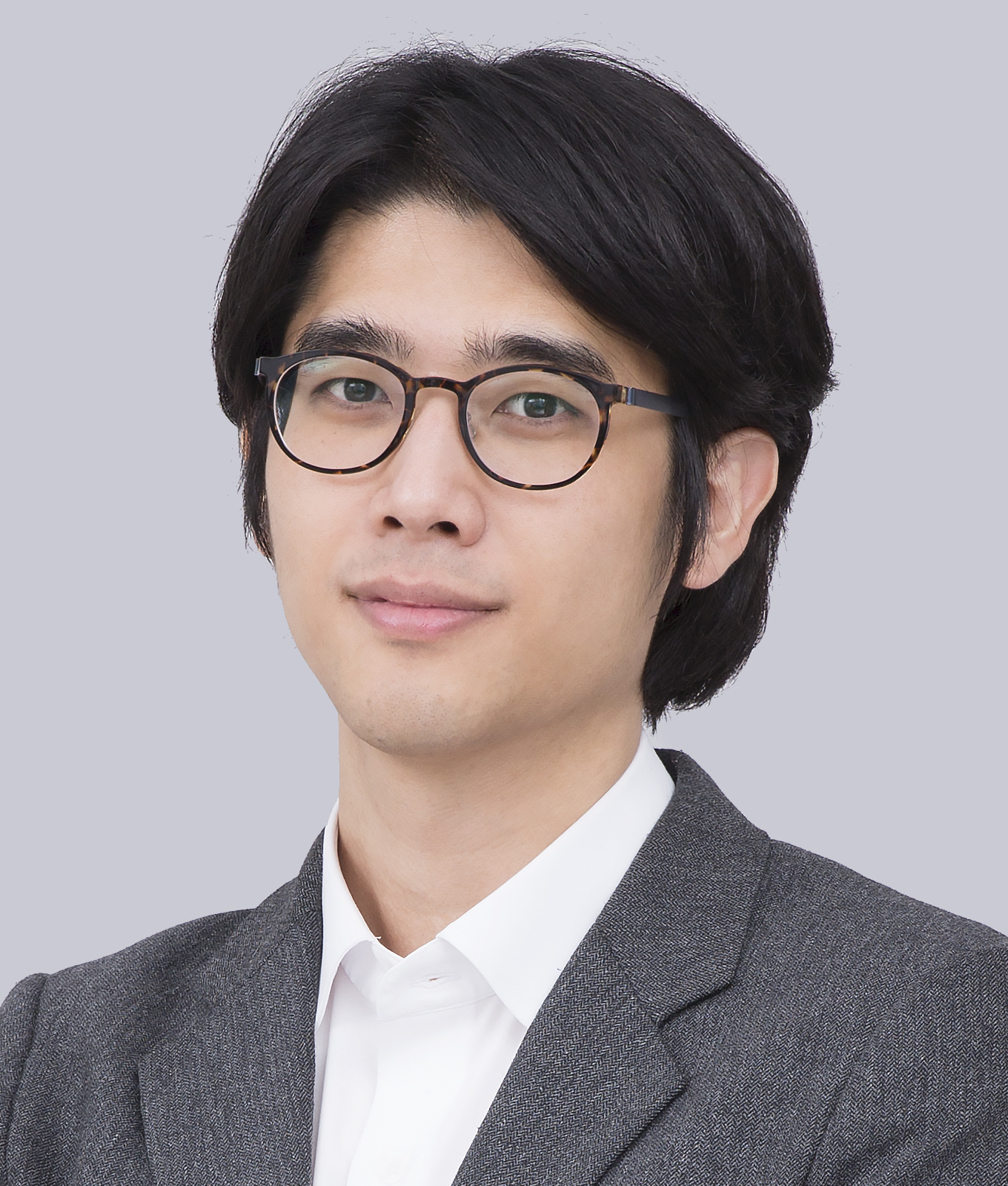}}]
	{Jihong Park} (S’09-M’16) is a Lecturer (assistant professor) at the School of IT, Deakin University, Australia. He received the B.S. and Ph.D. degrees from Yonsei University, Seoul, Korea, in 2009 and 2016, respectively. He was a Post-Doctoral Researcher with Aalborg University, Denmark, from 2016 to 2017; the University of Oulu, Finland, from 2018 to 2019. His recent research focus includes communication-efficient distributed machine learning, distributed control, and distributed ledger technology, as well as their applications for beyond 5G/6G communication systems. He served as a Conference/Workshop Program Committee Member for IEEE GLOBECOM, ICC, and WCNC, as well as NeurIPS, ICML, and IJCAI. He received the IEEE GLOBECOM Student Travel Grant in 2014, the IEEE Seoul Section Student Paper Contest Bronze Prize in 2014, and the 6th IDIS-ETNEWS (The Electronic Times) Paper Contest Award sponsored by the Ministry of Science, ICT, and Future Planning of Korea. Currently, he is an Associate Editor of Frontiers in Data Science for Communications, a Review Editor of Frontiers in Aerial and Space Networks, and a Guest Editor of MDPI Telecom SI on ``millimeter wave communiations and networking in 5G and beyond."
	
\end{IEEEbiography}

\begin{IEEEbiography}[{\includegraphics[width=1in,height=1.25in,clip,keepaspectratio]{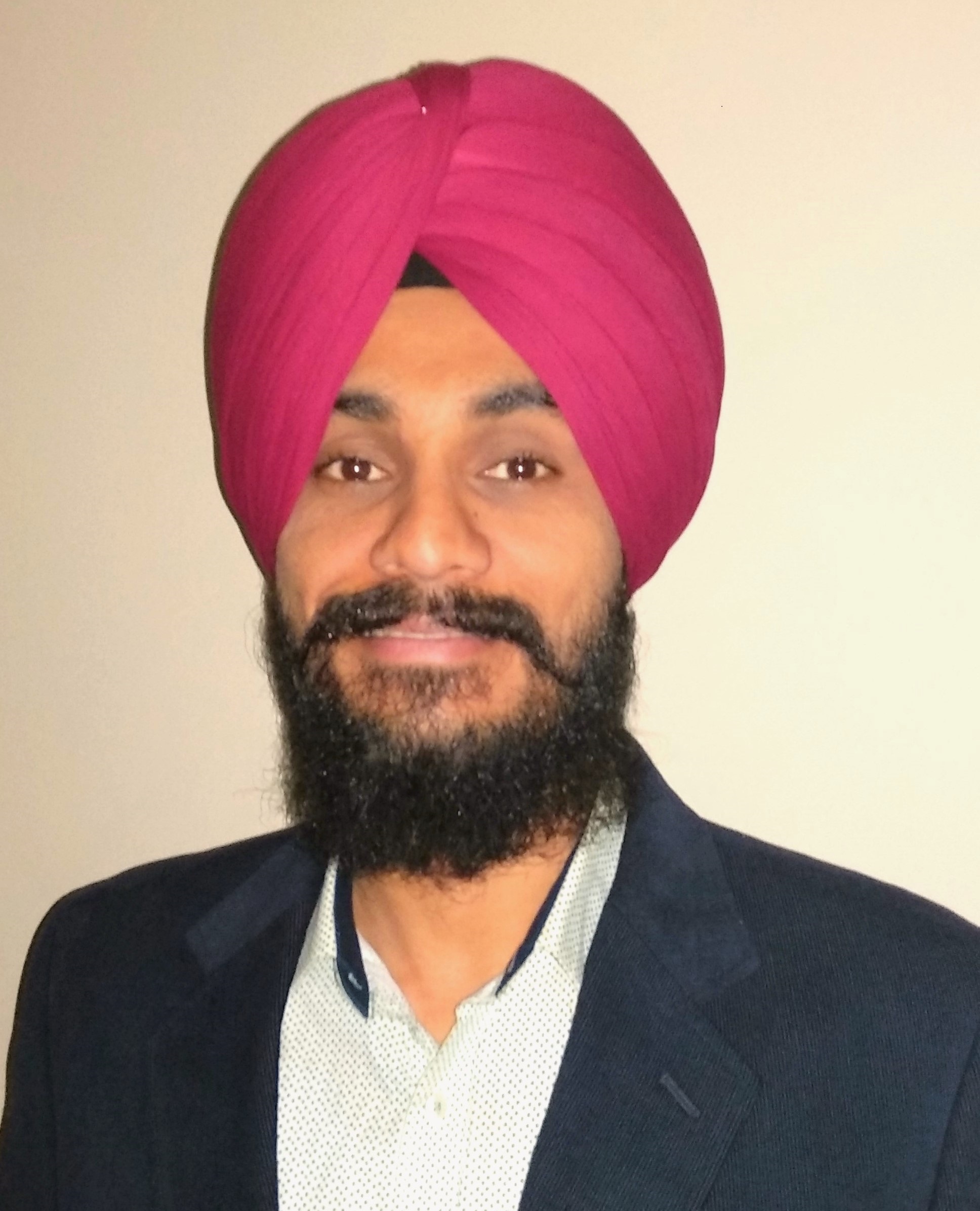}}]%
{Amrit Singh Bedi}
received the Diploma degree in electronics and communication engineering (ECE)
from Guru Nanak Dev Polytechnic, Ludhiana, India, in 2009, the B.Tech. degree in ECE from the Rayat and
Bahra Institute of Engineering and Bio-Technology, Kharar, India, in 2012, the M.Tech. and a Ph.D. degree in
electrical engineering (EE) from Indian Institute of Technology (IIT) Kanpur, in 2017 and 2018, respectively.
Currently, he is a postdoctoral fellow with the U.S. Army Research laboratory, Adelphi, MD, USA. He is
currently involved in developing stochastic optimization algorithms for supervised learning. His research
interests include distributed stochastic optimization for networks, time-varying optimization, and machine
learning. His paper selected as a Best Paper Finalist at the 2017 IEEE Asilomar Conference on Signals,
Systems, and Computers.

\end{IEEEbiography}

 \begin{IEEEbiography}[{\includegraphics[width=1in,height=1.25in, clip,keepaspectratio]{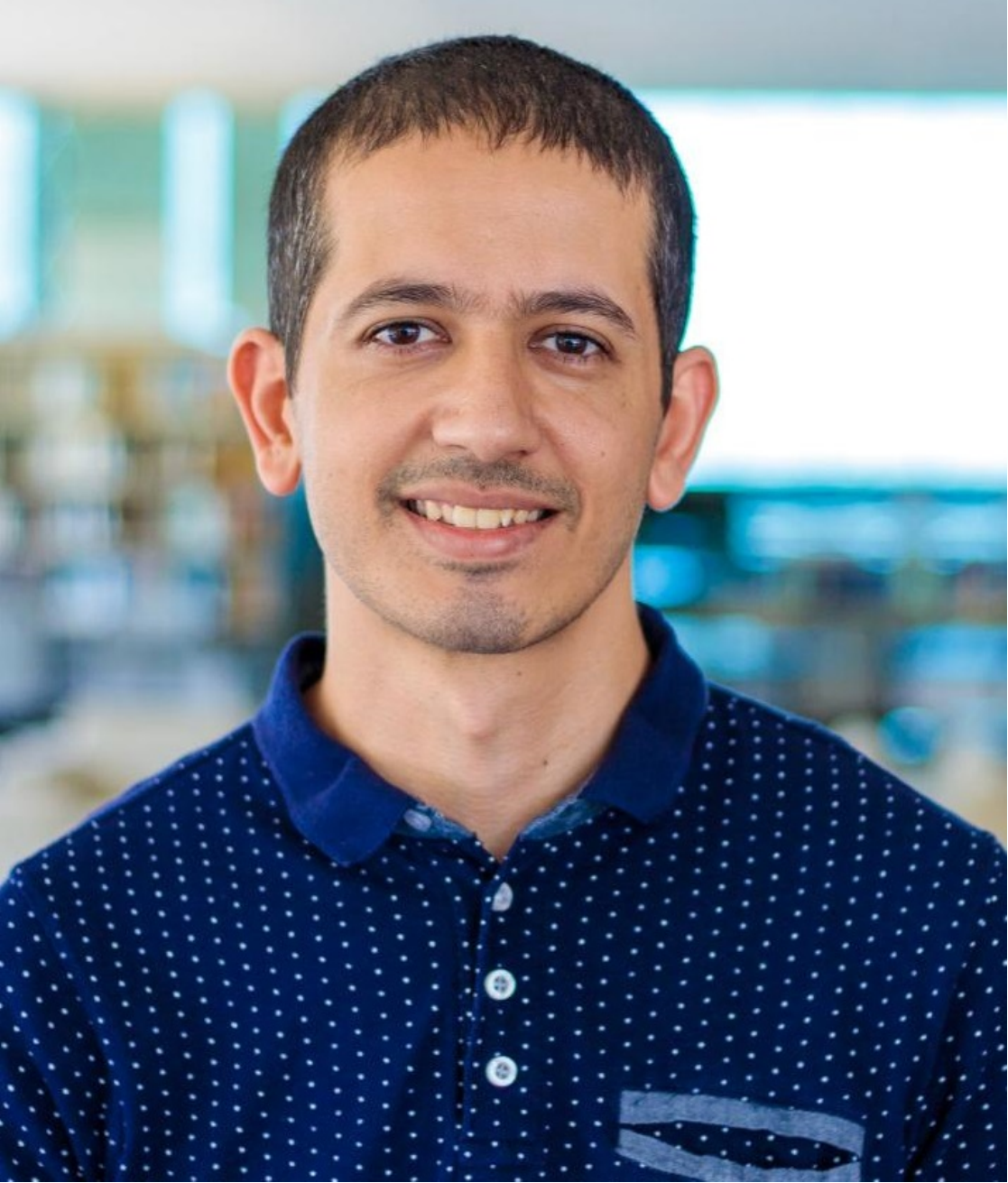}}]{Chaouki Ben Issaid} is a Postdoctoral Fellow at the Centre for Wireless Communications, University of Oulu. He received his Dipl\^{o}me d'Ingénieur with majors in Economics and Financial Engineering from Ecole Polytechnique de Tunisie (EPT) in 2013. Later on, he obtained his Master in Applied Mathematics and Computational Science (AMCS) and the Ph.D. degree in Statistics from KAUST in 2015 and 2019, repectively. His current research focus lies in the area of communication-efficient distributed machine learning. 
 \end{IEEEbiography}

\begin{IEEEbiography}
	[{\includegraphics[width=1in,height=1.25in,clip,keepaspectratio]{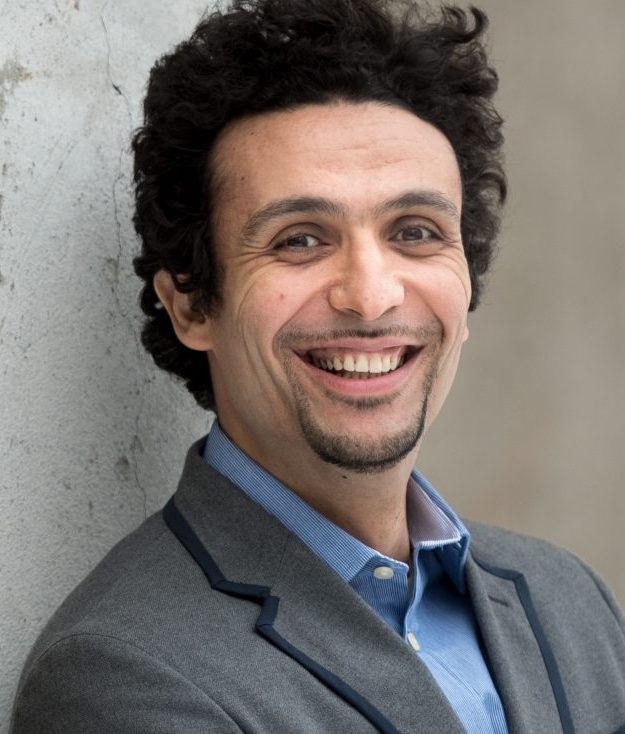}}]
	{Mehdi Bennis} is an Associate Professor at the Centre for Wireless Communications, University of Oulu, Finland, an Academy of Finland Research Fellow and head of the intelligent connectivity and networks/systems group (ICON). His main research interests are in radio resource management, heterogeneous networks, game theory and machine learning in 5G networks and beyond. He has co-authored one book and published more than 200 research papers in international conferences, journals and book chapters. He has been the recipient of several prestigious awards including the 2015 Fred W. Ellersick Prize from the IEEE Communications Society, the 2016 Best Tutorial Prize from the IEEE Communications Society, the 2017 EURASIP Best paper Award for the Journal of Wireless Communications and Networks, the all-University of Oulu award for research and the 2019 IEEE ComSoc Radio Communications Committee Early Achievement Award. Dr Bennis is an editor of IEEE TCOM.
\end{IEEEbiography}

\begin{IEEEbiography}
	[{\includegraphics[width=1in,height=1.25in,keepaspectratio]{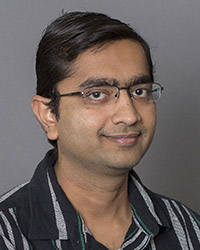}}]{Vaneet Aggarwal (S'08 - M'11 - SM'15)}
	received the B.Tech. degree from the Indian Institute of Technology, Kanpur, India in 2005, and the M.A. and Ph.D. degrees in 2007 and 2010, respectively from Princeton University, Princeton, NJ, USA, all in Electrical Engineering.
	
	He is currently an Associate Professor at Purdue University, West Lafayette, IN, where he has been since Jan 2015. He was a Senior Member of Technical Staff Research at AT\&T Labs-Research, NJ (2010-2014), Adjunct Assistant Professor at Columbia University, NY (2013-2014), and VAJRA Adjunct Professor at IISc Bangalore (2018-2019). His current research interests are in communications and networking, cloud computing, and machine learning.
	
	Dr. Aggarwal received Princeton University's Porter Ogden Jacobus Honorific Fellowship in 2009, the AT\&T Vice President Excellence Award in 2012, the AT\&T Key Contributor Award in 2013, the AT\&T Senior Vice President Excellence Award in 2014, the 2017 Jack Neubauer Memorial Award recognizing the Best Systems Paper published in the IEEE TRANSACTIONS ON VEHICULAR TECHNOLOGY, and the 2018 Infocom Workshop HotPOST Best Paper Award. He is on the Editorial Board of the IEEE TRANSACTIONS ON COMMUNICATIONS, the IEEE TRANSACTIONS ON GREEN COMMUNICATIONS AND NETWORKING, and the IEEE/ACM TRANSACTIONS ON NETWORKING.\end{IEEEbiography}

\end{document}